\documentclass{article}

\usepackage{microtype}
\usepackage{graphicx}
\usepackage{subfigure}
\usepackage{booktabs} 
\usepackage{multirow}

\usepackage{xcolor}

\usepackage[breaklinks=true]{hyperref}
\usepackage{etoolbox}
\appto\UrlBreaks{\do\-}

\usepackage[accepted]{icml2021}

\icmltitlerunning{GCExplainer: Human-in-the-Loop Concept-based Explanations for Graph Neural Networks}

\begin{document}

\twocolumn[
\icmltitle{GCExplainer: Human-in-the-Loop Concept-based Explanations for Graph Neural Networks}



\icmlsetsymbol{equal}{*}

\begin{icmlauthorlist}
\icmlauthor{Lucie Charlotte Magister}{to}
\icmlauthor{Dmitry Kazhdan}{to}
\icmlauthor{Vikash Singh}{to}
\icmlauthor{Pietro Li\`{o}}{to}
\end{icmlauthorlist}

\icmlaffiliation{to}{Department of Computer Science and Technology, University of Cambridge, Cambridge, United Kingdom}

\icmlcorrespondingauthor{Lucie Charlotte Magister}{lcm67@cam.ac.uk}

\icmlkeywords{Machine Learning, ICML, GNN, Concept, Explainability, Interpretability}

\vskip 0.3in
]



\printAffiliationsAndNotice{}  

\begin{abstract}
While graph neural networks (GNNs) have been shown to perform well on graph-based data from a variety of fields, they suffer from a lack of transparency and accountability, which hinders trust and consequently the deployment of such models in high-stake and safety-critical scenarios. Even though recent research has investigated methods for explaining GNNs, these methods are limited to single-instance explanations, also known as local explanations. Motivated by the aim of providing global explanations, we adapt the well-known Automated Concept-based Explanation approach \cite{Ghorbani2019} to GNN node and graph classification, and propose GCExplainer. GCExplainer is an unsupervised approach for post-hoc discovery and extraction of global concept-based explanations for GNNs, which puts the human in the loop. We demonstrate the success of our technique on five node classification datasets and two graph classification datasets, showing that we are able to discover and extract high-quality concept representations by putting the human in the loop. We achieve a maximum completeness score of 1 and an average completeness score of 0.753 across the datasets. Finally, we show that the concept-based explanations provide an improved insight into the datasets and GNN models compared to the state-of-the-art explanations produced by GNNExplainer~\cite{Ying2019}.
\end{abstract}

\section{Introduction}
\label{introduction}
Graph Neural Networks (GNNs) are a class of deep learning methods for reasoning about graphs~\cite{Bacciu2020}. Their significance lies in incorporating both the feature information and structural information of a graph, which allows deriving new insights from a plethora of data \cite{Ying2019, Lange2020, Barbiero2020, Xiong2021}. Unfortunately, similar to other deep learning methods, such as convolutional neural networks (CNNs) and recurrent neural networks (RNNs), GNNs have a notable drawback: the computations that lead to a prediction cannot be interpreted directly \cite{Tjoa2015}. In addition, it can be argued that the interpretation of GNNs is more difficult than that of alternative methods, as two types of information, feature and structural information, are combined during decision-making \cite{Ying2019}. This lack of transparency and accountability impedes trust \cite{Tjoa2015}, which in turn hinders the deployment of such models in safety-critical scenarios. Furthermore, this opacity hinders gaining an insight into potential shortcomings of the model and understanding what aspects of the data are significant for the task to be solved \cite{Ying2019}.

Recent research has attempted to improve the understanding of GNNs by producing various explainability techniques \cite{Pope2019, Baldassarre2019, Ying2019, Schnake2020, Luo2020, Vu2020}. A prominent example is GNNExplainer \cite{Ying2019}, which finds the sub-graph structures relevant for a prediction by maximising the mutual information between the prediction of a GNN and the distribution of possible subgraphs. However, a significant drawback of this method is that the explanations are local, which means they are specific to a single prediction.

The goal of our work is to improve GNN explainability through concept-based explanations, which are produced by putting the human in the loop. Concept-based explanations are explanations in the form of small higher-level units of information \cite{Ghorbani2019}. The exploration of such explanations is motivated by concept-based explanations being easily accessible by humans \cite{Ghorbani2019, Kazhdan2021}. Moreover, concept-based explanations act as a global explanation of the class, which improves the overall insight into the model \cite{Ghorbani2019, Yeh2019}. We pay particular attention to involving the user in the discovery and extraction process, as ultimately the user must reason about the explanations extracted.

To the best of our knowledge, this is the first work that explores human-in-the-loop concept representation learning in the context of GNNs. We make the following contributions, with the source code available on GitHub \footnote{Source code available at: \url{https://github.com/CharlotteMagister/GCExplainer}}:
\begin{itemize}
    \item an unsupervised method for concept discovery and extraction in GNNs for post-hoc explanation, which takes the user into account;
    \item the resulting tool Graph Concept Explainer (GCExplainer);
    \item a metric evaluating the purity of concepts discovered;
    \item an evaluation of the approach on five node classification and two graph classification datasets;
\end{itemize}

\begin{figure}[ht]
\vskip 0.2in
\begin{center}
\centerline{\includegraphics[width=\columnwidth]{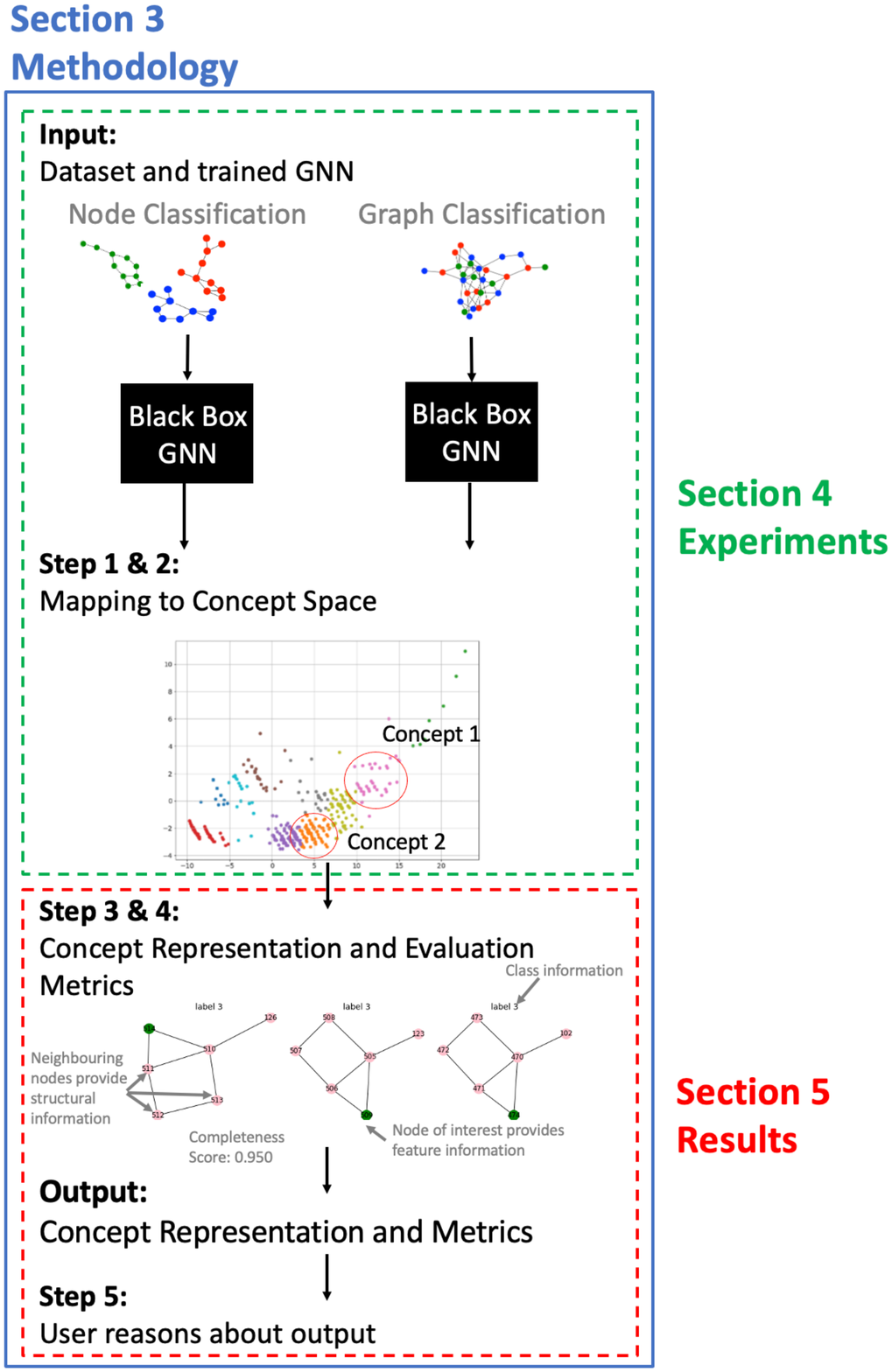}}
\caption{Overview of the methodology.}
\label{visualabstract}
\end{center}
\vskip -0.2in
\end{figure}

Our method is based on the Automated Concept-based Explanation (ACE) algorithm \cite{Ghorbani2019} and the work presented by \citet{Yeh2019}. Figure \ref{visualabstract} provides an overview of the methodology proposed. Our results show that GCExplainer allows to discover and extract high quality concepts, outperforming GNNExplainer \cite{Ying2019} based on our qualitative evaluation. Our results suggest that GCExplainer can be used as a general framework due to its strong degree of generality, which finds application across a wide variety of GNN-based learning tasks.

\section{Related Work}
\label{related_work}

\subsection{Graph Neural Network Explainability}

Early research on GNN explainability focuses on adapting CNN-based techniques \cite{Pope2019, Baldassarre2019, Schnake2020}, such as gradient-based saliency maps, class activation mapping and excitation backpropagation \cite{Pope2019}. While such techniques allow to highlight features important for a prediction, they have received critique for being misleading \cite{Adebayo2018} and susceptible to input perturbations \cite{Kindermans2019}. More recent research has focused on GNN-specific explainability techniques, which take into account the graph structure \cite{Ying2019, Luo2020, Vu2020}. A prominent example is the model-agnostic technique GNNExplainer \cite{Ying2019}. However, a drawback of GNNExplainer is that the design is focused on local explanations and requires retraining for each prediction. Arguably, local explanations are insufficient, as they can be misleading if the limits of the domain are unknown \cite{Mittelstadt2019, Lakkaraju2020}, which has been demonstrated for CNNs \cite{Alvarez-Melis2018}. Even though GNNExplainer attempts to produce global explanations for a given class by taking 10 instances of the class and performing graph alignment on the subgraph explanations, the ability to confirm the global explanation is limited by the efficiency of graph alignment, which is an NP-Hard problem \cite{Ying2019}.

PGExplainer \cite{Luo2020} aims to address the issues of GNNExplainer. PGExplainer is a model-agnostic explainer with the same optimisation task, however, the key difference is that it uses a deep neural network (DNN) for the parameterisation of the explanation generation process. While PGExplainer claims to provide global explanations, the explanations are not truly global, but simply multi-instance explanations. In a similar manner, PGM-Explainer \cite{Vu2020} extracts relevant subgraphs for a prediction with the added benefit of feature dependencies being indicating using conditional probabilities.

\subsection{Concept-Based Explainability}

Concept-based explanations have been explored in various ways for other neural networks, such as CNNs \cite{Alvarez-Melis2018, Kim2018, Ghorbani2019, Yeh2019,  Koh2020, Chen2020} and RNNs \cite{Kazhdan2020_MEME}, as well as for detecting dataset distribution shifts \cite{Wijaya2021}. In general, these approaches can be split into supervised and unsupervised concept extraction, where supervised approaches provide a predefined set of concepts. For example, \citet{Alvarez-Melis2018} propose a method for encouraging interpretability during training through a concept encoder, while \citet{Koh2020} introduce concept bottleneck models, which are interpretable by design and allow human intervention. Similarly, Testing with Concept Activation Vectors (TCAV) \cite{Kim2018} allows to quantify the importance of a predefined set of concepts for a prediction through directional derivatives computed from the activation space. In contrast, \citet{Kazhdan2020} present the CME framework for DNNs, which is a semi-supervised approach that learns an intermediate concept representation from an only partially-labelled dataset.

In comparison to the previously mentioned techniques, unsupervised concept extraction approaches perform automatic concept discovery. For example, the ACE algorithm \cite{Ghorbani2019} automatically extracts visual concepts in image classification by segmenting the input image and clustering similar segments in the activation space of deeper layers in a CNN. The approach is based on the observation that the learned activation space is similar to human perceptual judgement \cite{Zhang2018}. Similarly, \citet{Yeh2019} also extract concept-based explanations from the activation space in an unsupervised manner, by training a topic model on the features extracted.

\section{Methodology}
\label{method}

\subsection{Concept Discovery and Extraction}
We propose to perform concept discovery for GNNs by establishing a mapping from the activation space to the concept space via clustering. In particular, we suggest performing $k$-Means clustering on the raw activation space of the last neighbourhood aggregation layer in a GNN, where each of the $k$ clusters formed represents a concept. The proposed method is an adaption of the ACE algorithm \cite{Ghorbani2019} for concept discovery in CNNs, which is based on earlier research suggesting that the arrangement of the activation space shows similarities to human perceptual judgement \cite{Zhang2018}, as well as the work presented by \citet{Yeh2019}.

We suggest the use of the $k$-Means clustering algorithm for a number of reasons. Firstly, the $k$-Means algorithm minimises the variance within a cluster \cite{Rokach2009}. This is desirable as we would like to group nodes of the same type, rather than neighbouring nodes, for global explanations that meet the properties for concepts defined by \citet{Ghorbani2019}. Secondly, $k$-Means is a user-friendly algorithm that allows exploration by the user. $k$ can be picked by visualising the activation space and estimating the number of clusters using a silhouette plot or based on the user's understanding of the concepts. As a cluster is representative of a concept, an alternative approach is to fine-tune $k$ by minimising the redundancy in the set of concepts while maximising its completeness score \cite{Yeh2019, Kazhdan2020}. Lastly, $k$-Means is also efficient to compute and works on multiscale data \cite{Rokach2009}. 

We suggest clustering the activation space produced by the final neighbourhood aggregation layer in a GNN, as later layers provide improved node representations, allowing improved clustering. In the context of graph classification, there is potential for extracting concepts at the graph level by performing clustering on the pooling layers. We leave this for future work, as it requires additional reasoning across graphs to observe concepts, such as finding the maximum common subgraph between the graphs clustered together. Lastly, we propose performing this clustering on the raw activation space in order to avoid the loss of information through dimensionality reduction (DR). 

Using the recovered set of concepts, a global explanation for an input instance can be extracted by finding the cluster in which the input instance falls. In the context of node classification, this is achieved by mapping a node in the graph to its representation in the activation space, which in turn can be mapped to a cluster analogous of a concept. The representation of a node in the activation space can simply be found by passing the instance through the GNN. In the context of graph classification, concept extraction can be performed by extracting the concepts associated with each node and aggregating over those. The overall contribution of a concept to the prediction of a class can simply be computed by calculating how many of the nodes in a class are clustered as the given concept. The most dominant concepts found this way can then be used to explain the prediction. Computing the concept contribution for the predicted class labels and the actual set of a class labels, can potentially allow identifying shortcomings in the model.

\subsection{Concept Representation}
\label{concept_rep}

We propose a concept representation step in the method to allow the user to reason about the concepts as explanations. Specifically, we propose plotting the $n$-hop neighbours of a node to build a subgraph. This is based on both structural and feature information being included in the representation learned by a GNN, implying that a subgraph is the most suitable representation. It can also be argued that concept-based explanations naturally take the modality of the input data. For example, concept-based explanations for image classifiers are usually superpixels, while bags of words represent concepts in text classification \cite{Yeh2019}.

We leave the value $n$ to be chosen by the user. This is motivated by two key reasons. Firstly, the opportunity for interactive concept extraction allows the user to perform their own exploration based on their domain knowledge of the dataset \cite{Chari2020}. Secondly, concept representation learning is difficult as it is dependent on the model architecture and performance. Hence, allowing the user to choose a suitable value for $n$ allows greater flexibility in varying the size and complexity of the concept, which in turn makes it easier for the user to reason about the concept. 

\subsection{Evaluation Metrics}

\subsubsection{Concept Purity}

We propose evaluating the concept representations extracted on concept purity. We define concept purity as the concept representations across a cluster being identical, which indicates that the concept is a strong, coherent global concept. We propose using the graph edit distance as a measure for this. The graph edit distance measures the number of operations that must be performed to transform graph $g_1$ into graph $g_2$, which can be formalised as \cite{Abu-Aisheh2015}:

\begin{equation}
  GED(g_1, g_2) = \min_{e_1, \ldots, e_k \in \gamma(g_1, g_2)} \sum_{i=1}^{k} c(e_i)
\end{equation}

where, $c$ is the cost function for an edit operation $e_i$ and $\gamma(g_1, g_2)$ is the set of edits which allow to transform $g_1$ into $g_2$. If the graph edit distance between concept representations of the same class is 0, then the concept is pure. Else, it can be argued that the concept is not pure indicating that $n$ must be increased or decreased. It can potentially also indicate that multiple concepts are grouped in the same cluster and that consequently $k$ must be increased.

\subsubsection{Concept Completeness}

An alternative metric for evaluating the set of concepts discovered is to calculate the concept completeness score, as proposed by \citet{Yeh2019} and similar to the model fidelity score proposed by \citet{Kazhdan2020}. The completeness score is the accuracy of a classifier, which predicts the output label $y$ for an input instance $x$ using the assigned concept $c$ as input. The intuition behind this is that the testing accuracy of the classifier represents the completeness of the concepts, as it indicates the accuracy by which the final output can be predicted from the concepts. A suitable classifier choice is a decision tree \cite{Kazhdan2020}, as the way concept labels are assigned using $k$-Means does not necessarily construct a space easily separable by other classifiers, such as logistic regression.

Furthermore, we propose concept heuristics for evaluation, which we detail in Appendix \ref{A} for brevity. 


\section{Experiments}
\label{experiments}

\subsection{Experimental Setup}

To evaluate the concept representation learning method proposed, we define a set of experiments. The goal of these experiments is twofold. Firstly, we aim to demonstrate the benefits of putting the human in the loop in the definition of $k$ and $n$. Secondly, we aim to demonstrate the success of the method on a range of datasets. For a systematic evaluation, we define the following experiments:
\begin{enumerate}
    \item \textbf{Adaption of $k$:} We adapt $k$, changing the number of concepts extracted.
    \item \textbf{Adaption of $n$:} We adapt $n$, changing the complexity of the concepts.
    \item \textbf{Datasets:} We apply the method on a range of datasets.
\end{enumerate}

Furthermore, we benchmark the proposed method against different designs to validate out design choices. We refer the reader to Appendix \ref{B} for the rational and analysis.

\subsection{Datasets}
\label{datasets}

We perform the experiments on the same set of datasets as GNNExplainer \cite{Ying2019}, as GNNExplainer has been established as a benchmark by subsequent research~\cite{Luo2020, Vu2020}.

\subsubsection{Synthetic Datasets}

GNNExplainer~\cite{Ying2019} presents five synthetic node classification datasets, which have a ground truth motif encoded. A ground truth motif is a subgraph in the graph, which a successful GNN explainability technique should recognise. Table \ref{motifs} summarises the datasets and their motifs.

The first dataset is BA-Shapes, which consists of a single graph of 700 nodes \cite{Ying2019}. The base structure of the graph is a Barab\'asi-Albert (BA) graph ~\cite{Barabasi2016} with 300 nodes, which has 80 house structures and 70 random edges attached to it. The task is to classify the nodes into four categories: top-node in a house structure, middle node in a house structure, bottom node in a house structure or node in the base graph. 

The second dataset is BA-Community, which is a union of two BA-Shapes graphs with the task of classifying the nodes into eight classes. These classes differentiate between the types of nodes in the house structure, as well as the membership of the node in one of the graphs. Similarly, the third dataset BA-Grid is also based on the BA graph, however, a 3-by-3 grid structure is attached instead of the house structure ~\cite{Ying2019}.

The fourth synthetic dataset is Tree-Cycles. The base graph of this dataset is formed by a balanced binary tree with a depth of eight, which has 80 cycle structures attached to it. The classification task requires differentiating between nodes part of the tree structure and cycle structure. The Tree-Grid dataset is constructed in the same manner, however, the cycle structures are replaced with 3-by-3 grid structures.

\begin{table*}[t]
\caption{Summary of the node classification datasets, along with examples of the motifs to be extracted.}
\label{motifs}
\vskip 0.15in
\begin{center}
\begin{small}
\begin{sc}
{%
\begin{tabular}{|l|c|c|c|c|c|}
\hline
\textbf{Dataset} & \textbf{BA-Shapes} & \textbf{BA-Community} & \textbf{BA-Grid} & \textbf{Tree-Cycles} & \textbf{Tree-Grid} \\ \hline
\textbf{Base Graph}    & BA Graph           & BA Graph         &  BA Graph             & \begin{tabular}[c]{@{}c@{}}Binary Tree\end{tabular} & \begin{tabular}[c]{@{}c@{}}Binary Tree\end{tabular} \\ \hline
\textbf{Motif}        & \includegraphics[scale=0.1]{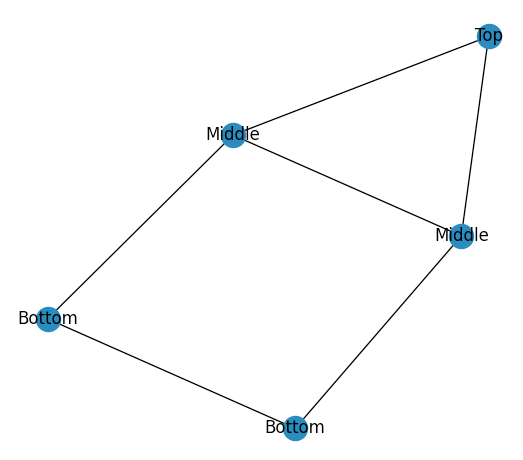} &  \includegraphics[scale=0.1]{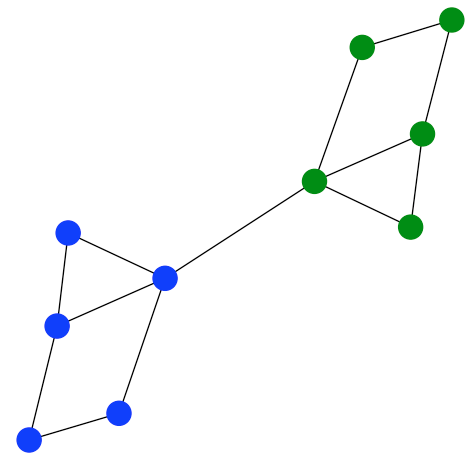} & \includegraphics[scale=0.1]{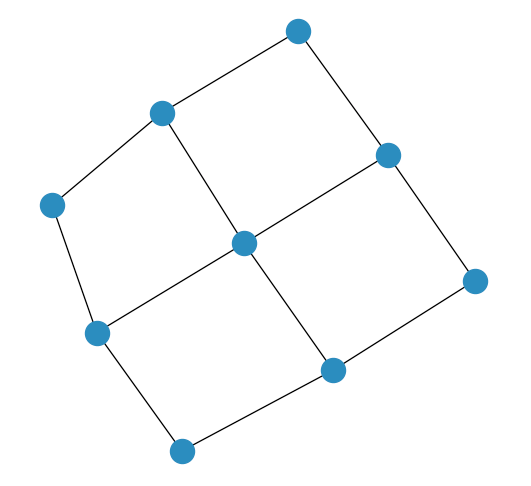} &  \includegraphics[scale=0.1]{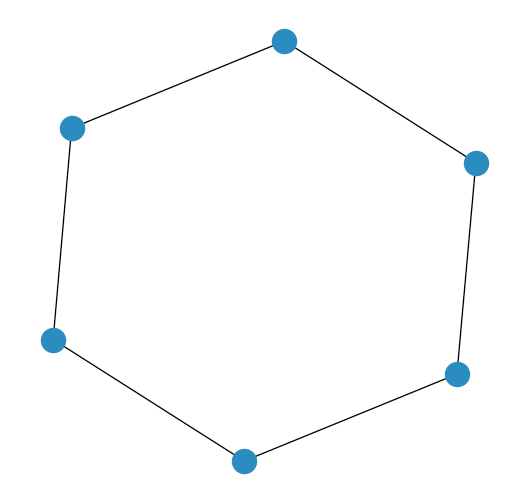} & \includegraphics[scale=0.1]{imgs/random/grid.png}                   \\ \hline
\textbf{Node Features} & None               & Community Id  & None          & None                & None \\ \hline
\end{tabular} %
}
\end{sc}
\end{small}
\end{center}
\vskip -0.1in
\end{table*}

\subsubsection{Real-World Datasets}

Besides allowing a comparison with GNNExplainer, the real-world datasets allow to establish the performance of the method on less structured data and on graph classification tasks. However, a challenge in evaluating these datasets is that there are no ground truth motifs. Table \ref{graphclass} summarises the two datasets and the motifs to be extracted as concepts.

The Mutagenicity dataset~\cite{KKMMN2016} is a collection of graphs representing molecules, which are labelled as mutagenic or non-mutagenic. A molecule is labelled as mutagenic if it causes a problem in the replication of the Gram-negative bacterium S. typhimurium \cite{Ying2019}. The motifs to be extracted are a cyclic structure and the substructure $NO_2$ for mutagenic molecules. 

The REDDIT-BINARY dataset~\cite{KKMMN2016} is a collection of graphs representing the structure of Reddit discussion threads, where nodes represent users and edges the interaction between users. The task is to classify the graphs into two types of online interactions. According to GNNExplainer~\cite{Ying2019}, the motifs to identify are reactions to a topic and expert answers to multiple questions.

\begin{table}[t]
\caption{Summary of the graph classification datasets, along with examples of the motifs to be extracted.}
\label{graphclass}
\vskip 0.15in
\begin{center}
\begin{small}
\begin{sc}
{%
\begin{tabular}{|l|l|l|}
\hline
\textbf{Dataset}     & \textbf{Mutagenicity} & \textbf{REDDIT-BINARY} \\ \hline
\textbf{Motifs}      &  \includegraphics[scale=0.1]{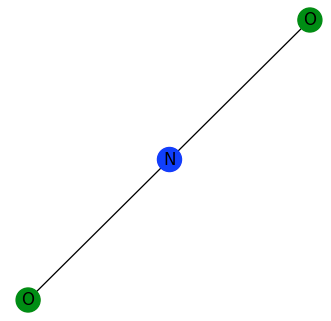} \includegraphics[scale=0.05]{imgs/random/cycle.png}  & \includegraphics[scale=0.1]{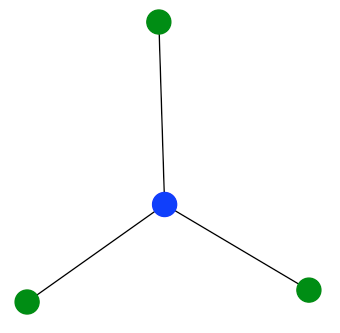} \includegraphics[scale=0.1]{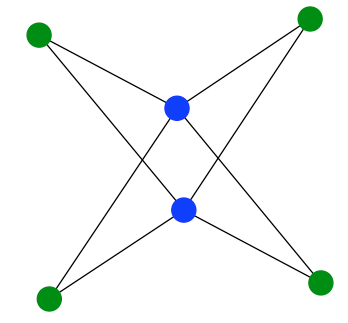} \\ \hline
\textbf{Node Labels} & Atom                  & None                   \\ \hline
\end{tabular}%
}
\end{sc}
\end{small}
\end{center}
\vskip -0.1in
\end{table}

\subsection{Implementation}
\label{impldetails}

We train a GNN model for each dataset. As recommended by GNNExplainer \cite{Ying2019}, we aim for a model accuracy of at least 95\% and 85\% for node and graph classification, respectively. We refer the reader to Appendix \ref{C} for implementation details.

\section{Results}
\label{results}
Section \ref{results2} focuses on the effects of $k$ and $n$, while Section \ref{results1} presents the performance of the method across datasets.

\subsection{Evaluation of Human-in-the-Loop Parameter Adaption}
\label{results2}

For brevity, we will focus on discussing the results for the BA-Shapes dataset, as the ground truth motif to be discovered is known, aiding the evaluation. We refer the reader to Appendix \ref{D} for results on further datasets. Table \ref{new_res1} summarises the concept completeness and purity scores obtained using the method presented when adapting the values $n$ and $k$. The choice in the range of $n$ was guided by the knowledge of the dataset: at least the 2-hop neighbourhood must be visualised to cover the full concept. The choice in the range of $k$ was guided by the examination of the clustered activation space.

Focusing on the concept completeness scores, it can be stated that the highest concept completeness scores are obtained when $k = 10$. The score is influenced by the number of concepts discovered, hence, it only depends on $k$ and not $n$. In contrast, the purity score depends on $n$, the size of the concept visualised. Reviewing the purity scores within the context of $k$, it can be stated that $n = 2$ is the favourable setting across the different $k$ defined. This is due to this value allowing to visualise the whole concept. Decreasing $n$ gives a limited overview of the concept, while a $n$ that is too large leads to additional nodes and edges that are not relevant to the concept to be included, which in turn leads to a worse concept purity score. Examining these settings together, it can be concluded that the best setting within the parameter range explored is $k = 10$ and $n = 2$.

Another observation that can be made is that $k$ partially influences the purity score. For example, in general worse concept purity scores are achieved when $k = 5$ in contrast to $k = 15$. This can be explained by a lower $k$ causing different concepts to be grouped together, which when compared lead to a high graph edit distance.

\begin{table}[t]
\caption{The concept completeness and purity score of the concepts discovered for the BA-Shapes dataset when adapting $n$ and $k$.}
\label{new_res1}
\vskip 0.15in
\begin{center}
\begin{small}
\begin{sc}
{%
\begin{tabular}{|c|c|c|c|}
\hline
\textbf{k}          & \textbf{n} & \textbf{\begin{tabular}[c]{@{}c@{}}Completeness \\ Score\end{tabular}} & \textbf{\begin{tabular}[c]{@{}c@{}}Average Purity\\ Score\end{tabular}} \\ \hline
\multirow{3}{*}{5}  & 1          & 0.810                                                                  & 8.375                                                                      \\ \cline{2-4} 
                    & 2          & 0.810                                                                  & 1.000                                                                      \\ \cline{2-4} 
                    & 3          & 0.810                                                                  & 6.000                                                                      \\ \hline
\multirow{3}{*}{10} & 1          & 0.964                                                                  & 3.500                                                                      \\ \cline{2-4} 
                    & 2          & 0.964                                                                  & 3.375                                                                      \\ \cline{2-4} 
                    & 3          & 0.964                                                                  & 5.500                                                                      \\ \hline
\multirow{3}{*}{15} & 1          & 0.956                                                                  & 3.455                                                                      \\ \cline{2-4} 
                    & 2          & 0.956                                                                  & 2.417                                                                      \\ \cline{2-4} 
                    & 3          & 0.956                                                                  & 0.000                                                                      \\ \hline
\end{tabular}%
}
\end{sc}
\end{small}
\end{center}
\vskip -0.1in
\end{table}

\begin{figure}[ht]
\vskip 0.2in
\begin{center}
\centerline{\includegraphics[width=\columnwidth, trim = 0 0cm 1cm 0cm, clip]{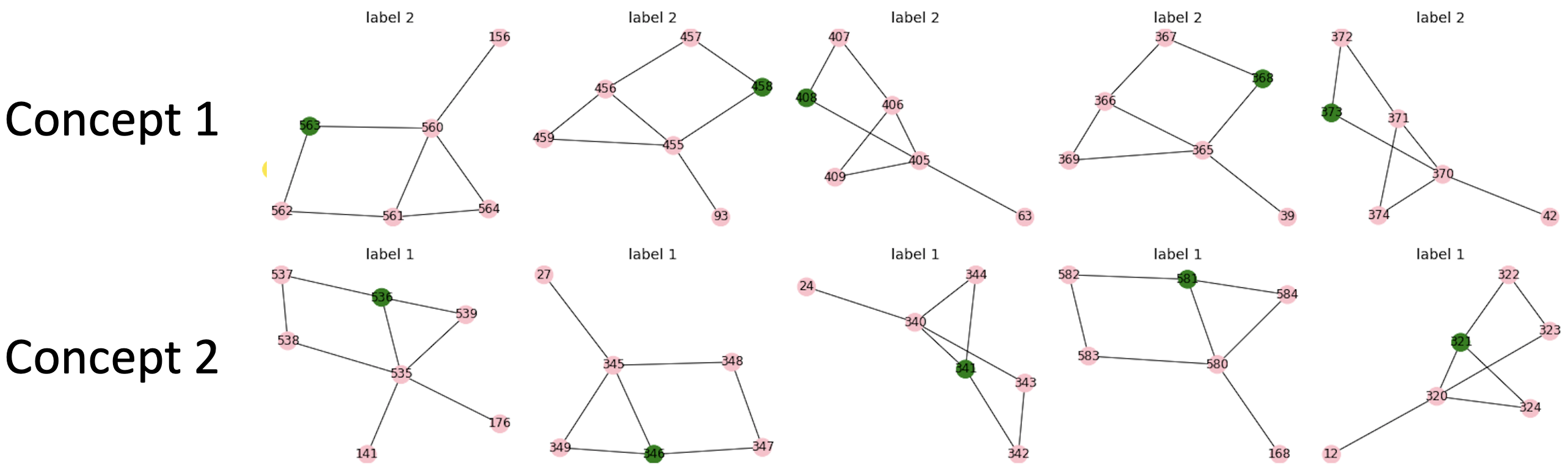}}
\caption{A subset of the concepts discovered for BA-Shapes, where $k = 5$ and $n = 2$.}
\label{new_res2}
\end{center}
\vskip -0.2in
\end{figure}

\begin{figure}[ht]
\vskip 0.2in
\begin{center}
\centerline{\includegraphics[width=\columnwidth, trim = 0 8cm 0 0cm, clip]{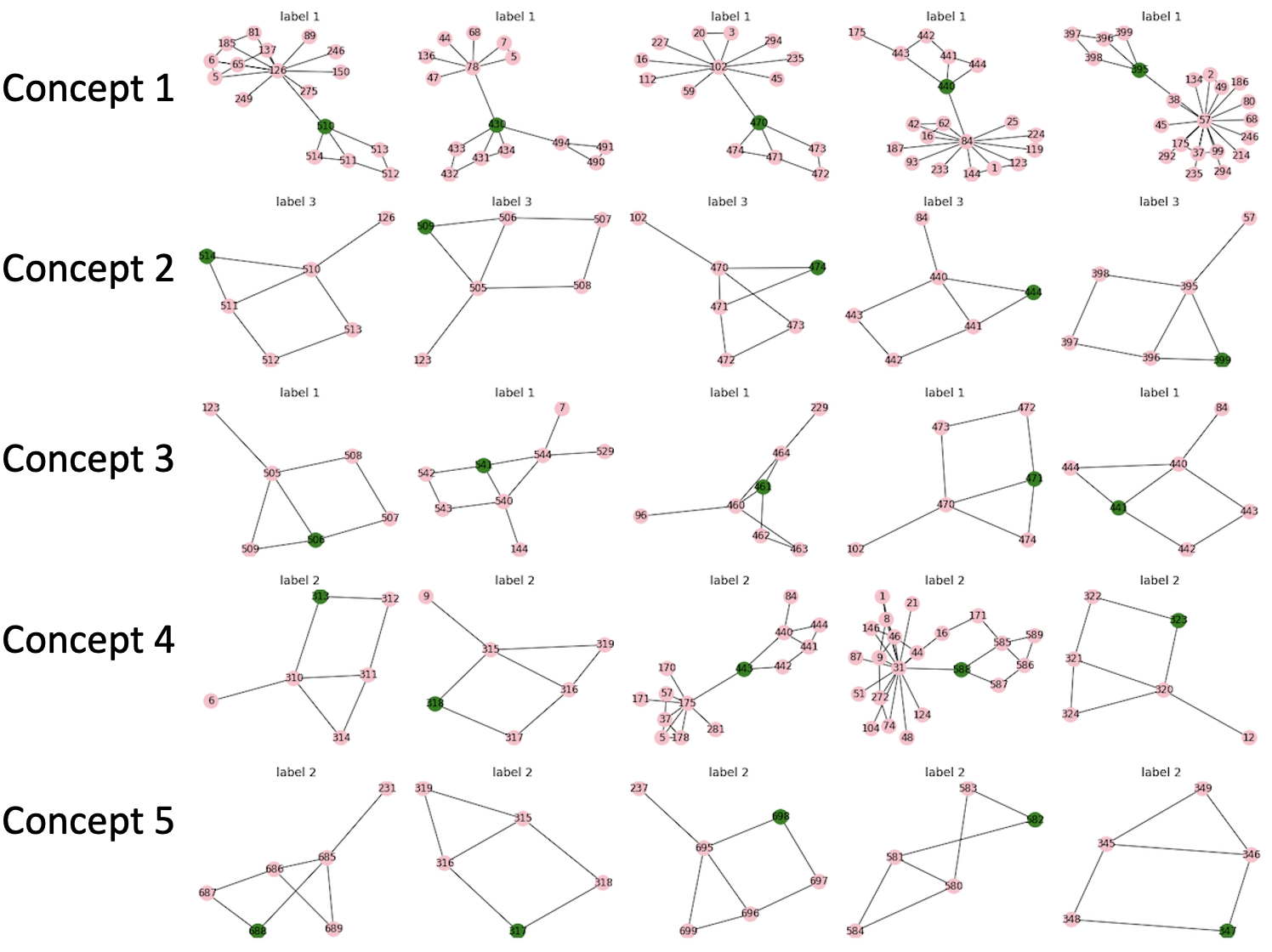}}
\caption{A subset of the concepts discovered for BA-Shapes, where $k = 10$ and $n = 2$.}
\label{new_res3}
\end{center}
\vskip -0.2in
\end{figure}

\begin{figure}[ht]
\vskip 0.2in
\begin{center}
\centerline{\includegraphics[width=\columnwidth, trim = 0 0 0 0cm, clip]{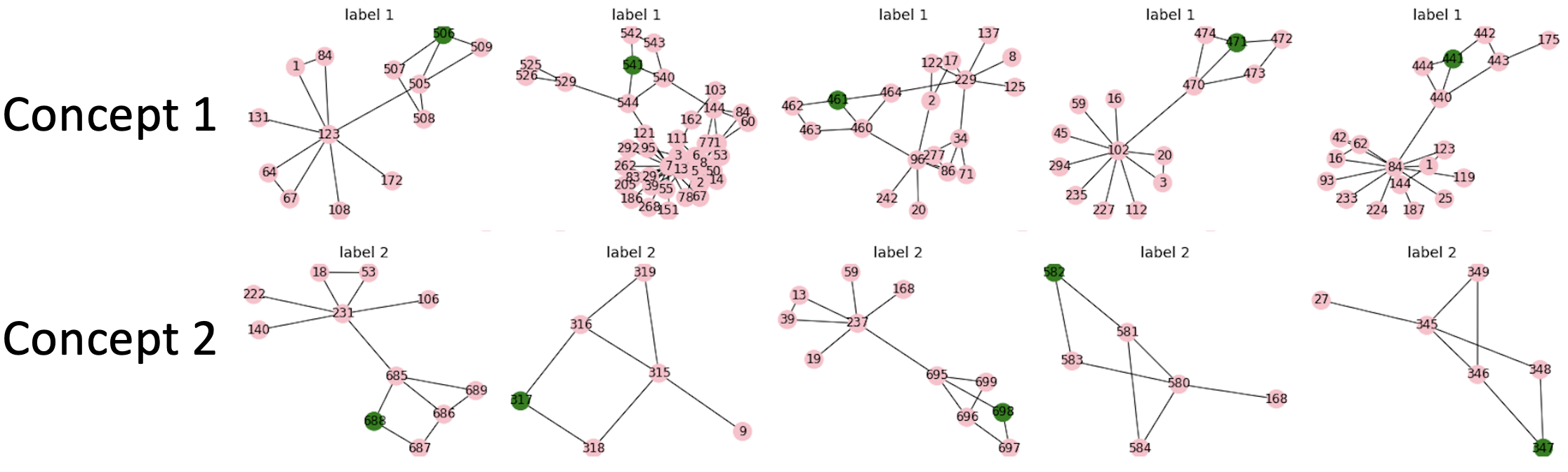}}
\caption{A subset of the concepts discovered for BA-Shapes, where $k = 10$ and $n = 3$.}
\label{new_res4}
\end{center}
\vskip -0.2in
\end{figure}

The conclusions drawn are also supported by the qualitative results obtained. Figure \ref{new_res2} shows a subset of the concepts discovered when $k = 5$ and $n = 2$, while Figure \ref{new_res3} shows a subset of the concepts discovered when $k = 10$ and $n = 2$. To allow reasoning about the concept represented by a cluster, we visualise the top five concept representations closest to a cluster centroid. The green nodes are the nodes clustered together, while the pink nodes are the neighbourhood explored. Comparing the visualisations, it can be stated that more fine grained concepts are extracted when $k = 10$, which also reflects in the improved concept completeness score seen previously. In contrast, Figure \ref{new_res4} shows a subset of concepts extracted when $k = 10$ and $n = 3$. Comparing the concepts extracted more closely, it can be stated that it is harder to identify the concept, because the subgraphs are larger and convoluted by nodes of the BA base graph, which are not part of the house motif to be identified. 

In summary, it can be stated that the user should choose $k$ based on the concept completeness score, as it is indicative of whether all concepts are discovered. However, this should be viewed in conjunction with the visualisations by the user to avoid redundancy in the concepts. Lastly, it can be argued that choosing an appropriate $n$ is most important, as it significantly impacts the complexity and consequently the understanding of the concept. As the user must interpret the concept and may have additional knowledge on the dataset, it is vital that this parameter is chosen by the user rather than by some heuristic. However, the average purity score can guide the user's exploration.

\subsection{Evaluation of the proposed Method across different Datasets}
\label{results1}

For brevity, we limit our qualitative analysis to BA-Shapes, but present the quantitative analysis across all datasets. We refer the reader to Appendix \ref{E} for further qualitative results. 

\subsubsection{Qualitative Results}

\begin{figure}[ht]
\vskip 0.2in
\begin{center}
\centerline{\includegraphics[width=\columnwidth, trim = 0 0 0 2cm, clip]{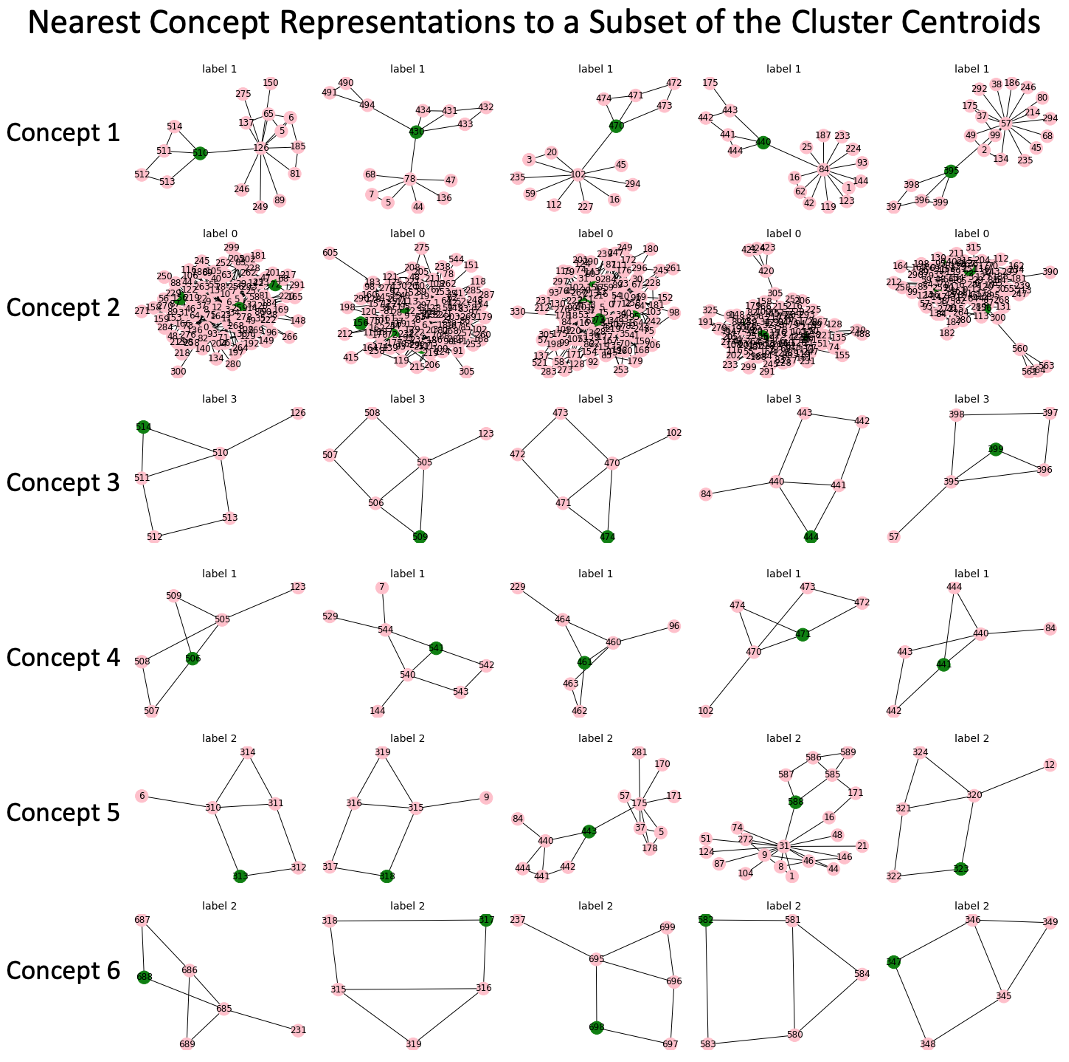}}
\caption{A subset of the concepts discovered for BA-Shapes.}
\label{ba_shapes}
\end{center}
\vskip -0.2in
\end{figure}

Figure \ref{ba_shapes} shows a subset of the concepts extracted for the BA-Shapes dataset, where the green node is the node part of the cluster and the pink nodes are the neighbourhood explored. Ten concepts are extracted in total, however, we reduce the visualisation to the most meaningful concepts. We remove noisy concepts caused by the BA base graph, such as concept 2, as these are not the motifs of interest. However, the presence of such concepts shows that the technique does not solely reason about a subset of nodes, which is beneficial for the discovery of new knowledge.

Figure \ref{ba_shapes} shows that the method successfully extracts the house motif as a concept. For example, concept 6 shows bottom nodes with their 2-hop neighbourhoods, forming the basic house structure. It highlights that the house structure is a reoccurring motif and that the structure is important for the prediction. While the additional edge displayed in the first and third structure could be seen as impurities, they are only random edges and can be disregarded. In contrast, concept 3 shows the structure with the edge, which attaches to the base graph. This shows that for the prediction of the top node, the attaching arm is of importance. When analysing concept 1 in conjunction with concept 4 it becomes evident that the GNN differentiates between housing structures with the arm attaching to the base graph on the close and far side of the node of interest. This shows that BA-Shapes has more fine-grained concept representations than the basic house structure. The similarities between different concept representations of a cluster highlight that the common structure is important for the prediction of such nodes globally.

\begin{figure}[ht]
\vskip 0.2in
\begin{center}
\centerline{\includegraphics[width=0.4\columnwidth, trim={0 0cm 0 0.8cm},clip]{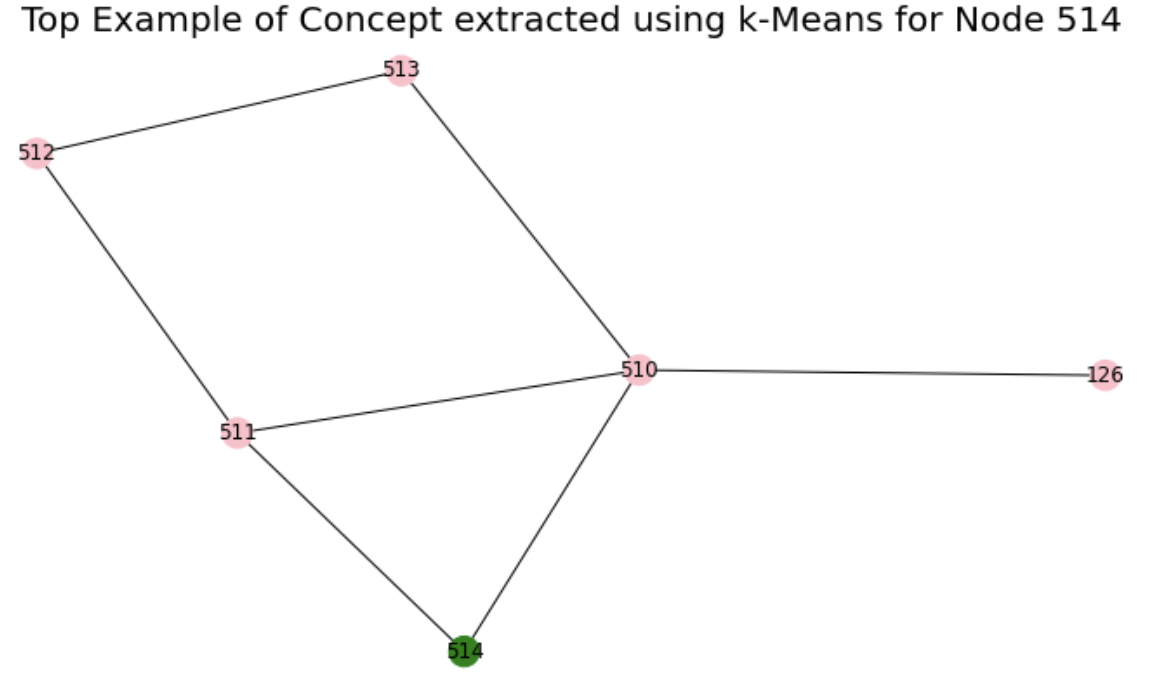}}
\caption{The concept representation representative of the cluster explaining a node in BA-Shapes.}
\label{ba_shapes_exp1}
\end{center}
\vskip -0.2in
\end{figure}

\begin{figure}[ht]
\vskip 0.2in
\begin{center}
\centerline{\includegraphics[width=0.8\columnwidth, trim={0 0 0 0.8cm},clip]{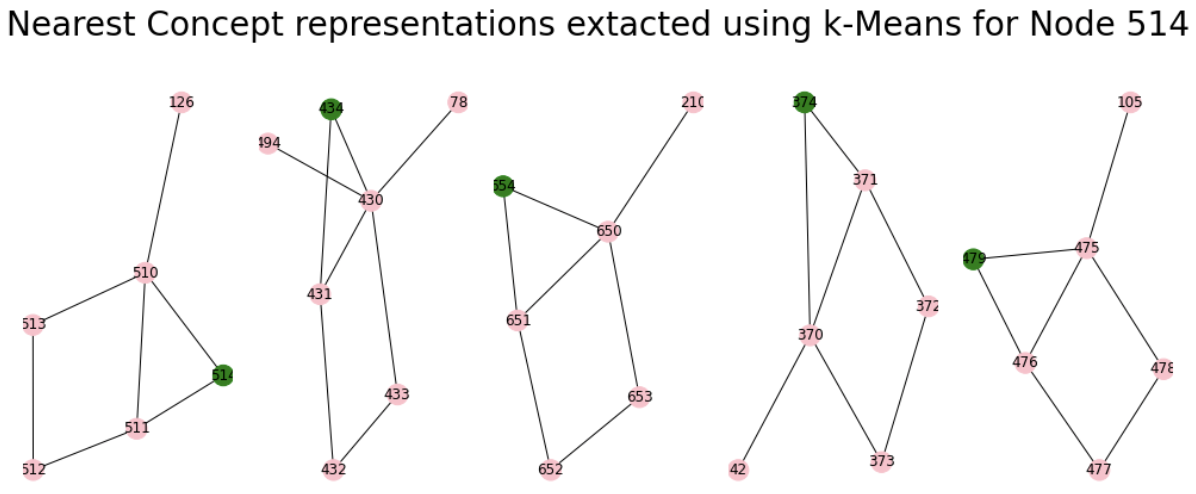}}
\caption{The concept representations nearest to the BA-Shapes node 514 explained, visualised to capture cluster variance.}
\label{ba_shapes_exp2}
\end{center}
\vskip -0.2in
\end{figure}

\begin{figure}[ht]
\vskip 0.2in
\begin{center}
\centerline{\includegraphics[width=0.6\columnwidth, trim={0 0 0 0.8cm},clip]{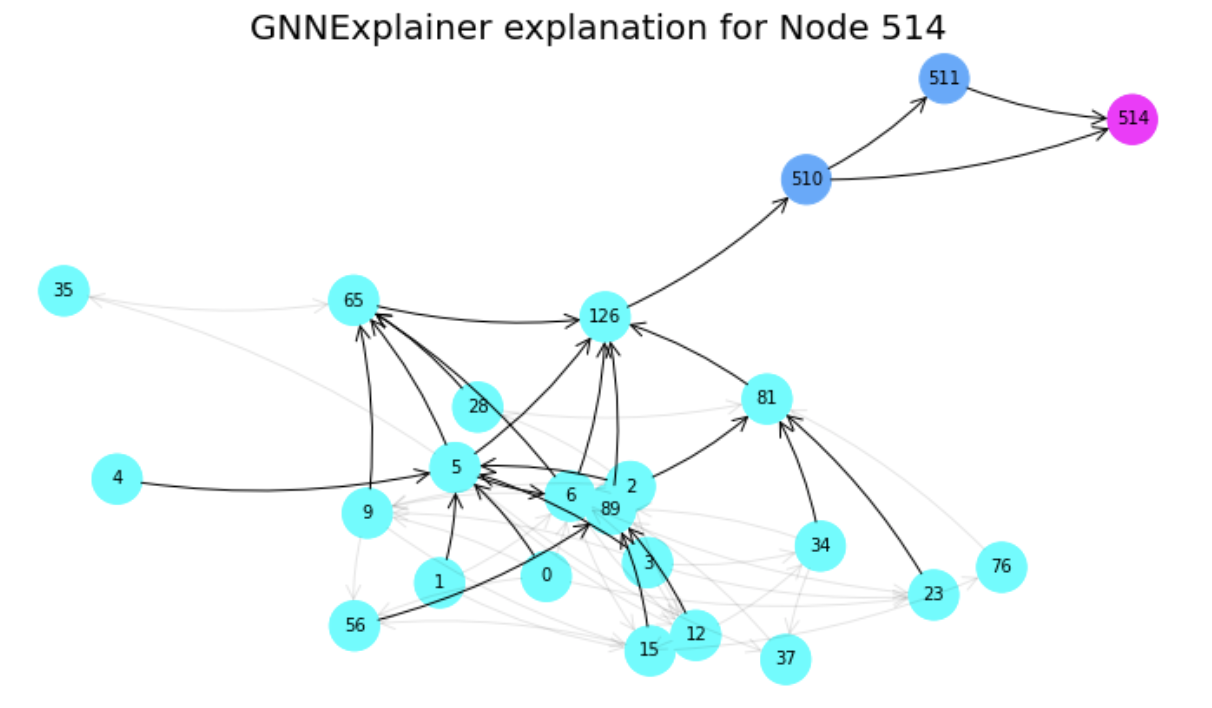}}
\caption{GNNExplainer explanation produced for BA-Shapes node 514, where the top of the house, middle of the house and BA base graph are coloured in purple, blue and turquoise, respectively.}
\label{ba_shapes_gnnexplainer}
\end{center}
\vskip -0.2in
\end{figure}

To further establish the success of our method, we compare our explanations against those of GNNExplainer \cite{Ying2019}. Figures \ref{ba_shapes_exp1} and \ref{ba_shapes_exp2} visualise the explanation extracted for a top node in the house structure. The former visualises the concept representation closest to the cluster centre, which can be seen as the global representation for the cluster. In contrast, the latter visualises the concept representations nearest to the node being explained to capture inter-cluster variance if $k$ is not chosen appropriately. It should be noted that in Figure \ref{ba_shapes_exp1} the node explained is not included unless it is closest to the cluster centroid, while in Figure \ref{ba_shapes_exp2} the first concept representation is the concept representation for the node itself. From the visualisations, it is evident that the house structure plays an important role for the prediction of the top node, as well as the arm attaching to the base graph. In contrast, Figure \ref{ba_shapes_gnnexplainer} visualises the explanation of GNNExplainer \cite{Ying2019}. It shows that the middle nodes (blue) of the house and part of the BA base structure (turquoise) are important. The explanation is enriched by the directed edges, which show the influence of the computation. However, in contrast to the concept-based explanation, the explanation is local and is not as intuitive to understand without specific knowledge about the computation of GNNs.

\subsection{Quantitative Results}

\subsubsection{Concept Completeness}

To evaluate whether the set of concepts discovered is sufficient for explanation, we compute the concept completeness score, summarised in Table \ref{completeness_score_decision_tree}. The completeness score computed ranges at the accuracy of the GNN for the BA-Shapes and Tree-Cycles dataset. This indicates that the set of concepts discovered represents the reasoning of the GNN well. However, a significantly lower concept score is obtained for BA-Community, which can be attributed to the BA-Graph dominating the concepts discovered. In total, the two BA base graphs have 600 nodes which make up two classes, while there are only 80 house structures of five nodes per community that make up six classes. This distribution causes concepts related to the BA base graph to dominate the set, impacting the completeness score.

Similar to the node classification tasks, the concept completeness score for REDDIT-BINARY ranges around the classifier accuracy. However, the completeness score for Mutagenicity is significantly lower. This means that not all concepts describing the dataset are discovered and extracted, indicating that $k$ must be increased. Nevertheless, the concepts allow to describe 71.3\% of the dataset.

\begin{table}[t]
\caption{The completeness score for the discovered set of concepts and model accuracy for each dataset.}
\label{completeness_score_decision_tree}
\vskip 0.15in
\begin{center}
\begin{small}
\begin{sc}
{%
\begin{tabular}{|l|c|c|}
\hline
\multicolumn{1}{|c|}{{\textbf{Dataset}}} & \textbf{\begin{tabular}[c]{@{}c@{}}Completeness \\ Score\end{tabular}} & \textbf{\begin{tabular}[c]{@{}c@{}}Model\\ Accuracy\end{tabular}} \\ \hline
BA-Shapes    & 0.964   & 0.956\\ \hline
BA-Community & 0.678 & 0.990 \\ \hline
BA-Grid      & 1.000 & 0.952 \\ \hline
Tree-Cycles   & 0.949 & 0.955\\ \hline
Tree-Grid    & 0.965 & 0.957 \\ \hline
Mutagenicity & 0.713 & 0.869 \\ \hline
REDDIT-BINARY & 0.967  & 0.899\\ \hline
\end{tabular}%
}
\end{sc}
\end{small}
\end{center}
\vskip -0.1in
\end{table}

\subsubsection{Concept Purity}

To evaluate the quality of the individual concepts, we compute the concept purity score, summarised in Table \ref{purity_score}. In general, the concepts discovered for the node classification datasets are pure, as the average concept purity scores are reasonably close to 0. Moreover, reviewing the range of the concept purity scores, at least one concept is perfectly pure. The less pure concepts discovered can be explained through the structure of the graph. For example, one of the concepts discovered for the BA-Community dataset has a concept purity score of 11, which is very high when seen in relation to the concept size. However, this can be explained by the concept being representative of nodes in the BA base graph, which is highly connected and produces noisy concept. It is important to analyse the concept purity in conjunction with the visualisations and information on the dataset to filter out less informative concepts.

Comparing the results for node and graph classification datasets, the method performs less well on the graph classification datasets. This is due to the graph classification datasets being real-world datasets that are less structured than the synthetic node classification datasets, which have a clear motif encoded. However, the poorer concept purity can also be attributed to $k$ and $n$ not being defined as well for these datasets, which again can be attributed to the imperfect knowledge of the datasets. The results could potentially be improved by further fine tuning these parameters. Nevertheless, the concept purity scores for the graph classification datasets are still within a reasonable range and perfectly pure concepts are discovered.

\begin{table}[t]
\caption{The minimum, maximum and average purity score of the concepts for each dataset.}
\label{purity_score}
\vskip 0.15in
\begin{center}
\begin{small}
\begin{sc}
{%
\begin{tabular}{|l|c|c|c|}
\hline
\multicolumn{1}{|c|}{{\multirow{2}{*}{\textbf{Dataset}}}} &
  \multicolumn{3}{|c|}{\textbf{Concept Purity}} \\ \cline{2-4} 
  \multicolumn{1}{|l|}{} &
  \textbf{Min} &
  \textbf{Max} &
  \textbf{Average} \\ \hline
BA-Shapes    & 0.000 & 10.000  & 3.375   \\ \hline
BA-Community & 1.000 & 11.000 &  4.923     \\ \hline
BA-Grid      & 0.000 & 0.000   & 0.000    \\ \hline
Tree-Cycles   & 0.000 & 4.000   & 1.167 \\ \hline
Tree-Grid    & 0.000 & 8.500 & 3.100   \\ \hline
Mutagenicity & 0.000 & 14.500 & 6.968       \\ \hline
REDDIT-BINARY & 0.000 & 14.000 & 5.400       \\ \hline
\end{tabular}
}
\end{sc}
\end{small}
\end{center}
\vskip -0.1in
\end{table}

\section{Conclusion}
\label{conclusion}
We present a post-hoc, unsupervised concept representation learning method for the discovery and extraction of concept-based explanations for GNNs. We successfully demonstrate that the proposed method allows to extract high quality concepts that are semantically meaningful and can be reasoned about by the user. This research can help increase the transparency and accountability of GNNs, as a user is be able to analyse predictions using global explanations. Furthermore, by visualising concept representations close to the input instance being explained, more localised explanations can be retrieved. The concept-based explanations are more intuitive to understand than the predictions of GNNExplainer~\cite{Ying2019}, because important structural and feature information is easier to understand when viewed across different examples. GCExplainer has a vast potential for application, including but not limited to explaining knowledge graphs and social networks.

An area for future work is the application of the method on link prediction tasks and multi-modal graph datasets. In particular, the use of MultiMap~\cite{Jain2021} as a DR technique could be explored to allow combining data in different formats from different datasets.  Future work could also investigate Deep Graph Mapper \cite{Bodnar2021} for visualisation or other modes of concept representation, as subgraphs can be difficult to reason about. Lastly, an alternative avenue for future work is the adaption of the work of ~\cite{Yeh2019}, such that the regulariser includes the graph edit distance to encourage the grouping of similar subgraph instances.

\bibliography{main}
\bibliographystyle{icml2021}


\appendix

\section{Concept Heuristics}
\label{A}

We define a set of fine-grained concepts for BA-Shapes, which we expect to recover. We use this as a heuristic to evaluate how well the concept discovery is performed. For example, a heuristic concept for BA-Shapes is a house structure grouped via the bottom node on the far side of the house, where the far side is the side opposite of the edge attaching to the main graph. We count the number of these concepts extracted for a succinct evaluation. 

The edge attaching the house to the main graph allows to distinguish between concepts in a fine-grained manner. This edge allows to differentiate middle and bottom nodes as nodes on the close and far side of the base graph. We use this to define the following set of concept heuristics for BA-Shapes:
\begin{enumerate}
    \item House with top-node
    \item House with middle-node
    \item House with bottom-node
    \item House with top-node with edge
    \item House with middle-node and edge on far side
    \item House with middle-node with edge on close side
    \item House with bottom-node with edge on far side
    \item House with bottom node with edge on close side
\end{enumerate}

This evaluation is only applicable to the BA-House and BA-Community dataset, due to the unique house structure. It does not work for the other node classification datasets, as the structures are invariant to translation. Furthermore, it does not apply to the graph classification tasks, as the ground truth motifs are too varied.

We perform the evaluation of the concept heuristics discovered on the methodology proposed, as well as variations of the methodology to establish the validity of the design. The variations of the methodology substitute the raw activation space for a Principle Component Analysis (PCA) \cite{Smith2002}, t-distributed Stochastic Neighbour Embedding (t-SNE) \cite{VanDerMaaten2008} and Uniform Manifold Approximation and Projection (UMAP) \cite{McInnes2018} reduced activation space, while $k$-Means clustering is substituted by agglomerative hierarchical clustering (AHC) \cite{Rokach2009} and Density-Based Spatial Clustering of Applications with Noise (DBSCAN) \cite{Ester1996}. 

\subsection{Results}

For a succinct comparison of the dimensionality reduction and clustering techniques explored for concept discovery, we visualise the concepts in the same manner as before and count the number of concept heuristics retrieved. 

Table \ref{samples_kmeans} summarises the number of such concept heuristics retrieved using the different dimensionality reduction techniques and clustering algorithms across the layers of the model. It appears as if applying t-SNE dimensionality reduction on the activation space and then performing $k$-Means clustering yields the best results in comparison to the other constellations, as consistently more concepts are identified. However, as five of the eight concepts are identified both via layer 0, 1, and 2, it can be argued that our assumption that improved clustering is produced in higher layers is refuted. Nevertheless, the assumption holds when applying the other dimensionality reduction methods.

\begin{figure}[ht]
\vskip 0.2in
\begin{center}
\centerline{\includegraphics[width=\columnwidth, trim={0 0 0 0.1cm},clip]{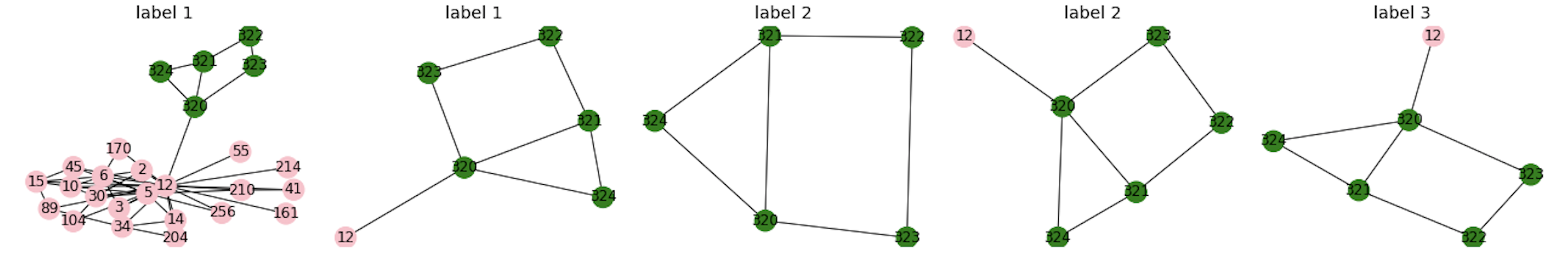}}
\caption{A single house instance discovered using AHC on the raw activation space of the final convolutional layer in BA-Shapes.}
\label{whole_house}
\end{center}
\vskip -0.2in
\end{figure}

\begin{figure}[ht]
\vskip 0.2in
\begin{center}
\centerline{\includegraphics[width=\columnwidth, trim={0 0 0 0.1cm},clip]{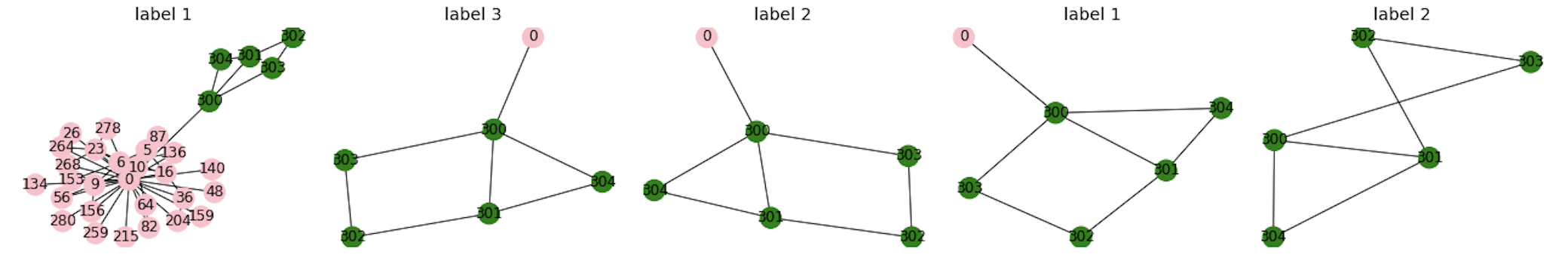}}
\caption{A single house instance discovered using DBSCAN on the raw activation space of the final convolutional layer in BA-Shapes.}
\label{whole_house2}
\end{center}
\vskip -0.2in
\end{figure}

The number of these fine-grained concepts found using AHC and DBSCAN is consistently 0. This is because both clustering algorithms discover concepts, which cluster nodes of the same house structure rather than nodes of the same type, as shown in Figures \ref{whole_house} and \ref{whole_house2} for AHC and DBSCAN, respectively. These concepts do not fit our defined schemata. Furthermore, these concepts are not useful, as they are local explanations. This highlights that the use of $k$-Means is desirable, as it allows to extract concepts based on node similarity, which allows producing global explanations. This can be explained by $k$-Means minimising inter-cluster variance. However, it should be noted that a different number of clusters is defined for each algorithm, which impacts the results.

\begin{table}[t]
\caption{Table summarising the number of concepts identified per constellation of dimensionality reduction and clustering technique for BA-Shapes. The highest number of concept heuristics across the layers for a given dimensionality reduction technique and clustering algorithm are highlighted in bold.}
\label{samples_kmeans}
\vskip 0.15in
\begin{center}
\begin{small}
\begin{sc}
{%
\begin{tabular}{|c|c|c|c|c|}
\hline
\multirow{2}{*}{\textbf{\begin{tabular}[c]{@{}c@{}}DR \\ Method\end{tabular}}} &
  \multirow{2}{*}{\textbf{Layer}} &
  \multicolumn{3}{c|}{\textbf{Num. of Concepts Recovered}} \\ \cline{3-5} 
      &   & \textbf{$k$-Means} & \textbf{AHC} & \textbf{DBSCAN} \\ \hline
Raw & 0 & 3 & 0 & 0\\ \hline
Raw & 1 & 2 & 0 & 0\\ \hline
Raw & 2 & 4 & 0 & 0\\ \hline
Raw & 3 & \textbf{5} & 0 & 0\\ \hline
Raw & 4 & 3 & 0 & 0\\ \hline
PCA & 0 & 2 & 0 & 0\\ \hline
PCA & 1 & 2 & 0 & 0\\ \hline
PCA & 2 & \textbf{4} & 0 & 0\\ \hline
PCA & 3 & \textbf{4} & 0 & 0\\ \hline
PCA & 4 & 3 & 0 & 0\\ \hline
t-SNE & 0 & 5 & 0 & 0\\ \hline
t-SNE & 1 & 5 & 0 & 0 \\ \hline
t-SNE & 2 & 5 & 0 & 0 \\ \hline
t-SNE & 3 & \textbf{6} & 0 & 0 \\ \hline
t-SNE & 4 & 4  & 0 & 0 \\ \hline
UMAP & 0 & 2 & 0 & 0 \\ \hline
UMAP & 1 & 4 & 0 & 0 \\ \hline
UMAP & 2  & \textbf{6} & 0 & 0\\ \hline
UMAP & 3  & \textbf{6} & 0 & 0 \\ \hline
UMAP & 4 & 4 & 0 & 0 \\ \hline
\end{tabular}%
}
\end{sc}
\end{small}
\end{center}
\vskip -0.1in
\end{table}

\section{Evaluation of the proposed Method against alternative Designs}
\label{B}

\subsection{Experimental Setup}
We benchmark the method proposed against alternative designs in order to validate our design choices. We take a systematic approach to this by varying different elements of the proposed design, as well as the dataset used:
\begin{enumerate}
    \item \textbf{Choice of DR:} We substitute the raw activation space for a PCA, t-SNE and UMAP reduced version.
    \item \textbf{Choice of layer:} We perform the concept discovery and extraction on all layers of the models trained.
    \item \textbf{Choice of clustering algorithm:} We substitute the $k$-Means algorithm with AHC and DBSCAN.
    \item \textbf{Choice of dataset:} We perform these experiments on a set of node and graph classification datasets.
\end{enumerate}

\subsection{Qualitative Results}

We visualise the activation space of each layer, similar to the analysis performed in \cite{Kazhdan2020} to confirm the clustering of similar node instances in GNNs. We limit our discussion to the analysis of the final convolutional layers in the model for brevity. 

\begin{figure}[ht]
\vskip 0.2in
\begin{center}
\centerline{\includegraphics[width=\columnwidth, trim={0 0 0 0cm},clip]{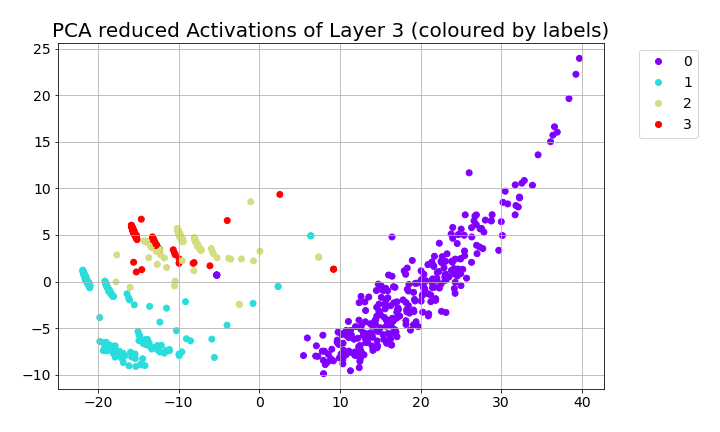}}
\caption{PCA reduced activation space of the final convolutional layer of the model trained on BA-Shapes.}
\label{pca2}
\end{center}
\vskip -0.2in
\end{figure}

\begin{figure}[ht]
\vskip 0.2in
\begin{center}
\centerline{\includegraphics[width=\columnwidth, trim={0 0 0 0cm},clip]{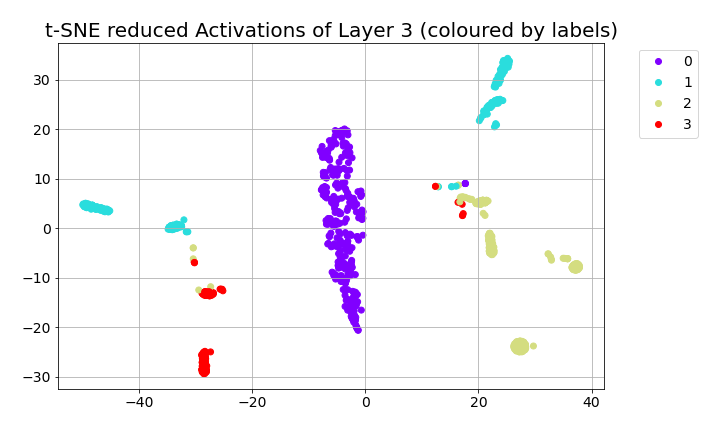}}
\caption{t-SNE reduced activation space of the final convolutional layer of the model trained on BA-Shapes.}
\label{tsne2}
\end{center}
\vskip -0.2in
\end{figure}

\begin{figure}[ht]
\vskip 0.2in
\begin{center}
\centerline{\includegraphics[width=\columnwidth, trim={0 0 0 0cm},clip]{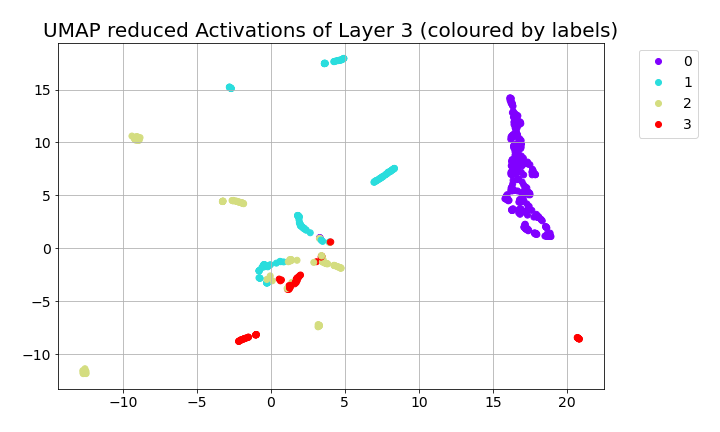}}
\caption{UMAP reduced activation space of the final convolutional layer of the model trained on BA-Shapes.}
\label{umap2}
\end{center}
\vskip -0.2in
\end{figure}

Figures \ref{pca2} - \ref{umap2} visualises the activation space of the final convolutional layer of the model trained on BA-Shapes using different dimensionality reduction techniques. The points plotted are the different input nodes coloured by their class label: where classes 0, 1, 2 and 3 represent the nodes not part of the house structure, top nodes, middle nodes and bottom nodes, respectively. Figures \ref{pca2} and \ref{tsne2} show the PCA and t-SNE reduced activation space of the final convolutional layer, respectively. It can be stated that more distinct clustering is achieved using t-SNE. For example, in Figure \ref{tsne2}, ten distinct clusters can be identified, however, four of these clusters group instances of different classes together. In comparison, Figure \ref{umap2} shows that UMAP achieves even stronger clustering, however, some similarities to t-SNE can be identified, such as the clustering of class 0. This can be attributed to the representation found by t-SNE aligning with that found by UMAP when the attractive forces between data points are stronger than the repulsive ones in the calculation \cite{Bohm2020}. Comparing and contrasting the three techniques, it can be argued that improved clustering is achieved by UMAP and t-SNE over PCA.

\begin{figure}[ht]
\vskip 0.2in
\begin{center}
\centerline{\includegraphics[width=\columnwidth, trim={0 0 0 0cm},clip]{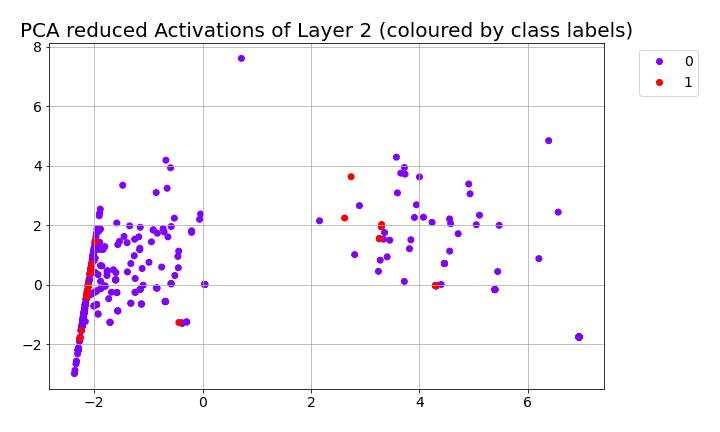}}
\caption{PCA reduced activation space of the middle convolutional layer of the model trained on the REDDIT-BINARY dataset (layers counted from 0).}
\label{pca_r2}
\end{center}
\vskip -0.2in
\end{figure}

\begin{figure}[ht]
\vskip 0.2in
\begin{center}
\centerline{\includegraphics[width=\columnwidth, trim={0 0 0 0cm},clip]{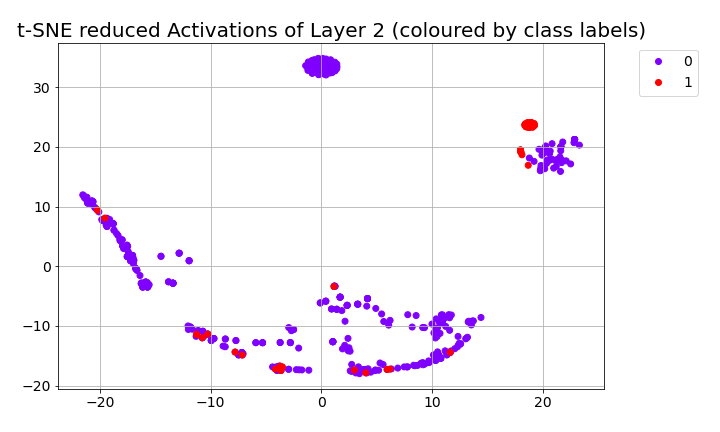}}
\caption{t-SNE reduced activation space of the middle convolutional layer of the model trained on the REDDIT-BINARY dataset (layers counted from 0).}
\label{tsne_r2}
\end{center}
\vskip -0.2in
\end{figure}

\begin{figure}[ht]
\vskip 0.2in
\begin{center}
\centerline{\includegraphics[width=\columnwidth, trim={0 0 0 0cm},clip]{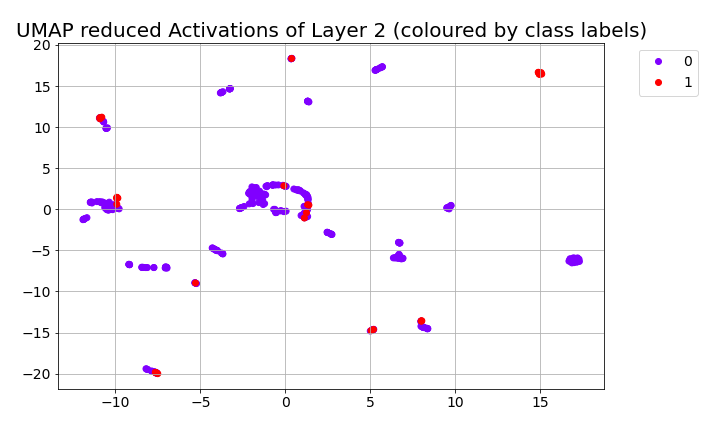}}
\caption{UMAP reduced activation space of the middle convolutional layer of the model trained on the REDDIT-BINARY dataset (layers counted from 0).}
\label{umap_r2}
\end{center}
\vskip -0.2in
\end{figure}

\begin{figure}[ht]
\vskip 0.2in
\begin{center}
\centerline{\includegraphics[width=\columnwidth, trim={0 0 0 0cm},clip]{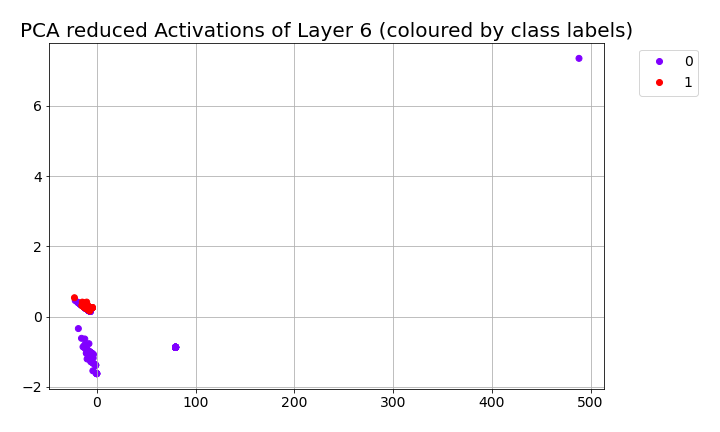}}
\caption{PCA reduced activation space of the final convolutional layer of the model trained on the REDDIT-BINARY dataset (layers counted from 0).}
\label{pca_r6}
\end{center}
\vskip -0.2in
\end{figure}

\begin{figure}[ht]
\vskip 0.2in
\begin{center}
\centerline{\includegraphics[width=\columnwidth, trim={0 0 0 0cm},clip]{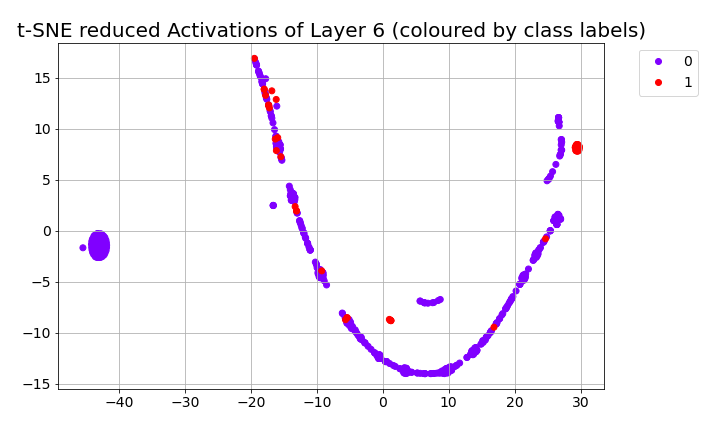}}
\caption{t-SNE reduced activation space of the final convolutional layer of the model trained on the REDDIT-BINARY dataset (layers counted from 0).}
\label{tsne_r6}
\end{center}
\vskip -0.2in
\end{figure}

\begin{figure}[ht]
\vskip 0.2in
\begin{center}
\centerline{\includegraphics[width=\columnwidth, trim={0 0 0 0cm},clip]{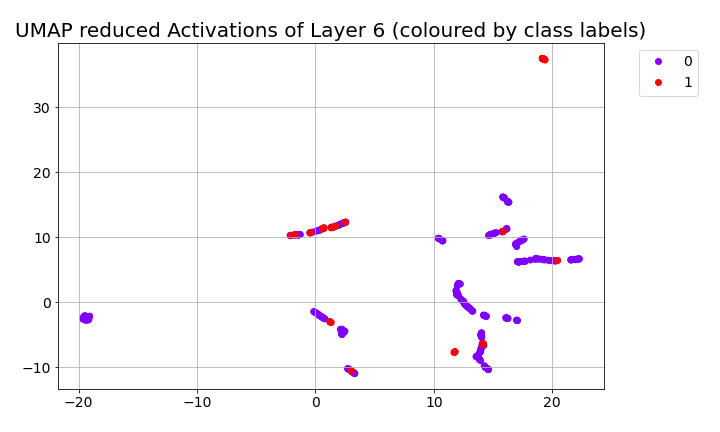}}
\caption{UMAP reduced activation space of the final convolutional layer of the model trained on the REDDIT-BINARY dataset (layers counted from 0).}
\label{umap_r6}
\end{center}
\vskip -0.2in
\end{figure}

To establish our observations, we also perform this analysis on REDDIT-BINARY, a graph classification dataset. Figures \ref{pca_r2} - \ref{umap_r6} show the dimensionality reduced activation space of a middle and the final convolutional layer for REDDIT-BINARY. In contrast to our earlier observations, the PCA reduced activation space appears more clustered for the later layers. Furthermore, the clustering does not seem to improve significantly for later layers of the t-SNE and UMAP reduced activation space. In general, the PCA reduced activation space of the final convolutional layer has the strongest clustering, with minimal overlap. However, the data points are more spread out in the UMAP reduced activation space, which may be better for concept discovery. This leads to the conclusion that the choice of dimensionality reduction technique depends on the dataset. Thus, performing clustering on the raw activation space remains the best choice. We refrain from a further analysis of the dimensionality reduced activation space of other models due to the limited insight gained.

\begin{figure}[ht]
\vskip 0.2in
\begin{center}
\centerline{\includegraphics[width=\columnwidth, trim={3cm 0 3cm 2cm},clip]{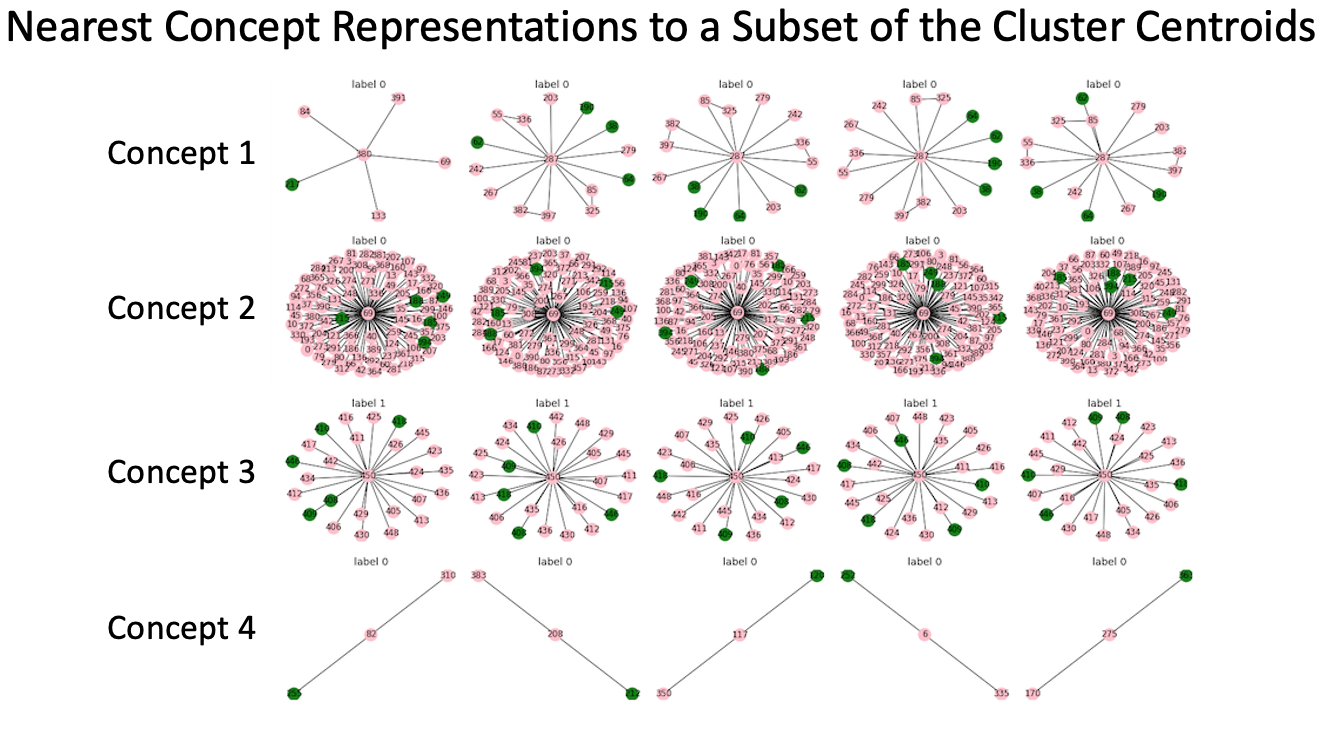}}
\caption{A subset of the concepts discovered for REDDIT-BINARY using $k$-Means on the raw activation space of the final convolutional layer.}
\label{reddit1}
\end{center}
\vskip -0.2in
\end{figure}

\begin{figure}[ht]
\vskip 0.2in
\begin{center}
\centerline{\includegraphics[width=\columnwidth, trim={3cm 0 3cm 3cm},clip]{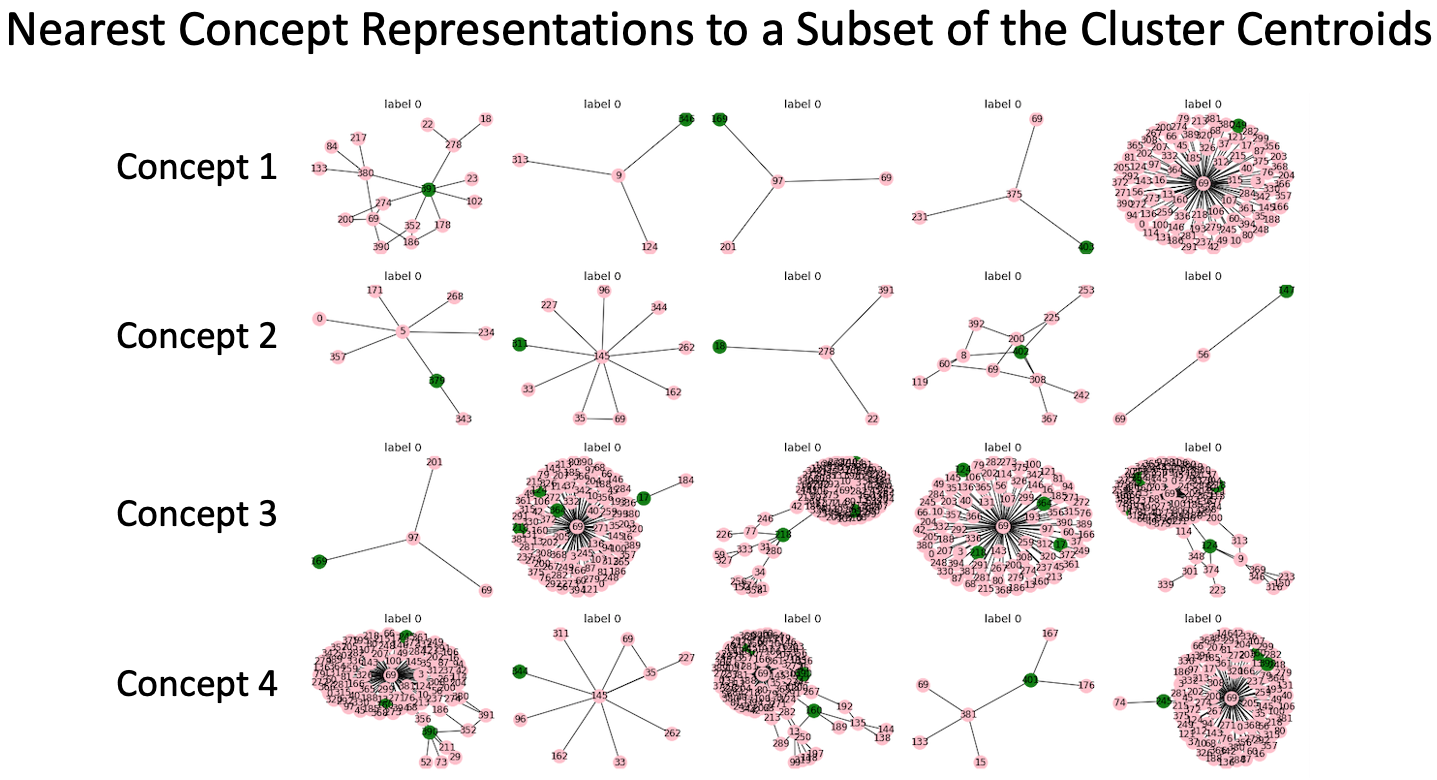}}
\caption{A subset of the concepts discovered for REDDIT-BINARY using AHC on the raw activation space of the final convolutional layer.}
\label{reddit2}
\end{center}
\vskip -0.2in
\end{figure}

\begin{figure}[ht]
\vskip 0.2in
\begin{center}
\centerline{\includegraphics[width=\columnwidth, trim={3cm 0 3cm 2.5cm},clip]{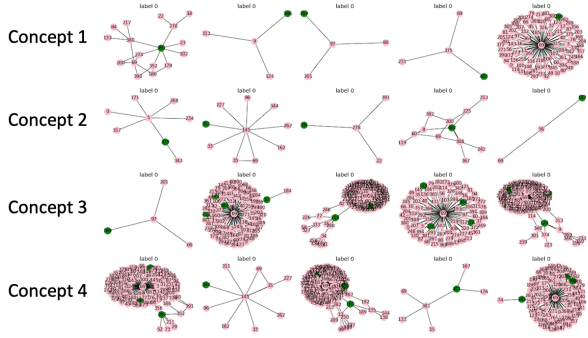}}
\caption{A subset of the concepts discovered for REDDIT-BINARY using DBSCAN on the raw activation space of the final convolutional layer.}
\label{reddit3}
\end{center}
\vskip -0.2in
\end{figure}

We substantiate our observations by examining the concepts extracted for REDDIT-BINARY. Figures \ref{reddit1} - \ref{reddit3} show the concepts discovered using $k$-Means, AHC and DBSCAN on the raw activation space of the final convolutional layer. While Figure \ref{reddit1} shows that a set of coherent, global concepts can be discovered using $k$-Means, Figures \ref{reddit2} and \ref{reddit3} show a failure to produce such using AHC and DBSCAN. Specifically, AHC only produces concepts representative of class 0 and the nodes grouped together do not exhibit the same neighbourhood. This highlights that $k$-Means achieves a more appropriate clustering of the activation space for the discovery of global concepts.

The qualitative results are indicative of the proposed method outperforming the alternative designs. However, the results are not conclusive and copious. Therefore, we decide to focus on the quantitative evaluation of the proposed method against alternative designs.

\subsection{Quantitative Results}
\label{quant}

The quantitative results in form of the completeness and purity score provide the greatest indication of the performance of the alternative designs. Hence, we perform an analysis of the completeness and purity score for a larger selection of the datasets. As the base motifs encoded in the node classification are similar, we limit our analysis to the exploration of the BA-Shapes, BA-Grid and Tree-Cycles dataset. To investigate performance of the method on graph classification, we explore the results for the REDDIT-BINARY dataset.

\subsubsection{Concept Completeness}

\textbf{BA-Shapes:} Table \ref{kmeanscompleteness} summarise the completeness scores of the concepts discovered using the different combinations of clustering algorithms and DR techniques on the BA-Shapes dataset. Our general assumption that improved concepts are extracted in deeper layers of the model is confirmed. Across the DR methods employed, the highest concept score is computed using the concepts discovered in the final convolutional layer. However, when clustering the raw activation space, the highest completeness score is obtained when recovering concepts from the last convolutional or linear layer. In contrast, the completeness score decreases for the liner layer across the DR techniques. The score obtained using the proposed method is the highest completeness score overall at 0.964, exceeding the testing accuracy of the model. This indicates that a complete set of concepts is discovered. The second best method appears to be the clustering of the PCA reduced activation space, achieving an a completeness score of 0.949. This stands in contrast to our previous observation of t-SNE yielding improved results and can be explained by the relation between PCA and $k$-Means \cite{Ding2004}.

Similar results are obtained using AHC for concept discovery and extraction. Contrary to the results obtained for $k$-Means clustering, the highest completeness score is obtained when performing clustering on the PCA-reduced activation space of the linear layer. In contrast, the highest concept completeness scores are achieved on the final convolutional layer when the activation space is reduced using t-SNE and UMAP. Across both clustering techniques, clustering on the t-SNE reduced activation space appears to produce the poorest set of concepts.

\begin{table}[t]
\caption{Completeness score for the set of concept discovered for the BA-Shapes dataset using different clustering and DR techniques.}
\label{kmeanscompleteness}
\vskip 0.15in
\begin{center}
\begin{small}
\begin{sc}
{%
\begin{tabular}{|c|c|c|c|c|}
\hline
\multirow{2}{*}{\textbf{\begin{tabular}[c]{@{}c@{}}DR \\ Method\end{tabular}}} &
  \multirow{2}{*}{\textbf{Layer}} &
  \multicolumn{3}{c|}{\textbf{Completeness Score}} \\ \cline{3-5} 
      &   & \textbf{$k$-Means} & \textbf{AHC} & \textbf{DBSCAN} \\ \hline
Raw           & 0              & 0.612 & 0.431 & 0.431 \\ \hline
Raw           & 1              & 0.752 & 0.518 & 0.620 \\ \hline
Raw           & 2              & \textbf{0.964} & 0.810 & \textbf{0.920} \\ \hline
Raw           & 3              & \textbf{0.964}  & \textbf{0.949} & 0.898 \\ \hline
PCA           & 0              & 0.701 & 0.701 & 0.679                   \\ \hline
PCA           & 1              & 0.766 & 0.801 & 0.774                  \\ \hline
PCA           & 2              & 0.869 & 0.949 & \textbf{0.898}                   \\ \hline
PCA           & 3              & \textbf{0.949}  & \textbf{0.956} & 0.883                  \\ \hline
t-SNE         & 0              & 0.613 & 0.613 & 0.431                    \\ \hline
t-SNE         & 1              & 0.752 & 0.752 & 0.620                 \\ \hline
t-SNE         & 2              & \textbf{0.810}  & \textbf{0.810} & \textbf{0.934}                 \\ \hline
t-SNE         & 3              & \textbf{0.810} & 0.796 & 0.898    \\ \hline
UMAP          & 0              & 0.701 & 0.701 & 0.701                   \\ \hline
UMAP          & 1              & 0.788 & 0.788 & 0.803                 \\ \hline
UMAP          & 2              & \textbf{0.934} & \textbf{0.934} & 0.949             \\ \hline
UMAP          & 3              & 0.796 & 0.781  & \textbf{0.964}                  \\ \hline
\end{tabular}
}
\end{sc}
\end{small}
\end{center}
\vskip -0.1in
\end{table}

In reference to the results for DBSCAN, it can be stated that the final convolutional layer or the linear layer of the model are the best for concept discovery and extraction. Contrary to the previous results, the highest completeness score is obtained on the UMAP-reduced activation space. Based on these results, it can be established that no DR technique works the best across the clustering methods. Moreover, the observation of improved concept representations in deeper layers is confirmed. In general, $k$-Means clustering on the raw activation space achieved the highest concept completeness score, confirming our design choices. However, it should be noted that the different clustering techniques form a different number of clusters.

\begin{figure}[ht]
\vskip 0.2in
\begin{center}
\centerline{\includegraphics[width=\columnwidth, trim={0cm 0 0cm 0.2cm},clip]{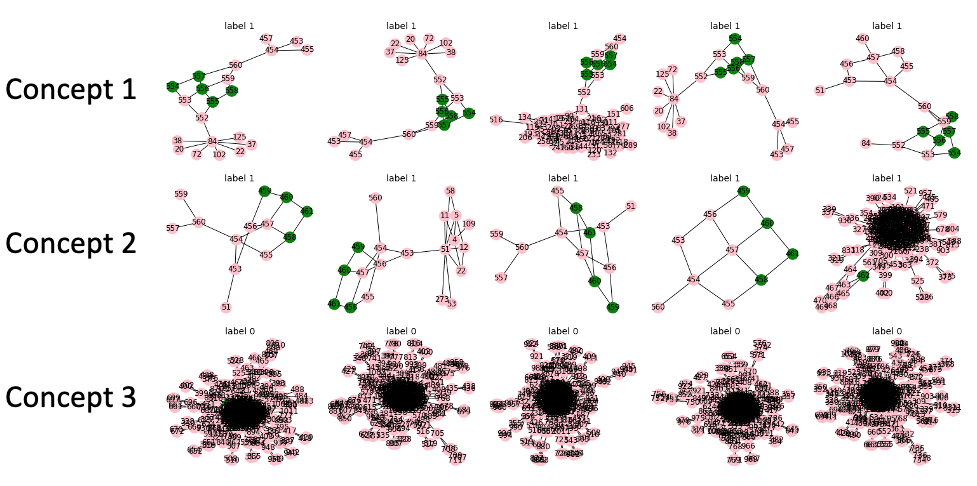}}
\caption{A subset of the concepts discovered for BA-Grid using AHC on the raw activation space of the final convolutional layer.}
\label{visual_ba_grid}
\end{center}
\vskip -0.2in
\end{figure}

\textbf{BA-Grid:} Table \ref{ba_grid_completeness} summarises the completeness scores computed across the different definitions of the base method applied to the BA-Grid dataset. In general, our previous observations are supported. For example, across the dimensionality reduction techniques and clustering algorithms, it can be observed that improved completeness scores are obtained in later layers of the model. However, AHC appears to outperform $k$-Means in concept discovery, achieving a completeness score of 1 across the different dimensionality reduction techniques. However, these results are not supported by the visual representation of the concepts. Figure \ref{visual_ba_grid} shows the concepts discovered using AHC on the raw activation space of the final convolutional layer. The concepts do not appear pure. In contrast to the concepts produced using $k$-Means, these concepts are less easy to reason about. 

\begin{table}[t]
\caption{Completeness score for the set of concept discovered for the BA-Grid dataset using different clustering and DR techniques.}
\label{ba_grid_completeness}
\vskip 0.15in
\begin{center}
\begin{small}
\begin{sc}
{%
\begin{tabular}{|c|c|c|c|c|}
\hline
\multirow{2}{*}{\textbf{\begin{tabular}[c]{@{}c@{}}DR \\ Method\end{tabular}}} &
  \multirow{2}{*}{\textbf{Layer}} &
  \multicolumn{3}{c|}{\textbf{Completeness Score}} \\ \cline{3-5} 
      &   & \textbf{$k$-Means} & \textbf{AHC} & \textbf{DBSCAN} \\ \hline
Raw           & 0              & 0.810 & 1.000 & 0.746 \\ \hline
Raw           & 1              & 0.834 & 0.829 & 0.751 \\ \hline
Raw           & 2              & 0.985 & 0.980 & 0.800 \\ \hline
Raw           & 3              & \textbf{1.000}  & \textbf{1.000} & 0.917 \\ \hline
Raw           & 4              & 0.990  & 0.995 &  \textbf{0.990}\\ \hline
PCA           & 0              & 0.990 & 0.746 & 0.746                    \\ \hline
PCA           & 1              & 0.927 & 0.829 & 0.751                \\ \hline
PCA           & 2              & 0.980 & 0.980 & 0.820                   \\ \hline
PCA           & 3              & \textbf{1.000}  & \textbf{1.000} & 0.921                  \\ \hline
PCA           & 4              & 0.995  & 0.995 & \textbf{0.990}                  \\ \hline
t-SNE         & 0              & 0.810 & 0.990 & \textbf{0.990}                    \\ \hline
t-SNE         & 1              & 0.834 & 0.951 & 0.951                \\ \hline
t-SNE         & 2              & 0.985  & 0.990 & 0.961                 \\ \hline
t-SNE         & 3              & \textbf{1.000} & \textbf{1.000}  & 0.761    \\ \hline
t-SNE         & 4              & 0.990 & 0.995 & 0.824    \\ \hline
UMAP          & 0              & 0.990 & 0.990 & 0.995                   \\ \hline
UMAP          & 1              & 0.956 & 0.956 & 0.951                 \\ \hline
UMAP          & 2              & 0.980 & 0.995 & 0.995             \\ \hline
UMAP          & 3              & 0.917 & 1.000   & \textbf{1.000}                  \\ \hline
UMAP          & 4              & \textbf{0.990} & \textbf{1.000}  & 0.995                   \\ \hline
\end{tabular}%
}
\end{sc}
\end{small}
\end{center}
\vskip -0.1in
\end{table}

\textbf{Tree-Cycles:} Table \ref{tree_cycle_completeness} summarises the completeness scores for the Tree-Cycles dataset. Reviewing the completeness scores obtained across the different techniques, it can be stated that the observation that improved results are obtained for later layers in the model is supported. Furthermore, the $k$-Means algorithm performs well, specifically on the raw activation space. While the AHC clustering algorithm appears to outperform $k$-Means again, this is refuted when examining the concept representations. Thus, our previous conclusions are supported.

\begin{table}[t]
\caption{Completeness scores for the set of concepts discovered for Tree-Cycles.}
\label{tree_cycle_completeness}
\vskip 0.15in
\begin{center}
\begin{small}
\begin{sc}
{%
\begin{tabular}{|c|c|c|c|c|}
\hline
\multirow{2}{*}{\textbf{\begin{tabular}[c]{@{}c@{}}DR \\ Method\end{tabular}}} &
  \multirow{2}{*}{\textbf{Layer}} &
  \multicolumn{3}{c|}{\textbf{Completeness Score}} \\ \cline{3-5} 
      &   & \textbf{$k$-Means} & \textbf{AHC} & \textbf{DBSCAN} \\ \hline
Raw           & 0              & 0.841 & 0.932 & 0.847 \\ \hline
Raw           & 1              & 0.807 & 0.926 & 0.903 \\ \hline
Raw           & 2              & \textbf{0.949} & 0.949 & 0.898 \\ \hline
Raw           & 3              & 0.938  & \textbf{0.949} & \textbf{0.909} \\ \hline
PCA           & 0              & 0.841 & 0.932 & 0.847                   \\ \hline
PCA           & 1              & 0.807 & 0.858 & 0.892                  \\ \hline
PCA           & 2              & 0.938 & \textbf{0.960} & 0.892                   \\ \hline
PCA           & 3              & \textbf{0.938}  & 0.949 & \textbf{0.909}                  \\ \hline
t-SNE         & 0              & 0.915 & \textbf{0.966} & \textbf{0.767}                    \\ \hline
t-SNE         & 1              & 0.869 & 0.886 & 0.739                 \\ \hline
t-SNE         & 2              & 0.903 & 0.949 & 0.642                 \\ \hline
t-SNE         & 3              & \textbf{0.949} & 0.949 & 0.682    \\ \hline
UMAP          & 0              & \textbf{0.875} & \textbf{0.983} & 0.727                   \\ \hline
UMAP          & 1              & 0.733 & 0.733 & \textbf{0.864}                 \\ \hline
UMAP          & 2              & 0.858 & 0.920 & 0.818             \\ \hline
UMAP          & 3              & 0.813 & 0.881  & 0.841                  \\ \hline
\end{tabular}%
}
\end{sc}
\end{small}
\end{center}
\vskip -0.1in
\end{table}

\begin{figure}[ht]
\vskip 0.2in
\begin{center}
\centerline{\includegraphics[width=\columnwidth, trim={0cm 0 0cm 0.2cm},clip]{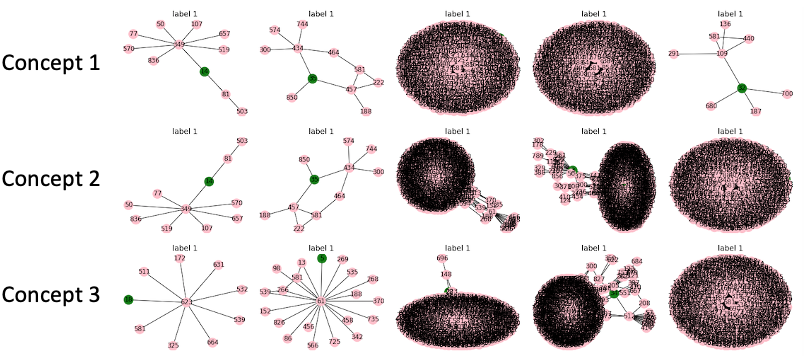}}
\caption{A subset of the concepts discovered for REDDIT-BINARY using AHC on the raw activation space of the penultimate pooling layer.}
\label{reddit_example}
\end{center}
\vskip -0.2in
\end{figure}

\textbf{REDDIT-BINARY:} Table \ref{reddit_completeness} summarises the completeness scores obtained on the REDDIT-BINARY dataset. To evaluate whether performing concept discovery is preferable on the neighbourhood aggregation layer instead of the pooling layer of the model, we evaluate the activation space of the convolutional and pooling layers. In Table \ref{reddit_completeness}, all even layers are convolutional layers, while the odd layers are pooling layers. In general, it appears as if a more complete set of concepts is extracted on the pooling layers. However, presenting the user with a set of full graphs from the dataset does not easily allow to understand the concept, wherefore, we argue clustering on the activation space is more suitable. The $k$-Means technique performs the best on the raw activation space with a completeness score of 0.969. This combination is only outperformed by AHC clustering on the raw activation space, however, this technique does not allow a clean grouping of nodes in the activation space. Rather impure concept representations are discovered, as shown in Figure \ref{reddit_example}. Thus, $k$-Means is preferable, as a clustering algorithm as it discovers purer concept representations that can easily be reasoned about by a user.

\begin{table}[t]
\caption{Completeness scores for the set of concepts discovered for REDDIT-BINARY.}
\label{reddit_completeness}
\vskip 0.15in
\begin{center}
\begin{small}
\begin{sc}
{%
\begin{tabular}{|c|c|c|c|c|}
\hline
\multirow{2}{*}{\textbf{\begin{tabular}[c]{@{}c@{}}DR \\ Method\end{tabular}}} &
  \multirow{2}{*}{\textbf{Layer}} &
  \multicolumn{3}{c|}{\textbf{Completeness Score}} \\ \cline{3-5} 
      &   & \textbf{$k$-Means} & \textbf{AHC} & \textbf{DBSCAN} \\ \hline
Raw           & 0              & 0.924 & 0.862 & \textbf{0.952} \\ \hline
Raw           & 1              & 0.375 & 0.429 & 0.500 \\ \hline
Raw           & 2              & \textbf{0.969} & 0.936 & 0.946 \\ \hline
Raw           & 3              & 0.429  & 0.500 & 0.400 \\ \hline
Raw           & 4              & 0.949  & 0.914 & 0.923 \\ \hline
Raw           & 5              & 0.429  & \textbf{1.000} & 0.333 \\ \hline
Raw           & 6              & 0.952  & 0.964 & 0.920 \\ \hline
Raw           & 7              & 0.500  & 0.500 & 0.800 \\ \hline
PCA           & 0              & 0.571 & 0.182 & 0.533                   \\ \hline
PCA           & 1              & 0.333 & 0.333 & 0.333                  \\ \hline
PCA           & 2              & 0.286 & 0.167 & 0.400                   \\ \hline
PCA           & 3              & 0.429  & 0.800 & 0.571                  \\ \hline
PCA           & 4              & 0.556  & 0.286 & \textbf{0.571}                  \\ \hline
PCA           & 5              & \textbf{0.889}  & \textbf{0.900} & 0.500                  \\ \hline
PCA           & 6              & 0.571  & 0.750 & 0.444                  \\ \hline
PCA           & 7              & 0.571  & 0.625 & 0.538                  \\ \hline
t-SNE         & 0              & 0.571 & 0.333 & 0.000                    \\ \hline
t-SNE         & 1              & 0.308 & 0.500 & 0.556                 \\ \hline
t-SNE         & 2              & 0.333 & 0.333 & 0.571                 \\ \hline
t-SNE         & 3              & \textbf{0.800} & 0.462 & 0.375    \\ \hline
t-SNE         & 4              & 0.385 & 0.545 & 0.375    \\ \hline
t-SNE         & 5              & 0.375 & \textbf{0.800} & 0.556    \\ \hline
t-SNE         & 6              & 0.444 & 0.333 & 0.667    \\ \hline
t-SNE         & 7              & 0.667 & 0.700 & \textbf{0.875}    \\ \hline
UMAP          & 0              & 0.533 & 0.100 & 0.273                   \\ \hline
UMAP          & 1              & 0.333 & 0.533 & 0.286                 \\ \hline
UMAP          & 2              & 0.209 & 0.429 & 0.667             \\ \hline
UMAP          & 3              & 0.429 & \textbf{1.000}  & 0.300                  \\ \hline
UMAP          & 4              & \textbf{0.667} & 0.500  & 0.250                  \\ \hline
UMAP          & 5              & 0.500 & 0.750  & 0.667                  \\ \hline
UMAP          & 6              & 0.417 & 0.167  & 0.500                  \\ \hline
UMAP          & 7              & 0.556 & 0.600  & \textbf{0.833}                  \\ \hline
\end{tabular}%
}
\end{sc}
\end{small}
\end{center}
\vskip -0.1in
\end{table}

\subsubsection{Concept Purity}

\textbf{BA-Shapes:} Lastly, we analyse the concept purity of the discovered concepts across the proposed method and alternative designs. Table \ref{concept_purity_all} summarises the purity of the concepts discovered using the different clustering and DR techniques on BA-Shapes. The concept purity scores across the different clustering algorithms need to be compared with caution, as a different number of clusters is used. Additionally, a different number of samples is used across the averages due to the limits imposed on the calculation of the concept purity score. 

In reference to Table \ref{concept_purity_all}, it can be stated that $k$-Means allows to extract the purest concepts, as the best purity score per DR technique is consistently lower than those for AHC and DBSCAN. Furthermore, the average purity scores for the concepts discovered for some layers using AHC and DBSCAN are significantly higher than those for $k$-Means. For example, a purity score of 0.667, 12.333 and 4.909 is achieved using $k$-Means, AHC and DBSCAN clustering on the t-SNE reduced activation space of layer 1, respectively. In general, it can be argued that $k$-Means allows to produce the purest concepts. The purity scores suggest that performing clustering on layer 1 or 2 of the raw activation space produces the best results, however, this should be viewed in conjunction with previous results due to the limitations of the purity score. Due to these limitations, we attribute higher value to the results indicated by the completeness score.

\begin{table}[t]
\caption{Average purity score for concepts extracted for the BA-Shapes dataset per layer using the different designs.}
\label{concept_purity_all}
\vskip 0.15in
\begin{center}
\begin{small}
\begin{sc}
{%
\begin{tabular}{|c|c|c|c|c|}
\hline
\multirow{2}{*}{\textbf{\begin{tabular}[c]{@{}c@{}}DR \\ Method\end{tabular}}} &
  \multirow{2}{*}{\textbf{Layer}} &
  \multicolumn{3}{c|}{\textbf{Average Concept Purity}} \\ \cline{3-5} 
      &   & \textbf{$k$-Means} & \textbf{AHC}    & \textbf{DBSCAN} \\ \hline
Raw   & 0 & 2.250            & 10.000         & -               \\ \hline
Raw   & 1 & \textbf{0.000}   & 8.000          & 2               \\ \hline
Raw   & 2 & \textbf{0.000}   & 7.667          & 5.200           \\ \hline
Raw   & 3 & 3.500            & 3.375          & 6.300           \\ \hline
Raw   & 4 & 5.375            & \textbf{1.000} & \textbf{1.333}  \\ \hline
PCA   & 0 & 0.750            & 5.000          & -               \\ \hline
PCA   & 1 & \textbf{0.000}   & 9.750          & \textbf{1.667}  \\ \hline
PCA   & 2 & 0.667            & 2.000          & 12.750          \\ \hline
PCA   & 3 & 0.667            & \textbf{1.000} & 4.000           \\ \hline
PCA   & 4 & 2.600            & 3.500          & 3.500           \\ \hline
t-SNE & 0 & 1.500            & \textbf{1.500} & 5.132           \\ \hline
t-SNE & 1 & \textbf{0.667}   & 12.333         & 4.909           \\ \hline
t-SNE & 2 & 1.417            & 7.810          & \textbf{4.694}  \\ \hline
t-SNE & 3 & 3.500            & 8.600          & 8.300           \\ \hline
t-SNE & 4 & 2.333            & 8.500          & 7.545           \\ \hline
UMAP  & 0 & 2.5              & 10.000         & -               \\ \hline
UMAP  & 1 & 2.333            & \textbf{2.438} & 10.000          \\ \hline
UMAP  & 2 & 0.667            & 4.375          & \textbf{3.045}  \\ \hline
UMAP  & 3 & \textbf{0.500}   & 3.364          & 5.833           \\ \hline
UMAP  & 4 & 1.000            & 6.833          & 6.400           \\ \hline
\end{tabular}
}
\end{sc}
\end{small}
\end{center}
\vskip -0.1in
\end{table}

\textbf{BA-Grid:} Table \ref{ba_grid_purity} lists the average concept purity for the proposed method and alternative designs applied to BA-Grid. Simply comparing the number of concept purity scores computed across the different techniques, it can be state that the $k$-Means clustering algorithm is preferable. This is based on the lack of scores computed for designs using AHC and DBSCAN clustering. The reason the scores are not computed is because the concept contained too many nodes. However, considering that the grid structure only has 9 nodes this indicates that the BA base graph is discovered, which is not a valuable motif in this task. Moreover, considering the concept scores computed using $k$-Means clustering, improved results are obtained on the raw activation space in later convolutional layers.

\begin{table}[t]
\caption{Average purity scores for concepts discovered per layer using the different designs on BA-Grid.}
\label{ba_grid_purity}
\vskip 0.15in
\begin{center}
\begin{small}
\begin{sc}
{%
\begin{tabular}{|c|c|c|c|c|}
\hline
\multirow{2}{*}{\textbf{\begin{tabular}[c]{@{}c@{}}DR \\ Method\end{tabular}}} &
  \multirow{2}{*}{\textbf{Layer}} &
  \multicolumn{3}{c|}{\textbf{Average Concept Purity}} \\ \cline{3-5} 
      &   & \textbf{$k$-Means} & \textbf{AHC}    & \textbf{DBSCAN} \\ \hline
Raw   & 0 & -  & -  & -               \\ \hline
Raw   & 1 & 0.000  & - & -                \\ \hline
Raw   & 2 & 0.000  & - & -            \\ \hline
Raw   & 3 & \textbf{0.000}  & - & -           \\ \hline
Raw   & 4 & -  & - & -            \\ \hline
PCA   & 0 & -  & - & -               \\ \hline
PCA   & 1 & - & - & -  \\ \hline
PCA   & 2 & 0.000 & - & -          \\ \hline
PCA   & 3 &\textbf{0.000}  & - & \textbf{0.000}           \\ \hline
PCA   & 4 & - & - & -           \\ \hline
t-SNE & 0 & 8.000  & -  & -            \\ \hline
t-SNE & 1 & 0.000  & - & -           \\ \hline
t-SNE & 2 & -  & 6.000 & -  \\ \hline
t-SNE & 3 & \textbf{0.000}  & - & -            \\ \hline
t-SNE & 4 & -  & - & -           \\ \hline
UMAP  & 0 & -  & \textbf{0.000} & 8.000                \\ \hline
UMAP  & 1 & -  & - & -          \\ \hline
UMAP  & 2 & -  & - & -  \\ \hline
UMAP  & 3 & 3.000  & - & \textbf{0.000}            \\ \hline
UMAP  & 4 & \textbf{2.500}  & - & -       \\ \hline
\end{tabular}%
}
\end{sc}
\end{small}
\end{center}
\vskip -0.1in
\end{table}

\textbf{Tree-Cycle:} These results are supported by the concept purity scores computed for the Tree-Cylces dataset, presented in Table \ref{tree_cycle_purity}. While fewer values are missing for AHC and DBSCAN, the concept purity scores are generally higher than those for $k$-Means, indicating that the concepts discovered are less pure. The results for $k$-Means across the different dimensionality reduction techniques demonstrate that high-quality concepts are recovered. However, contrary to our earlier conclusions the purest concepts appear to be achieved on the penultimate neighbourhood aggregation layer.

\begin{table}[t]
\caption{Average purity scores for concepts discovered per layer using the different designs on Tree-Cycles.}
\label{tree_cycle_purity}
\vskip 0.15in
\begin{center}
\begin{small}
\begin{sc}
{%
\begin{tabular}{|c|c|c|c|c|}
\hline
\multirow{2}{*}{\textbf{\begin{tabular}[c]{@{}c@{}}DR \\ Method\end{tabular}}} &
  \multirow{2}{*}{\textbf{Layer}} &
  \multicolumn{3}{c|}{\textbf{Average Concept Purity}} \\ \cline{3-5} 
      &   & \textbf{$k$-Means} & \textbf{AHC}    & \textbf{DBSCAN} \\ \hline
Raw   & 0 & 1.200  & 5.913  & -               \\ \hline
Raw   & 1 & \textbf{0.286}  & 10.077 & \textbf{2}               \\ \hline
Raw   & 2 & 1.167  & 5.500 & 5.200           \\ \hline
Raw   & 3 & 0.667  & \textbf{3.000} & 6.300           \\ \hline
PCA   & 0 & 1.000  & 11.150 & -               \\ \hline
PCA   & 1 &\textbf{0.250}  & 5.000  & \textbf{1.667}  \\ \hline
PCA   & 2 & 2.000  & 3.947 & 12.750          \\ \hline
PCA   & 3 & 0.571  & \textbf{3.000} & 4.000           \\ \hline
t-SNE & 0 & 2.000  & 6.190  & 5.132           \\ \hline
t-SNE & 1 & 0.250  & 7.833 & 4.909           \\ \hline
t-SNE & 2 & \textbf{0.000}  & \textbf{5.833} & \textbf{4.694}  \\ \hline
t-SNE & 3 & 0.688  & 6.767 & 8.300           \\ \hline
UMAP  & 0 & 2.214  & \textbf{4.345} & -               \\ \hline
UMAP  & 1 & \textbf{0.000}  & 8.100 & 10.000          \\ \hline
UMAP  & 2 & 0.167  & 5.016 & \textbf{3.045}  \\ \hline
UMAP  & 3 & 1.000  & 6.382 & 5.833           \\ \hline
\end{tabular}%
}
\end{sc}
\end{small}
\end{center}
\vskip -0.1in
\end{table}

\textbf{REDDIT-BINARY:} Finally, Table \ref{reddit_purity} summarises the concept purity scores obtained for the REDDIT-BINARY dataset. The number of the layers corresponds to that in the evaluation of the completeness score for the REDDIT-BINARY dataset. We observe a similar trend as to that observed on the completeness scores: concept discovery on the pooling layers appears to provide purer concepts. However, this must be set into context with the number of concepts included in this score. In general, fewer concepts are included int the calculations for the pooling layers. Moreover, the need for graph alignment to extract concepts from the clustered graphs is a significant drawback in the use of pooling layers. Hence, operating on the convolutional layers is still desirable. Furthermore, improved concept purity scores are obtained for later layers. The choice in dimensionality technique does not appear to make a difference. However, $k$-Means clustering appears to outperform AHC and DBSCAN in the discovery of pure concepts. While AHC has more concept purity scores of 0, the range reaches up to 8.582, while the purity scores for $k$-Means are consistently below 4.200. In contrast, DBSCAN appears to perform the worst, with a number of scores not being computed because the concepts exceed the limitations. In summary, it can be stated that the results support the selected method.

\begin{table}[t]
\caption{Average purity score for concepts discovered per layer using the different designs for REDDIT-BINARY.}
\label{reddit_purity}
\vskip 0.15in
\begin{center}
\begin{small}
\begin{sc}
{%
\begin{tabular}{|c|c|c|c|c|}
\hline
\multirow{2}{*}{\textbf{\begin{tabular}[c]{@{}c@{}}DR \\ Method\end{tabular}}} &
  \multirow{2}{*}{\textbf{Layer}} &
  \multicolumn{3}{c|}{\textbf{Average Concept Purity}} \\ \cline{3-5} 
      &   & \textbf{$k$-Means} & \textbf{AHC}    & \textbf{DBSCAN} \\ \hline
Raw   & 0 & 0.923 & 1.000 & 9.667                \\ \hline
Raw   & 1 & 0.000 & 0.000 & -               \\ \hline
Raw   & 2 & 2.033 & 8.583 & 4.500           \\ \hline
Raw   & 3 & 0.000  & 0.000 & -           \\ \hline
Raw   & 4 & 2.825  & 6.000 & 3.45               \\ \hline
Raw   & 5 & 0.000  & 0.000 & \textbf{0.000}  \\ \hline
Raw   & 6 & 2.731  & 8.182 & 7.375\\ \hline
Raw   & 7 & \textbf{0.000}  & \textbf{0.000} & - \\ \hline
PCA   & 0 & 3.167   & -     & -               \\ \hline
PCA   & 1 & 0.000  & -     & -  \\ \hline
PCA   & 2 & 3.222  & 0.000 & 2.333          \\ \hline
PCA   & 3 & 0.000  & 0.000 & 0.000           \\ \hline
PCA   & 4 & 3.200  & 0.000 & 0.000   \\ \hline
PCA   & 5 & 0.000  & 0.000 & 0.000           \\ \hline
PCA   & 6 & 0.400  & 1.333 & \textbf{0.000}           \\ \hline
PCA   & 7 & \textbf{0.000}  & \textbf{0.000} & -           \\ \hline
t-SNE & 0 & 2.143  & 0.000 & -           \\ \hline
t-SNE & 1 & 0.000  & 0.000 & 0.000           \\ \hline
t-SNE & 2 & 0.000  & 0.000 & 6.000  \\ \hline
t-SNE & 3 & -  & - & -           \\ \hline
t-SNE & 4 & 4.167  & 0.000 & \textbf{0.000}            \\ \hline
t-SNE & 5 & 0.000  & 3.333 & -           \\ \hline
t-SNE & 6 & 3.250  & 0.000 & -           \\ \hline
t-SNE & 7 & \textbf{0.000}  & \textbf{0.000} & -           \\ \hline
UMAP  & 0 & 0.000  & 0.500 & 0.000              \\ \hline
UMAP  & 1 & 0.000  & - & 0.000        \\ \hline
UMAP  & 2 & 0.000  & 0.000 & 0.000   \\ \hline
UMAP  & 3 & 0.000  & 0.000 & -          \\ \hline
UMAP  & 4 & 2.786  & 0.000 & 0.000          \\ \hline
UMAP  & 5 & 0.000  & \textbf{0.000} & 0.000          \\ \hline
UMAP  & 6 & 1.333  & 2.500 & 0.000          \\ \hline
UMAP  & 7 & \textbf{0.000}  & - & \textbf{0.000}          \\ \hline
\end{tabular}%
}
\end{sc}
\end{small}
\end{center}
\vskip -0.1in
\end{table}

\subsubsection{Key Results}

We show that improved concepts can be discovered using $k$-Means on later convolutional layers in the model. We demonstrate that AHC and DBSCAN provide less reasonable concept representations to the user, as neighbouring nodes, rather than similar nodes, are grouped. Furthermore, we demonstrate the applicability of the proposed method across different node and graph classification datasets.

\section{Implementation Details}
\label{C}

\subsection{Graph Neural Network Definitions}

As previously stated, we train a GCN model for each dataset according to the recommendations set by GNNExplainer \cite{Ying2019}. In total, we define six architectures to accommodate the different datasets. Table \ref{architectures} summarises the architectures and training details defined, as well as the applicable datasets. The general structure of the models is a number of convolutional layers for neighbourhood aggregation followed by a linear prediction layer. An exception to this are the network architectures for graph classification, which have an additional pooling layer after the final convolutional layer. We determine the number of convolutional layers and hidden units experimentally by reviewing the model performance. We pass the final activations through a log Softmax function to receive the probabilities of the classes predicted \cite{NEURIPS2019_9015}.

We train the models on an 80:20 training-testing split of the data in a basic training loop using gradient descent and backpropagation. We choose the Adam optimisation algorithm for gradient descent, as it has shown to produce more stable training results than stochastic gradient descent \cite{Kingma2015}. The loss function used is the negative log likelihood loss, as this yields the cross entropy loss in combination with the Softmax function applied in the models \cite{NEURIPS2019_9015}. We implement this training functionality and the model architectures using the framework PyTorch \cite{NEURIPS2019_9015}. This framework is chosen based on the availability of the PyTorch Geometric library \cite{Fey/Lenssen/2019}, which provides various functionalities for defining GNN architectures. Table \ref{node_class_acc} summarises the testing and training accuracies achieved.

\begin{table*}[t]
\caption{These architectures were defined experimentally with the goal of meeting the minimum threshold of 95\% and 85\% training accuracy for node and graph classification tasks, respectively.}
\label{architectures}
\vskip 0.15in
\begin{center}
\begin{small}
\begin{sc}
{%
\begin{tabular}{|l|l|c|c|}
\hline
\textbf{Dataset} &
  \textbf{Model Definition} & \textbf{Epochs} & \textbf{Learning Rate}\\ \hline %
BA-Shapes &
  \begin{tabular}[c]{@{}l@{}}3x Convolution Layer with 20 units, ReLU function\\ 1x Linear Layer with 20 units, log Softmax function\end{tabular} & 3000 & 0.001\\ \hline
BA-Grid &
  Same architecture as for BA-Shapes & 3000 & 0.001 \\ \hline
BA-Community & \begin{tabular}[c]{@{}l@{}}6x Convolutional Layer with 50 units, ReLU function\\ 1x Linear Layer with 50 hidden units, log Softmax function\end{tabular} & 6000 & 0.001
  \\ \hline
Tree-Cycles &
  \begin{tabular}[c]{@{}l@{}}3x Convolutional Layer with 50 units, ReLU function\\ 1x Linear Layer with 50 hidden units, log Softmax function\end{tabular} & 7000 & 0.001\\ \hline
Tree-Grid &
  \begin{tabular}[c]{@{}l@{}}7x Convolutional Layer with 20 units, ReLU function\\ 1x Linear Layer with 20 hidden units, log Softmax function\end{tabular} & 10000 & 0.001\\ \hline
Mutagenicity &
  \begin{tabular}[c]{@{}l@{}}4x Convolutional Layer with 30 units, ReLU function\\ 1x Pooling Layer pooling on the global maximum\\ 1x Linear layer with 30 units, log Softmax function\end{tabular} & 10000 & 0.005 \\ \hline
    REDDIT-BINARY &
  \begin{tabular}[c]{@{}l@{}} 4x Convolutional Layer with 40 units, ReLU function\\ 1x Pooling Layer pooling on the global maximum\\ 1x Linear layer with 40 units, log Softmax function\end{tabular} & 3000 & 0.005 \\ \hline
\end{tabular}%
}
\end{sc}
\end{small}
\end{center}
\vskip -0.1in
\end{table*}

\begin{table}[t]
\caption{Accuracy of the Models trained for each dataset.}
\label{node_class_acc}
\vskip 0.15in
\begin{center}
\begin{small}
\begin{sc}
{%
\begin{tabular}{|l|c|c|}
\hline
\textbf{Dataset / Model} & \textbf{\begin{tabular}[c]{@{}l@{}}Training \\ Accuracy\end{tabular}} & \textbf{\begin{tabular}[c]{@{}l@{}}Testing\\ Accuracy\end{tabular}} \\ \hline %
BA-Shapes & 97.2\% & 95.6\% \\ \hline
BA-Grid & 99.0\% & 99.0\% \\ \hline
BA-Community & 95.7\% & 95.2\% \\ \hline
Tree-Cycles & 96.0\% & 95.5\% \\ \hline
Tree-Grid & 95.1\% & 95.7\% \\ \hline
Mutagenicity & 89.1\% & 86.9\% \\ \hline
REDDIT-BINARY & 93.0\% & 89.9\% \\ \hline 
\end{tabular}
}
\end{sc}
\end{small}
\end{center}
\vskip -0.1in
\end{table}

\subsection{Extraction of the Activation Space}

Given a dataset and trained GNN, the first step in the method proposed is the extraction of the raw activation space of the model. In order to extract the raw activation space, we register hooks \cite{NEURIPS2019_9015} for each layer of the model. These hooks allow to attach a function to the layer, which is executed when a forward pass if performed. The function we define simply returns the output of the layer, which is the activation space of the model for the given layer and data. In order to get the full activation space, a full pass of the given dataset is performed.

\subsection{Dimensionality Reduction and Clustering Techniques}

The second step of the method is to perform a mapping from the activation space to the concept space. In our experimental setup (Appendix \ref{B}) this step is broken into two parts due to the additional preprocessing step applied to the activation space. We substitute the raw activation space for an activation space reduced using PCA, t-SNE and UMAP to explore the impact of dimensionality reduction on the quality of the results. We implement this using functionality available as part of the scikit-learn \cite{scikit-learn} and SciPy \cite{2020SciPy-NMeth} library.

The second part of the second step is to perform a mapping from the activation space to the concept space, for which we propose $k$-Means clustering. In order to validate our design choice, we substitute the $k$-Means algorithm with AHC using Ward's method and DBSCAN clustering. We choose these two clustering techniques based on their prominence and wide application \cite{ArchanaPatel2016}. We continue using the implementations provided by scikit-learn \cite{scikit-learn} and SciPy \cite{2020SciPy-NMeth} for this. 

The number of clusters defined for the experiments detailed in Appendix \ref{B} varies between the clustering techniques and datasets. In respect to $k$-Means clustering, we initially trial $5 \leq k \leq 30$ in steps increasing by 5, which is based on the number of clusters that can be observed when visualising the activation space and our knowledge of motifs. We observe redundancy in the concepts extracted for values greater than 10, which motivates our choice of $k=10$ for the experiments on node classification datasets. Moreover, this choice in $k$ avoids overly optimistic results and allows comparing the results across datasets. An exception to this is the BA-Community dataset for which we define $k=30$, as this produced enhanced concepts. This can be explained by the dataset consisting of two BA-Graphs. We also choose $k=30$ for the graph classification datasets due to the increased variety exhibited in the individual graphs of the datasets.

Similarly, we experiment with the parameter choices of DBSCAN. DBSCAN does not require defining the number of clusters, but the value for $\epsilon$ and $MinPts$. We choose the value of $\epsilon$ based on the range of the activation space and the $MinPts$ based on the number of motifs encoded in the graph. In contrast, we take a more systematic approach in defining the number of clusters for AHC. We plot the dendrogram of the activations for each layer, which visualises the splitting of clusters as a tree, where the depth is the distance at which a new cluster is formed. We use this to estimate the number of suitable clusters. While the number of clusters and thus concepts varies between the individual methods, it avoids overestimating or underestimating the performance of the different clustering techniques.

\subsection{Visualisation Module}

We implement a visualisation module for the concept representation step described in Section \ref{concept_rep}. We choose the parameter $n$ for the experiments detailed in Appendix \ref{B} based on the knowledge about the motif to be extracted. For example, in the case of BA-Shapes, the nodes within a two-hop distance are explored, as this allows to visualise the full house structure. We also use this technique for visualising the concepts extracted for graph classification tasks. We provide additional functionality for customising the visualisation, such as displaying the node features. This is performed using the libraries Matplotlib \cite{Hunter:2007} and NetworkX \cite{Hagberg2008} for visualisation and graph representation, respectively.

\subsection{Evaluation Module}

\subsubsection{Concept Completeness and Purity}

The evaluation module includes functionality implementing the concept completeness score \cite{Yeh2019} and concept purity score. We implement the former by using the clustering model trained to assign concepts to the input data of the datasets. The training-split of the predicted concepts and the provided class labels is then used to train a decision tree. We implement this using the scikit-learn \cite{scikit-learn} library implementation of the classifiers.

In order to evaluate the purity of a concept, the graph similarity of the top representations of a concept is computed. We use the NetworkX \cite{Hagberg2008} implementation of the graph edit distance for this. As computing the graph edit distance is expensive, we only perform this for the three concept representations closest to a cluster centre and take an average. We choose these concept representations as they can be seen as the strongest representations of the concept. The limit on the number of concept representations included in the calculation is required due to the large number of nodes and consequently substructures that would need to be compared. For example, the BA-Shapes dataset has 700 nodes \cite{Ying2019}, while the Mutagenicity dataset consists of 4337 graphs \cite{KKMMN2016}. We also limit the calculation to concept representations with up to 15 nodes. However, this limit is sufficient for comparing the concepts we expect to extract, as the motifs to be extracted consist of less than 15 nodes. The limit on the number of nodes and concept representations compared should be adjusted for other datasets based on the number of samples and the complexity of the expected concepts, which is indicated by $n$ in the number of $n$-hop neighbours visualised. Alternatively, the concept representations closest to a cluster centroid and the concept representations furthest away could be compared to explore the variance within a cluster.

\subsection{Computing Resources}

The experiments are run on a machine with a 2,3 GHz 8-Core Intel Core i9 CPU and a AMD Radeon Pro 5500M 8 GB GPU.

\section{Further Evaluation of Human-in-the-Loop Parameter Adaption}
\label{D}

To substantiate the conclusions drawn, we perform the same analysis for REDDIT-BINARY, a real-world dataset. Table \ref{new_res_otherdata} summarises the completeness and concept purity scores obtained when varying $n$ and $k$. Here $n$ and $k$ are picked in the same manner as before. Reviewing the concept completeness scores computed, it can be stated that $k = 20$ is sufficient for the dataset, as the concept completeness scores repeat across the different settings for $k$. Even though the same concept completeness score is achieved with higher values for $k$, a lower $k$ is desirable to avoid redundancy in the concepts, as a higher $k$ merely splits the same concept into multiple clusters. While the BA-Shapes dataset had consistent completeness scores for a given $k$, here the values vary. However, this is not caused by the variation of $n$ but rather by the implementation of $k$-Means clustering causing slightly varied clusters across runs.

Reviewing the average purity scores obtained, it is evident that $n = 1$ is desirable, as the lowest concept purity score is obtained. This can be attributed to this being a graph classification tasks with widely varying graphs. As more neighbours are visualised, the concept representations are more likely to differ. Given these observations, the most suitable hyperparameter setting appears to be $k = 20$ and $n = 1$.

\begin{table}[t]
\caption{The concept completeness and purity score of the concepts discovered for the REDDIT-BINARY dataset when adapting $n$ and $k$.}
\label{new_res_otherdata}
\vskip 0.15in
\begin{center}
\begin{small}
\begin{sc}
{%
\begin{tabular}{|c|c|c|c|}
\hline
\textbf{k}          & \textbf{n} & \textbf{\begin{tabular}[c]{@{}c@{}}Completeness \\ Score\end{tabular}} & \textbf{\begin{tabular}[c]{@{}c@{}}Average Purity\\ Score\end{tabular}} \\ \hline
\multirow{3}{*}{20}  & 1          & \textbf{0.948}                                                                  & \textbf{2.056}                                                                      \\ \cline{2-4} 
                    & 2          & 0.938                                                                  & 9.000                                                                      \\ \cline{2-4} 
                    & 3          & 0.940                                                                  & 6.000                                                                      \\ \hline
\multirow{3}{*}{30} & 1          & \textbf{0.948}                                                                  & \textbf{2.056}                                                                      \\ \cline{2-4} 
                    & 2          & 0.937                                                                  & 3.115                                                                      \\ \cline{2-4} 
                    & 3          & 0.940                                                                  & 6.000                                                                      \\ \hline
\multirow{3}{*}{40} & 1          & \textbf{0.948}                                                                   & \textbf{2.056}                                                                      \\ \cline{2-4} 
                    & 2          & 0.937                                                                  & 4.500                                                                      \\ \cline{2-4} 
                    & 3          & 0.940                                                                 & 6.000                                                                      \\ \hline
\end{tabular}%
}
\end{sc}
\end{small}
\end{center}
\vskip -0.1in
\end{table}

\begin{figure}[ht]
\vskip 0.2in
\begin{center}
\centerline{\includegraphics[width=\columnwidth, trim = 0 0cm 0cm 0cm, clip]{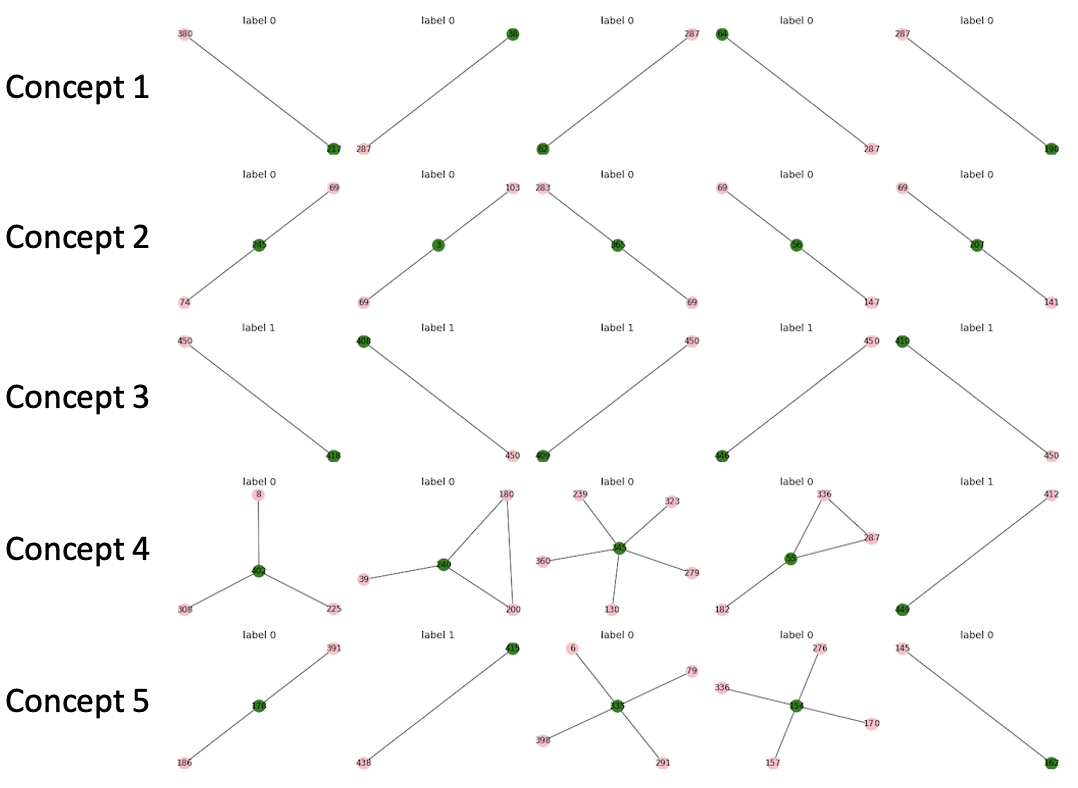}}
\caption{A subset of the concepts discovered for REDDIT-BINARY, where $k = 20$ and $n = 1$.}
\label{new_res5}
\end{center}
\vskip -0.2in
\end{figure}

\begin{figure}[ht]
\vskip 0.2in
\begin{center}
\centerline{\includegraphics[width=\columnwidth, trim = 0 0cm 0 0cm, clip]{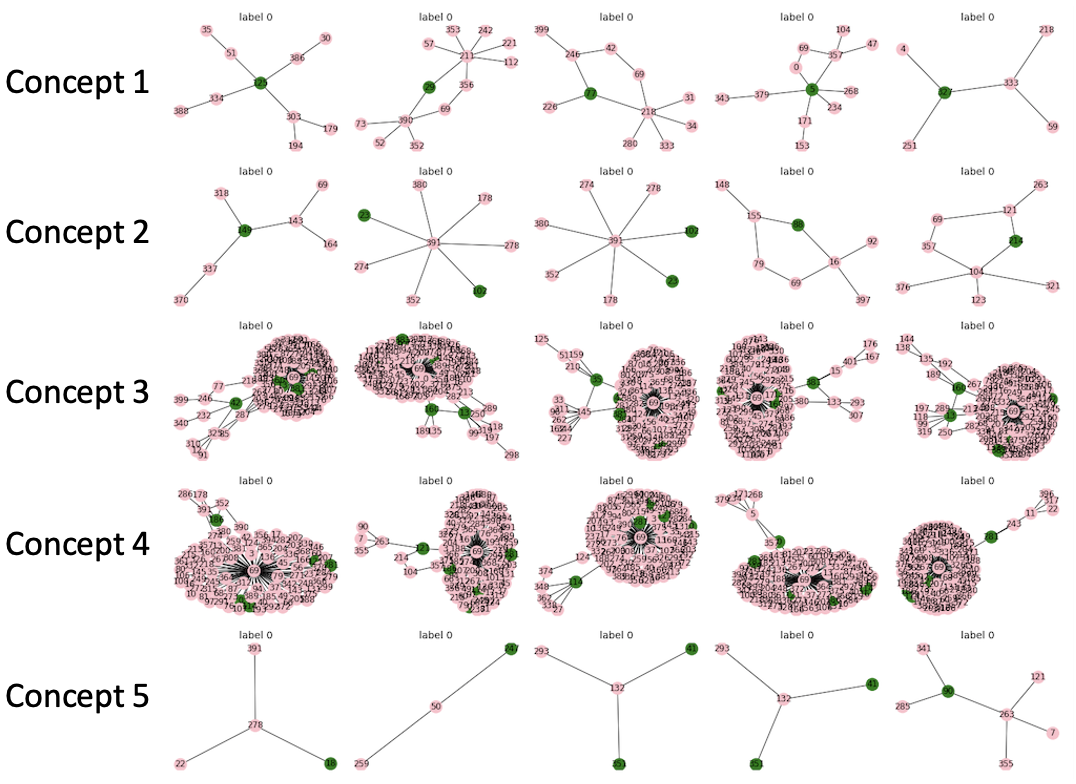}}
\caption{A subset of the concepts discovered for REDDIT-BINARY, where $k = 20$ and $n = 2$.}
\label{new_res6}
\end{center}
\vskip -0.2in
\end{figure}

\begin{figure}[ht]
\vskip 0.2in
\begin{center}
\centerline{\includegraphics[width=\columnwidth, trim = 0 0 0 0cm, clip]{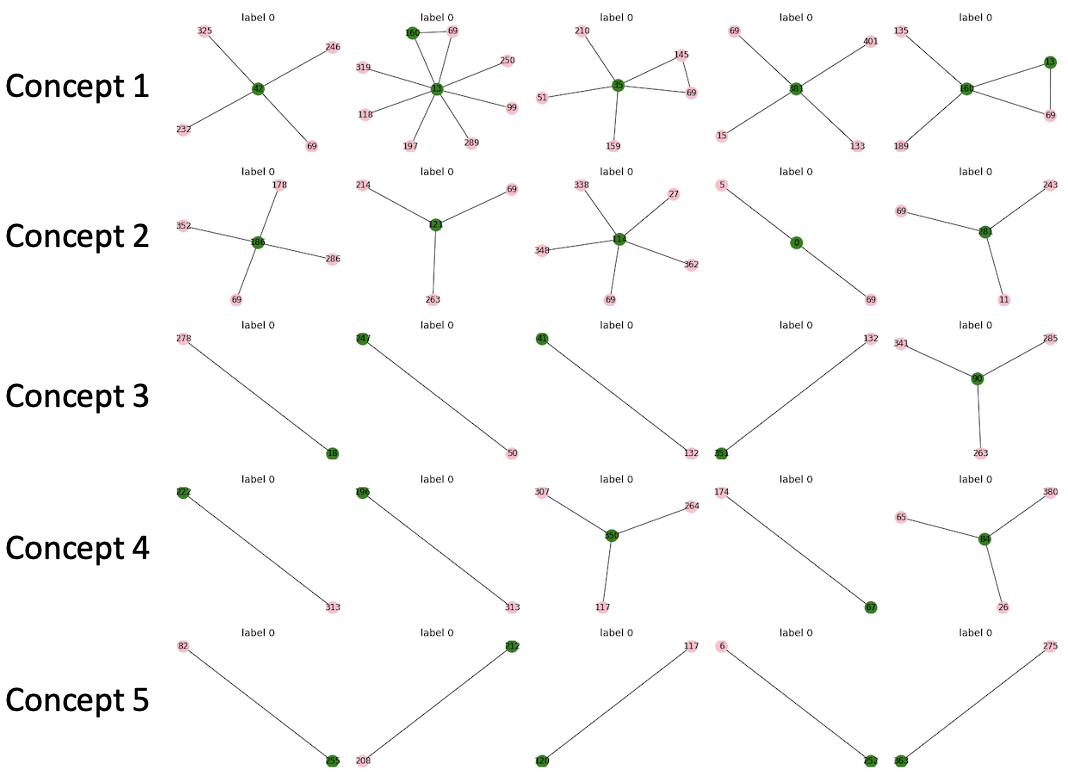}}
\caption{A subset of the concepts discovered for REDDIT-BINARY, where $k = 30$ and $n = 1$.}
\label{new_res7}
\end{center}
\vskip -0.2in
\end{figure}

Examining the qualitative results, slightly different conclusions can be drawn. For example, Figures \ref{new_res5} and \ref{new_res6} show a subset of the concepts extracted when $k = 20$ and $n$ is set to 1 and 2, respectively. Arguably, $n = 1$ does not visualise enough of the neighbourhood and makes the concepts appear redundant, as only very simple structures are visualised. In turn, when $n = 2$ more complex structure are visualised and it becomes clearer which concepts are representative of users reacting to a topic versus experts replying to questions. Examining Figure \ref{new_res7} in conjunction with these, our earlier observation, that redundancy in the concepts can be observed for higher values of $k$, is supported. Based on this, the best setting can be argued to be $k = 20$ and $n = 2$. Reasoning about these results in conjunction with the quantitative results, it can be stated that it is vital for the user to chose the parameters to ensure that the user can interpret the explanations and make sense of them. While the concept completeness and purity score can guide the user, they are insufficient to pick a suitable $k$ and $n$. They should only be used in conjunction with the visualisations and the user's judgement of the usefulness and meaning of these.

\section{Further Evaluation of the Proposed Method across different Datasets}
\label{E}

\subsection{Quantitative Results}

\subsubsection{BA-Community}

\begin{figure}[ht]
\vskip 0.2in
\begin{center}
\centerline{\includegraphics[width=\columnwidth, trim={3cm 0 3cm 2cm},clip]{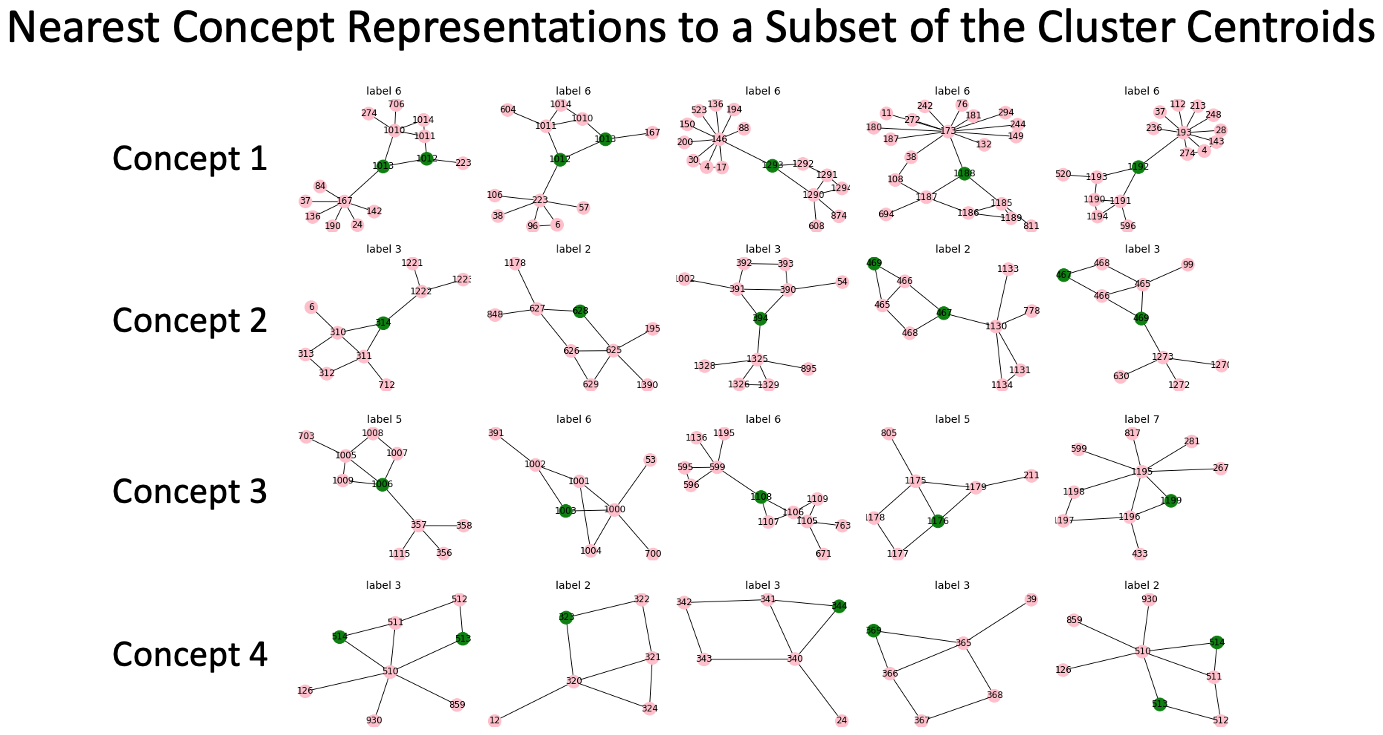}}
\caption{A subset of the concepts discovered for BA-Community. Green nodes are the nodes clustered together, while the pink nodes form the expanded neighbourhood.}
\label{ba_community}
\end{center}
\vskip -0.2in
\end{figure}

Figure \ref{ba_community} visualises the main concepts extracted for BA-Community. The house structure encoded is successfully discovered as part of the set of concepts, however, there is some misclassification. For example, in concept 2 the top node and bottom node of a house are grouped together, indicating that the GNN does not clearly separate the concept space. This is also evident when comparing the labels of the nodes printed above the concept representations. However, the nodes between house structures belonging to different communities seem to be separated in general. This indicates that the GNN draws a clear distinction based on the node features of the dataset. In conclusion, the desired motifs are extracted, however, the activation space is not separated clearly, leading to noisy concepts. 

\begin{figure}[ht]
\vskip 0.2in
\begin{center}
\centerline{\includegraphics[width=0.6\columnwidth, trim={0 0 0 1cm},clip]{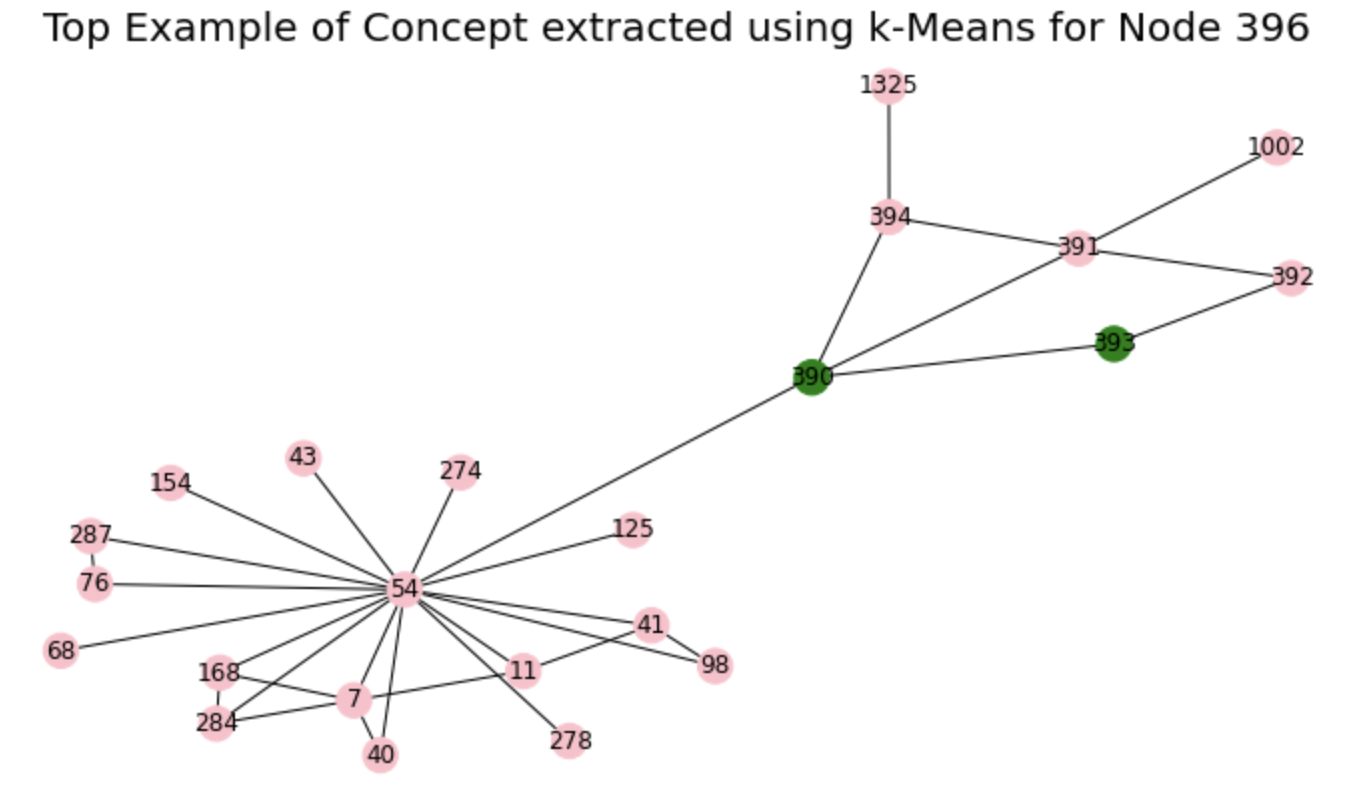}}
\caption{The concept representation representative of the cluster explaining a node in BA-Community. The pink nodes are the neighbourhood of the clustered node in green.}
\label{ba_community_exp1}
\end{center}
\vskip -0.2in
\end{figure}

\begin{figure}[ht]
\vskip 0.2in
\begin{center}
\centerline{\includegraphics[width=\columnwidth, trim={0cm 0 0cm 1.5cm},clip]{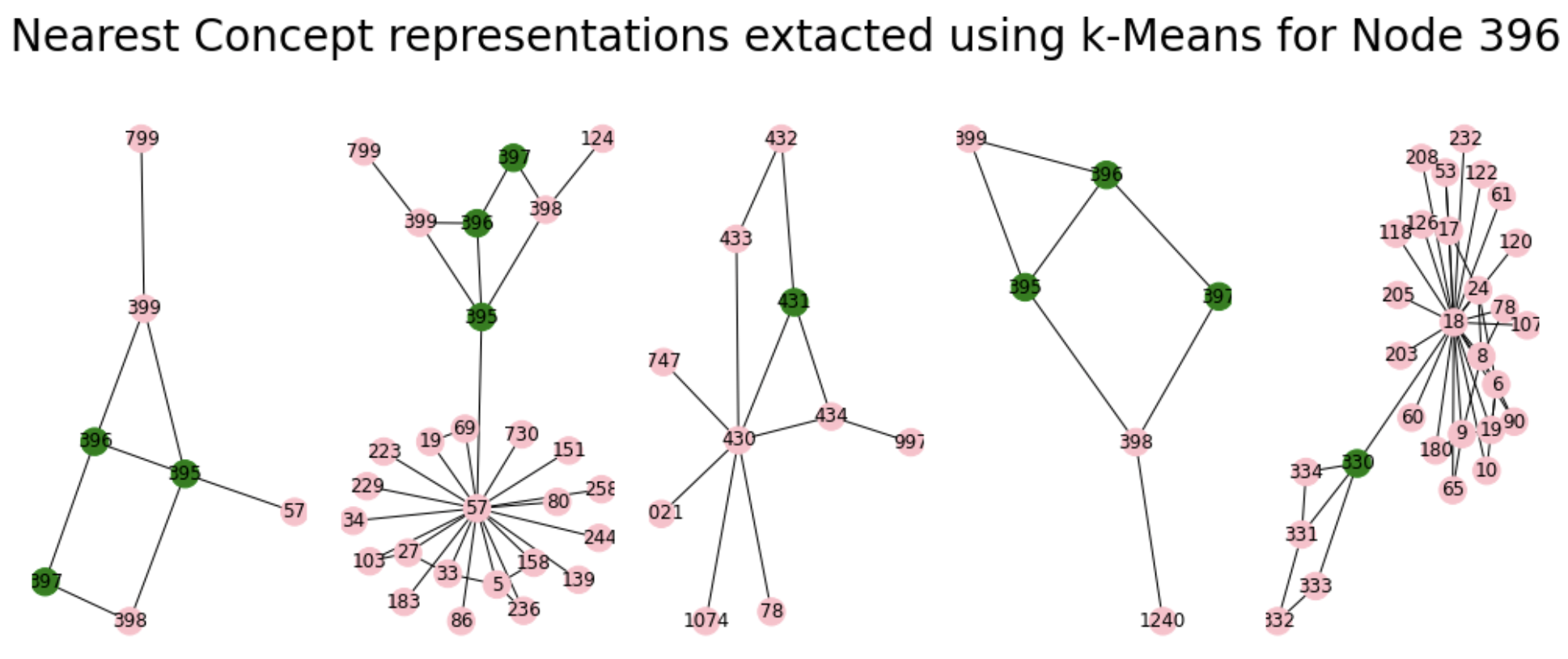}}
\caption{The concept representations nearest to the BA-Community node 396 explained, visualised to capture cluster variance. The pink nodes are the neighbourhood of the clustered nodes in green.}
\label{ba_community_exp2}
\end{center}
\vskip -0.2in
\end{figure}

\begin{figure}[ht]
\vskip 0.2in
\begin{center}
\centerline{\includegraphics[width=0.7\columnwidth, trim={0 0 0 1cm},clip]{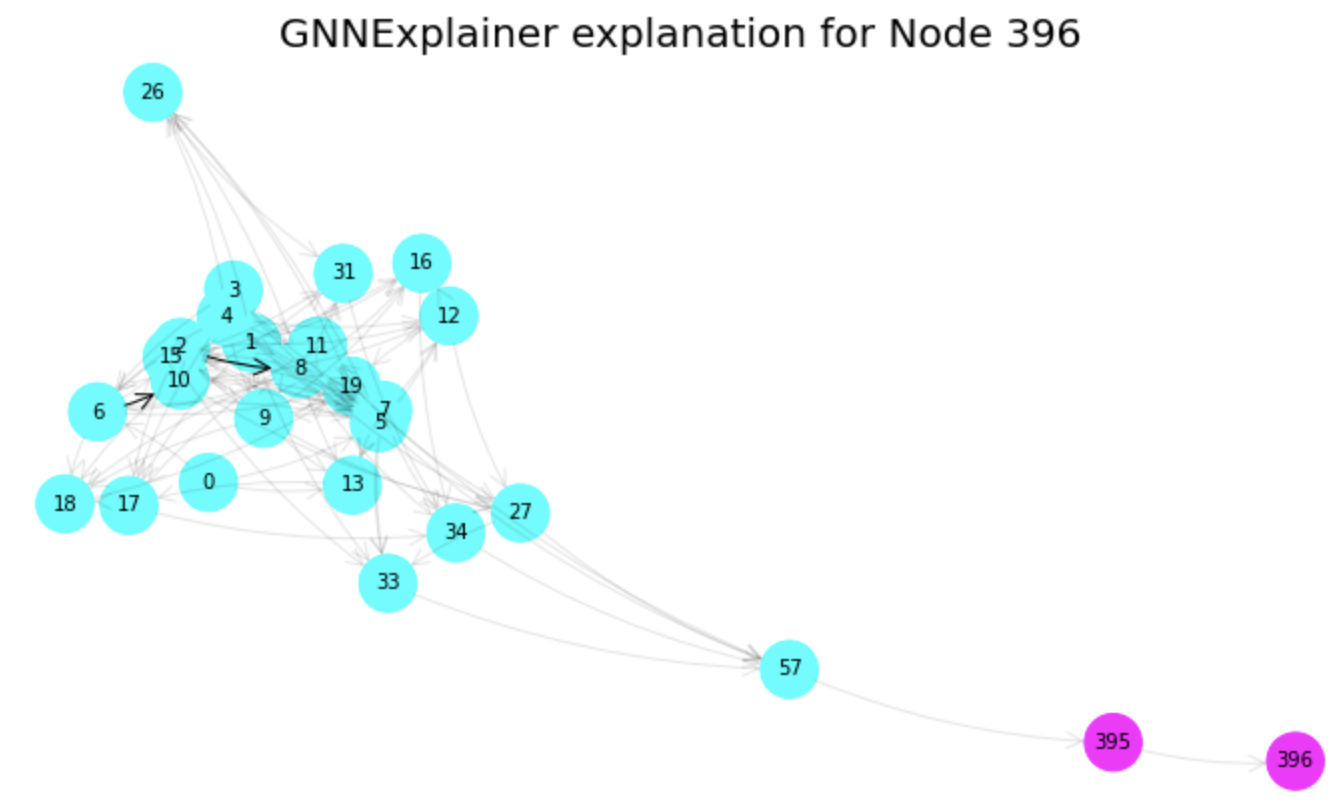}}
\caption{GNNExplainer explanation produced for BA-Community node 396. The purple nodes are part of the house structure.}
\label{ba_community_gnnexplainer}
\end{center}
\vskip -0.2in
\end{figure}

Figure \ref{ba_community_exp1} and \ref{ba_community_exp2} depict the concept-based explanations extracted for node 396, which is a middle node in a house structure. The explanation extracted is a house structure attached to one of the BA base graphs. This is confirmed by the closest representations of the concept in Figure \ref{ba_community_exp2}, however, inter-cluster variance can be observed through nodes of the same house structure being grouped together. This indicates that the GNN does not separate the nodes well in the activation space, which can be attributed to the simplistic house structure being crucial for the prediction of six entangled classes. These observations are not in line with the general trend observed across the other datasets, which do not have node features. One conclusion that could be drawn is that the method performs poorly on graphs including node features. However, it could also be indicative of shortcomings of the model. Nevertheless, it can be argued that the concept-based explanation is superior to the one of GNNExplainer \cite{Ying2019}, depicted in Figure \ref{ba_community_gnnexplainer}. This is based on the GNNExplainer explanation only highlighting the base graph and not identifying the house motif.

\subsubsection{BA-Grid}

\begin{figure}[ht]
\vskip 0.2in
\begin{center}
\centerline{\includegraphics[width=\columnwidth, trim={1.8cm 0 1.8cm 2cm},clip]{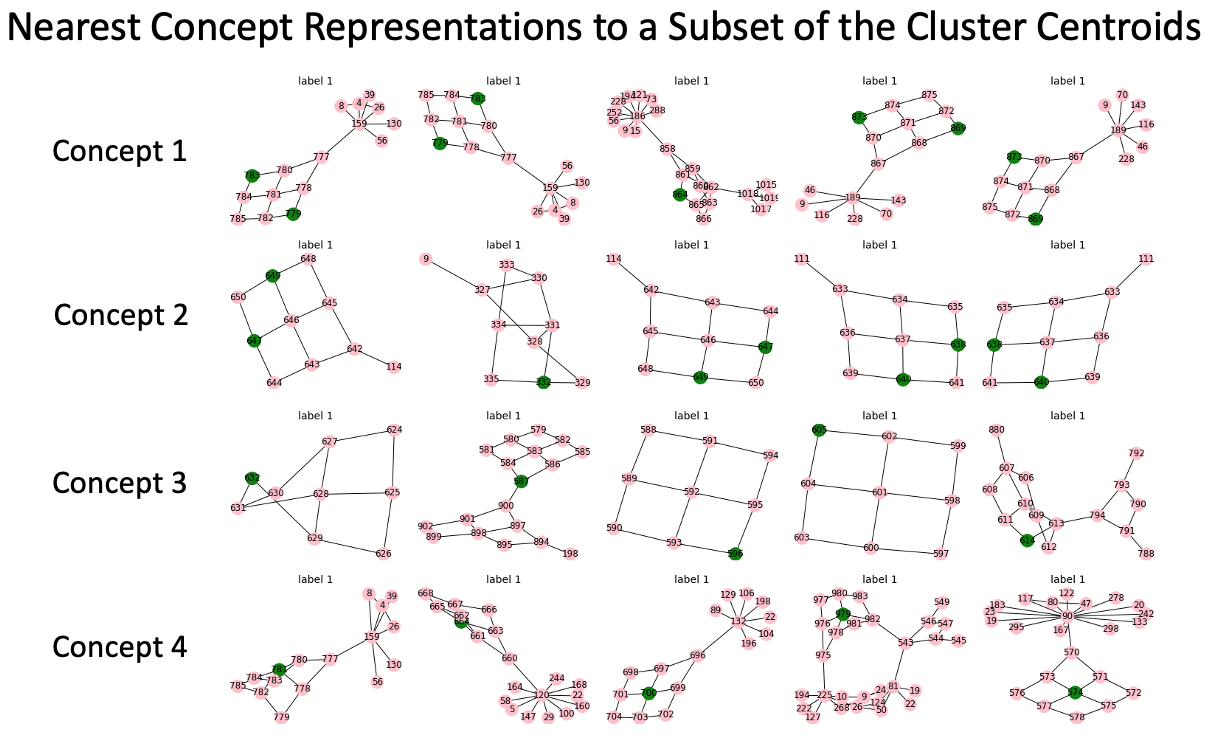}}
\caption{A subset of the concepts discovered for BA-Grid. Green nodes are the nodes clustered together, while the pink nodes form the expanded neighbourhood.}
\label{ba_grid}
\end{center}
\vskip -0.2in
\end{figure}

We extract high quality concepts for the BA-Grid dataset, as shown in Figure \ref{ba_grid}. We omit some of the concepts discovered, as these are representative of the BA base graph. The first concept discovered shows the grid motif, as well as its attachment to the BA base graph. Four of the five concept representations have two green nodes, which can be explained by these nodes both being part of the same grid. As the grid is invariant to translation in space, the nodes both represent the same node in the structure. Concept 2 also shows the base graph, however, less of the BA base graph is visualised. This is due to the location of the nodes. Again, the structures have two green nodes, because the grid is invariant to orientation. Concepts 3 and 4 also visualise the grid motif, however, the concepts appear more convoluted, which is due to the random edges in the graph leading to greater connectivity. However, it can clearly be observed that a corner node and the middle node are grouped together as a concept. In conclusion, the desired motif can be recovered. The visualisations show global concepts associated with the prediction of given nodes in a grid.

\begin{figure}[ht]
\vskip 0.2in
\begin{center}
\centerline{\includegraphics[width=0.6\columnwidth, trim={0 0 0 1cm},clip]{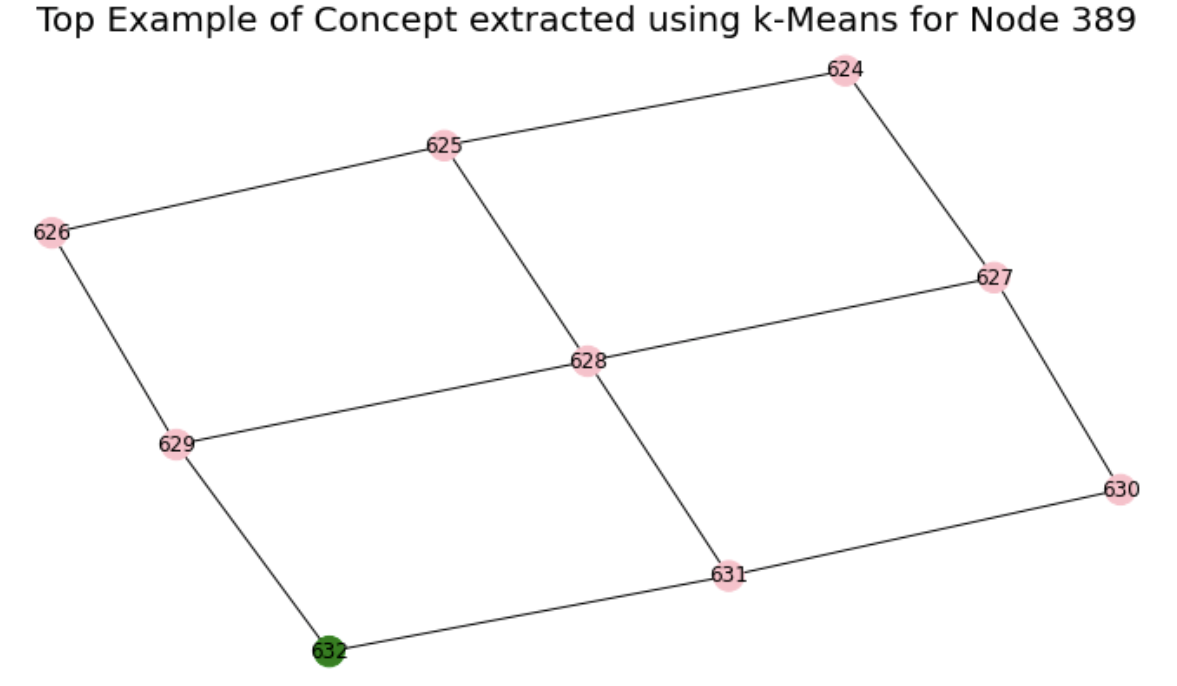}}
\caption{The concept representation representative of the cluster explaining a node in BA-Grid. The pink nodes are the neighbourhood of the clustered node in green.}
\label{ba_grid_exp1}
\end{center}
\vskip -0.2in
\end{figure}

\begin{figure}[ht]
\vskip 0.2in
\begin{center}
\centerline{\includegraphics[width=\columnwidth, trim={0 0 0 1cm},clip]{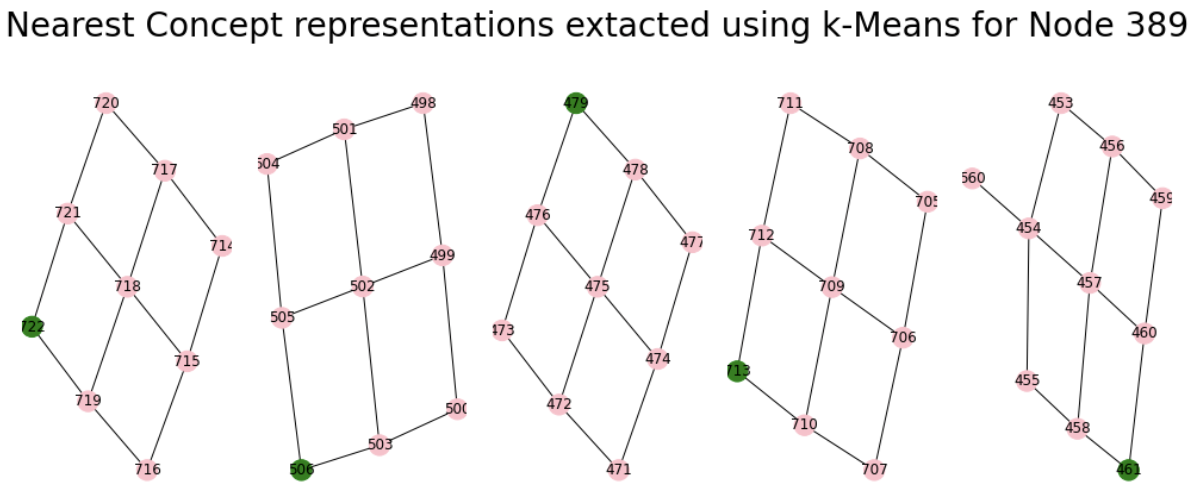}}
\caption{The concept representations nearest to the BA-Grid node 389 explained, visualised to capture cluster variance. The pink nodes are the neighbourhood of the clustered nodes in green.}
\label{ba_grid_exp2}
\end{center}
\vskip -0.2in
\end{figure}

\begin{figure}[ht]
\vskip 0.2in
\begin{center}
\centerline{\includegraphics[width=0.6\columnwidth, trim={0 0 0 1cm},clip]{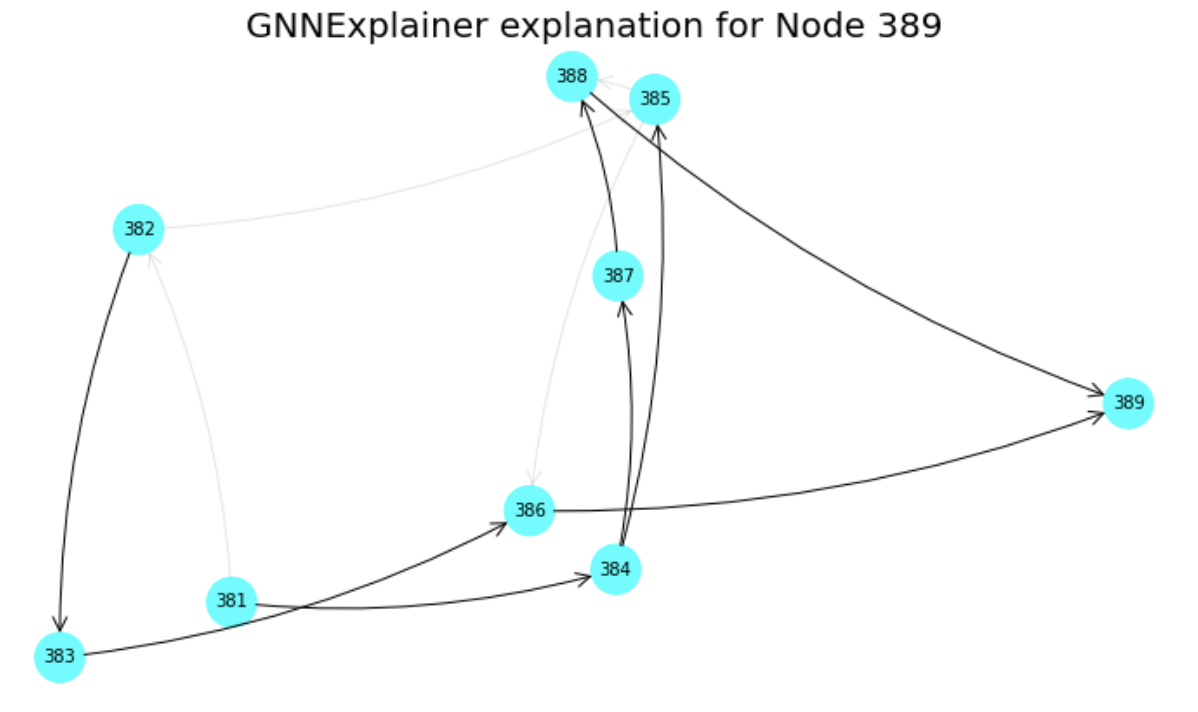}}
\caption{GNNExplainer explanation produced for BA-Grid node 389, where the turquoise nodes are part of the grid.}
\label{ba_grid_gnnexplainer}
\end{center}
\vskip -0.2in
\end{figure}

We also examine explanations produced for this dataset. For example, Figure \ref{ba_grid_exp1} and \ref{ba_grid_exp2} are explanations for the prediction of node 389, which is a node part of a grid. Across both visualisations, the grid structure is extracted as a concept. This shows the purity of the concept representation and indicates that the grid structure is important for the prediction of the node globally. In contrast, Figure \ref{ba_grid_gnnexplainer} visualises the explanation produced by GNNExplainer \cite{Ying2019}, which confirms that the grid structure plays an important role in the prediction of the node. In comparison with the concept-based explanations produced, the explanation produced by GNNExplainer is less intuitive to understand, as the whole motif is not visualised. 

\subsubsection{Tree-Cycles}

\begin{figure}[ht]
\vskip 0.2in
\begin{center}
\centerline{\includegraphics[width=\columnwidth, trim={2cm 0 2cm 2cm},clip]{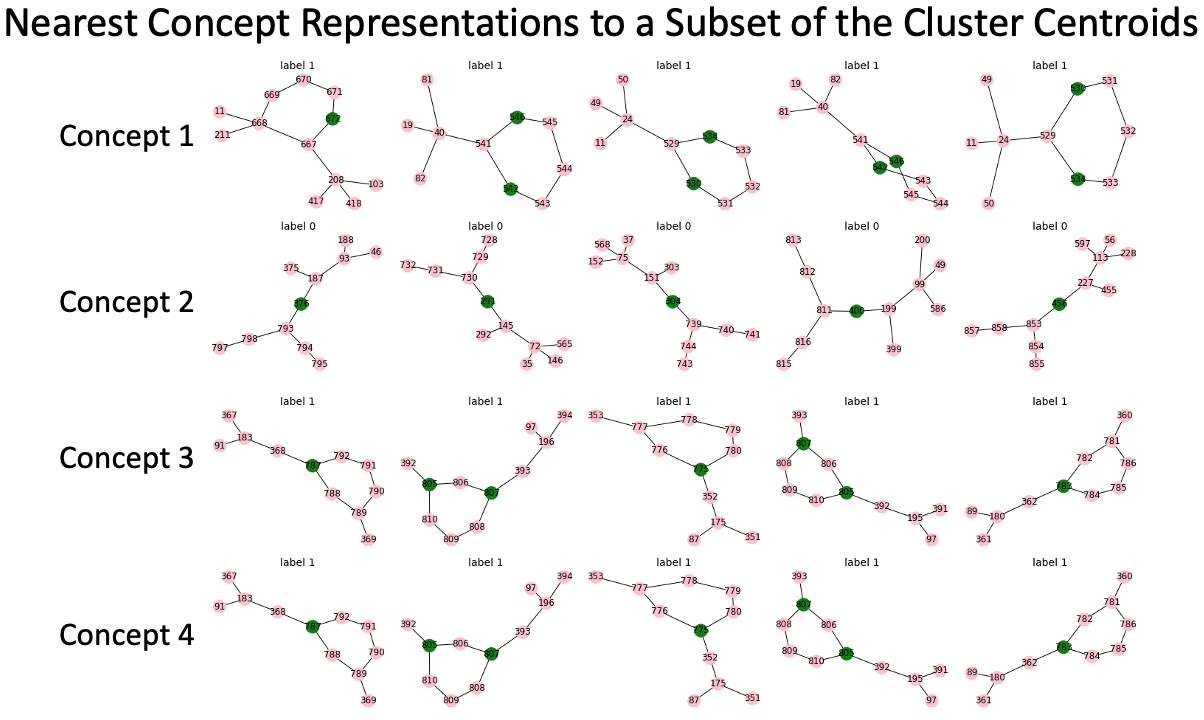}}
\caption{A subset of the concepts discovered for Tree-Cycles. Green nodes are the nodes clustered together, while the pink nodes form the expanded neighbourhood.}
\label{tree-cycle}
\end{center}
\vskip -0.2in
\end{figure}

We continue our evaluation by examining the concepts extracted for Tree-Cycles. Figure \ref{tree-cycle} visualises the concepts extracted in the same manner as before, however, we set $n = 3$ to accommodate for the larger network motif. Examining concept 1, 3, and 4 it can be stated that the cycle structure is extracted successfully. In contrast, concept 2 highlights the tree structure, which forms the base graph. The concepts extracted are coherent, which is highlighted when examining the nodes grouped together more closely. For example, in concept 1 multiple graphs have two green nodes. This is because the graph is invariant to translation, wherefore, the nodes can be seen as equivalent.

\begin{figure}[ht]
\vskip 0.2in
\begin{center}
\centerline{\includegraphics[width=0.6\columnwidth, trim={0 0 0 3cm},clip]{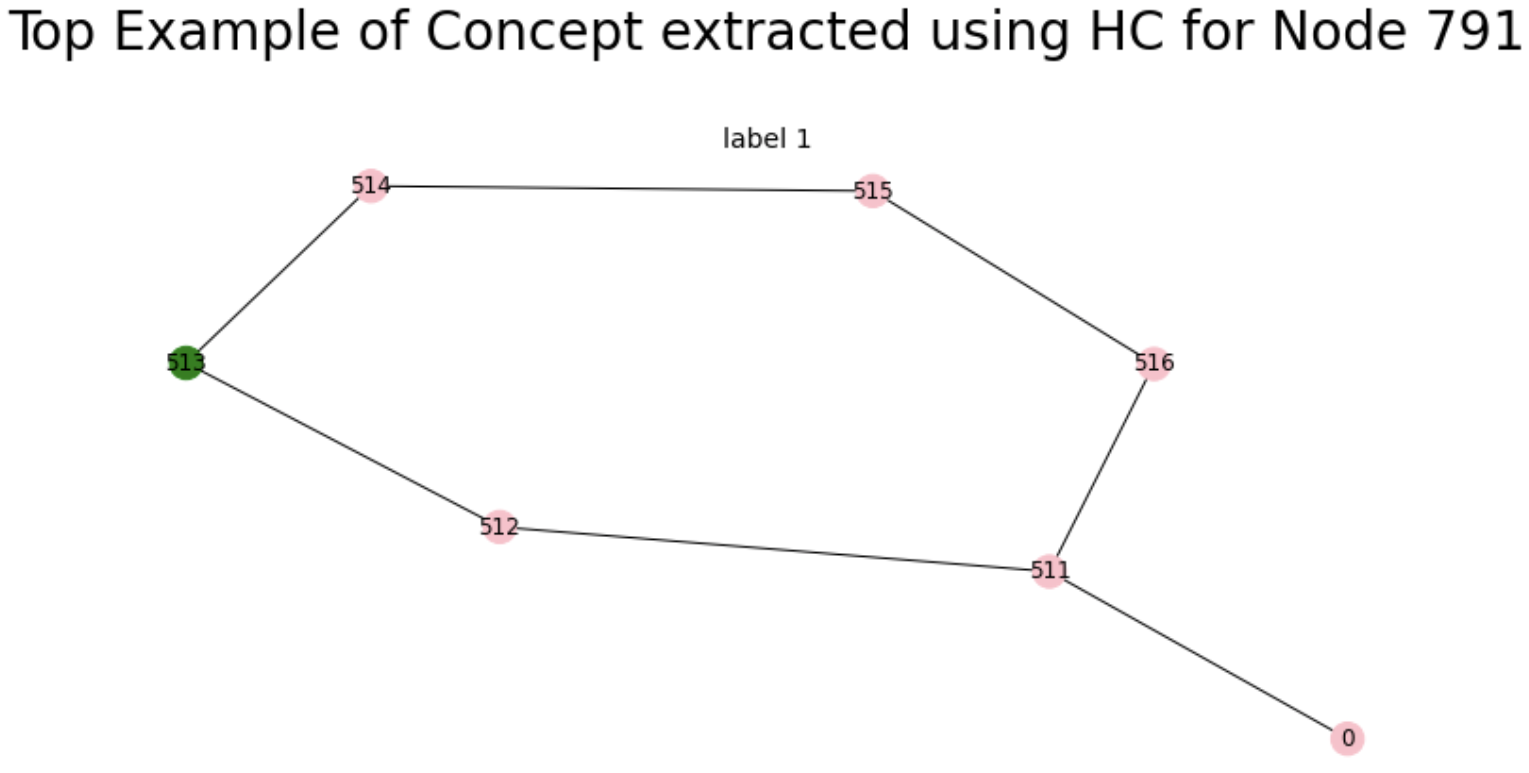}}
\caption{The concept representation representative of the cluster explaining a node in Tree-Cycles. The pink nodes are the neighbourhood of the clustered node in green.}
\label{tree_cycle_exp1}
\end{center}
\vskip -0.2in
\end{figure}

\begin{figure}[ht]
\vskip 0.2in
\begin{center}
\centerline{\includegraphics[width=\columnwidth, trim={0 0 0 2cm},clip]{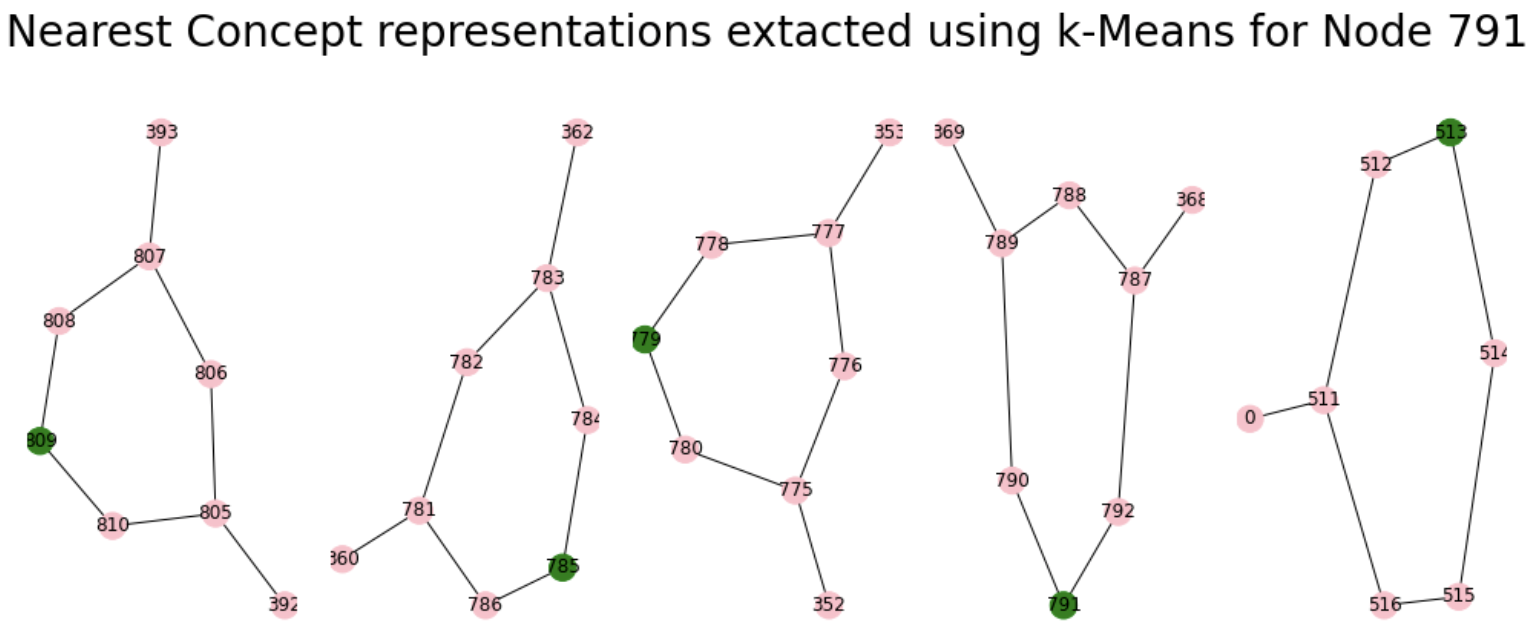}}
\caption{The concept representations nearest to the Tree-Cycles node 791 explained, visualised to capture cluster variance. The pink nodes are the neighbourhood of the clustered nodes in green.}
\label{tree_cycle_exp2}
\end{center}
\vskip -0.2in
\end{figure}

\begin{figure}[ht]
\vskip 0.2in
\begin{center}
\centerline{\includegraphics[width=0.6\columnwidth, trim={0 0 0 1cm},clip]{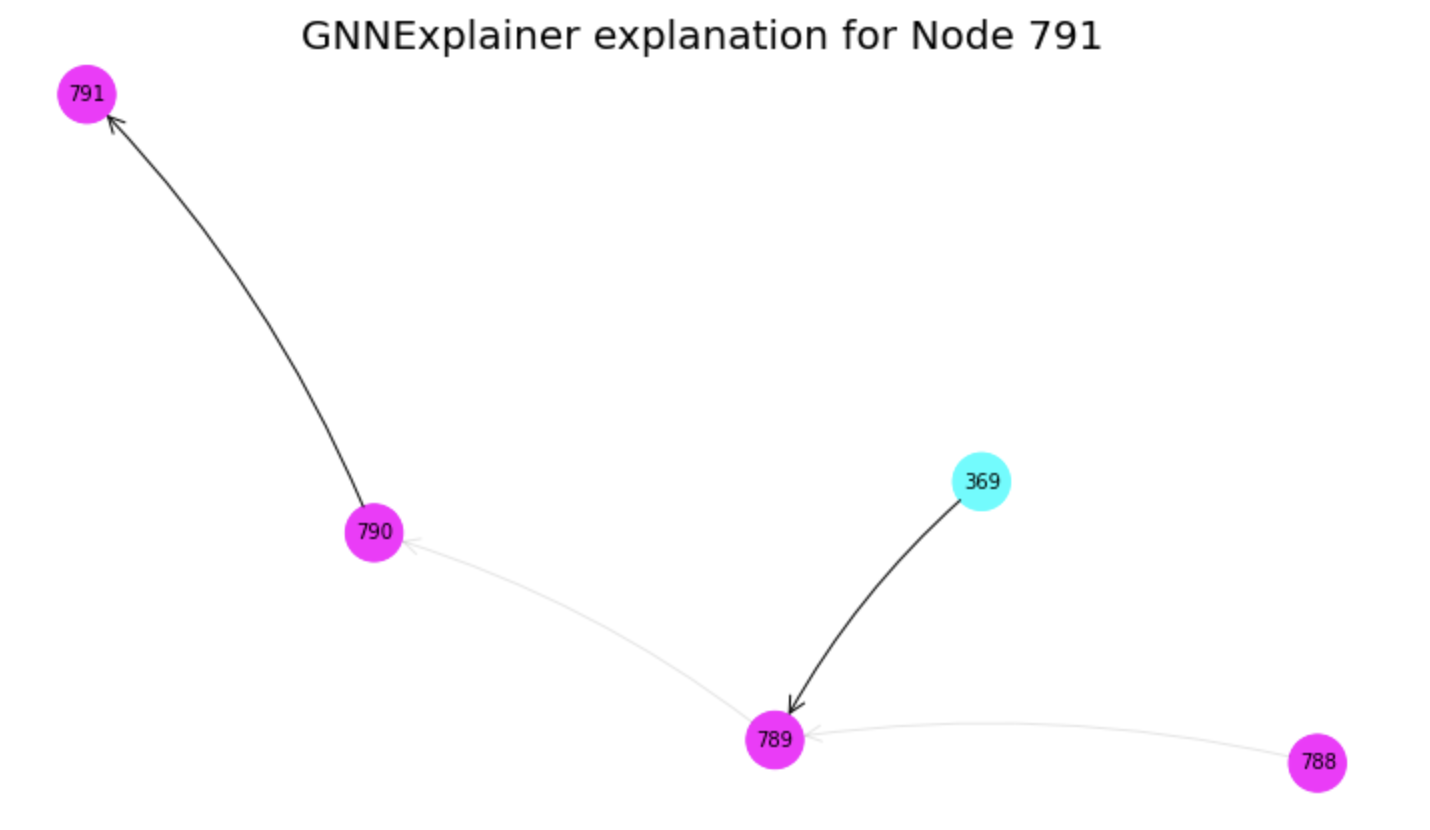}}
\caption{GNNExplainer explanation produced for Tree-Cycles node 791. The purple nodes are nodes part of the cycle structure.}
\label{tree_cycle_gnnexplainer}
\end{center}
\vskip -0.2in
\end{figure}

Figure \ref{tree_cycle_exp1} and \ref{tree_cycle_exp2} visualise the explanations produced. Both visualisations of the concept show that a the circular structure is identified as important for the prediction, as well as the edge attaching to the main graph. The additional edges displayed in Figure \ref{tree_cycle_exp2} are the random edges added to the graph. In comparison, the explanation produced by GNNExplainer \cite{Ying2019} does not depict the full cycle, as shown in Figure \ref{tree_cycle_gnnexplainer}. Only half of the cycle structure and the connecting edge are shown. While the explanation is more closely founded in the feature extraction performed by the GNN, the concept-based explanation is more intuitive to understand and allows recovering more fine-grained concepts than the coarse network motifs encoded in Tree-Cycles.

\subsubsection{Tree-Grid}

\begin{figure}[ht]
\vskip 0.2in
\begin{center}
\centerline{\includegraphics[width=\columnwidth, trim={2cm 0 2cm 2cm},clip]{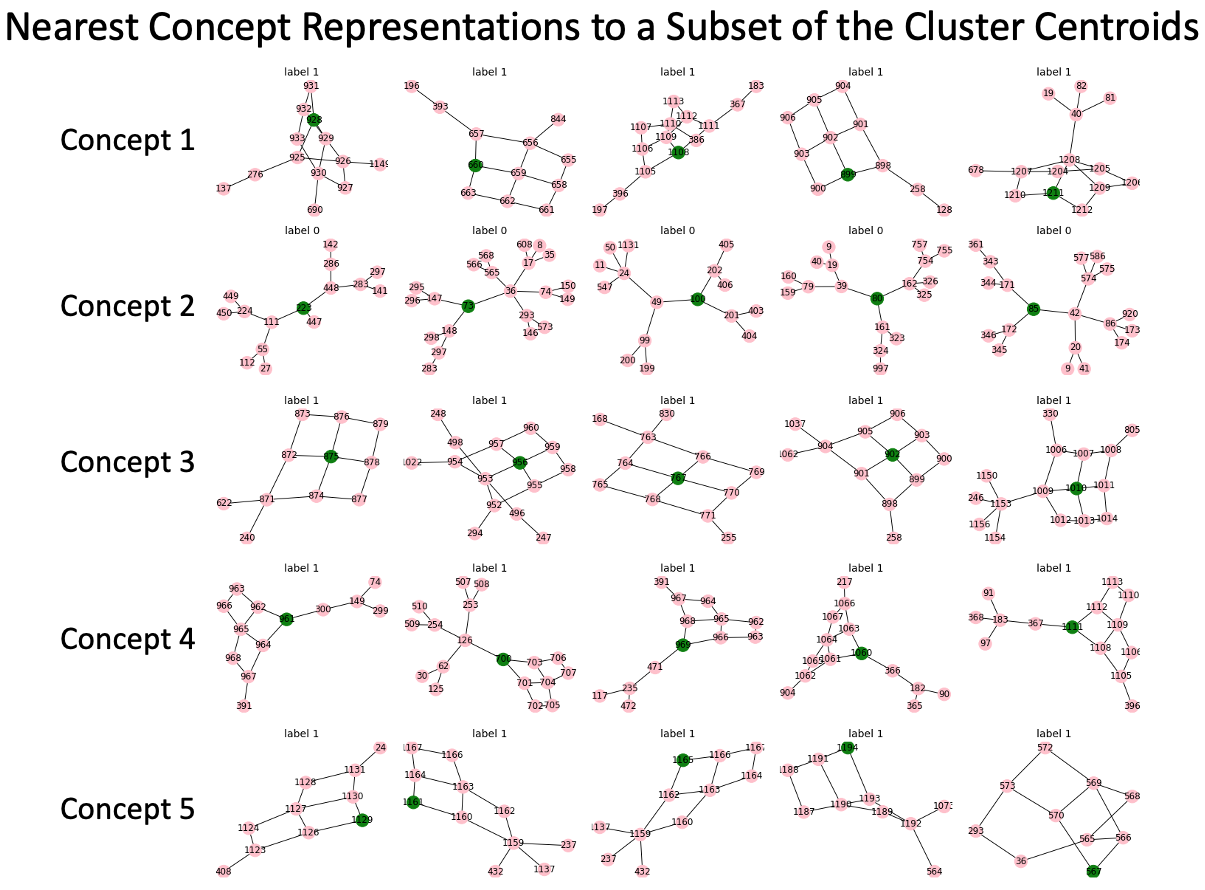}}
\caption{A subset of the concepts discovered for BA-Grid. Green nodes are the nodes clustered together, while the pink nodes form the expanded neighbourhood.}
\label{tree-grid}
\end{center}
\vskip -0.2in
\end{figure}

Figure \ref{tree-grid} visualises the top five concepts discovered for the Tree-Grid dataset. All concepts, except for concept 2, show the grid structure. The concept representation is more fine-grained than the basic grid structure, which shows in different nodes of the grid being highlighted. For example, concept 3 highlights the grid concept for the middle node, while concept 5 highlights the grid structure for a corner node. In general, it can be stated that the grid motif is successfully discovered as a concept and that the concept representations highlight the structure important for predictions. 

\begin{figure}[ht]
\vskip 0.2in
\begin{center}
\centerline{\includegraphics[width=0.6\columnwidth, trim={0 0 0 3cm},clip]{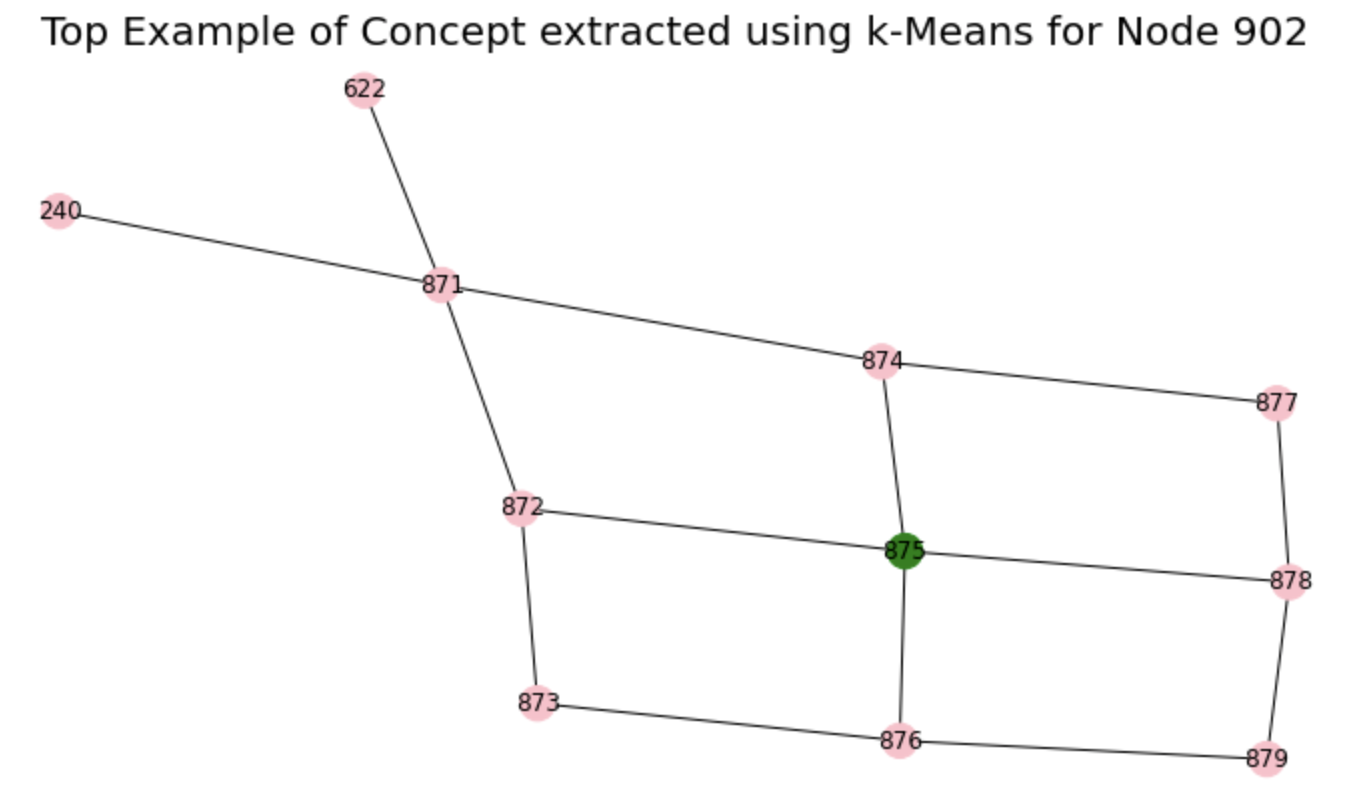}}
\caption{The concept representation representative of the cluster explaining a node in Tree-Grid. The pink nodes are the neighbourhood of the clustered node in green.}
\label{tree_grid_exp1}
\end{center}
\vskip -0.2in
\end{figure}

\begin{figure}[ht]
\vskip 0.2in
\begin{center}
\centerline{\includegraphics[width=\columnwidth, trim={0 0 0 3cm},clip]{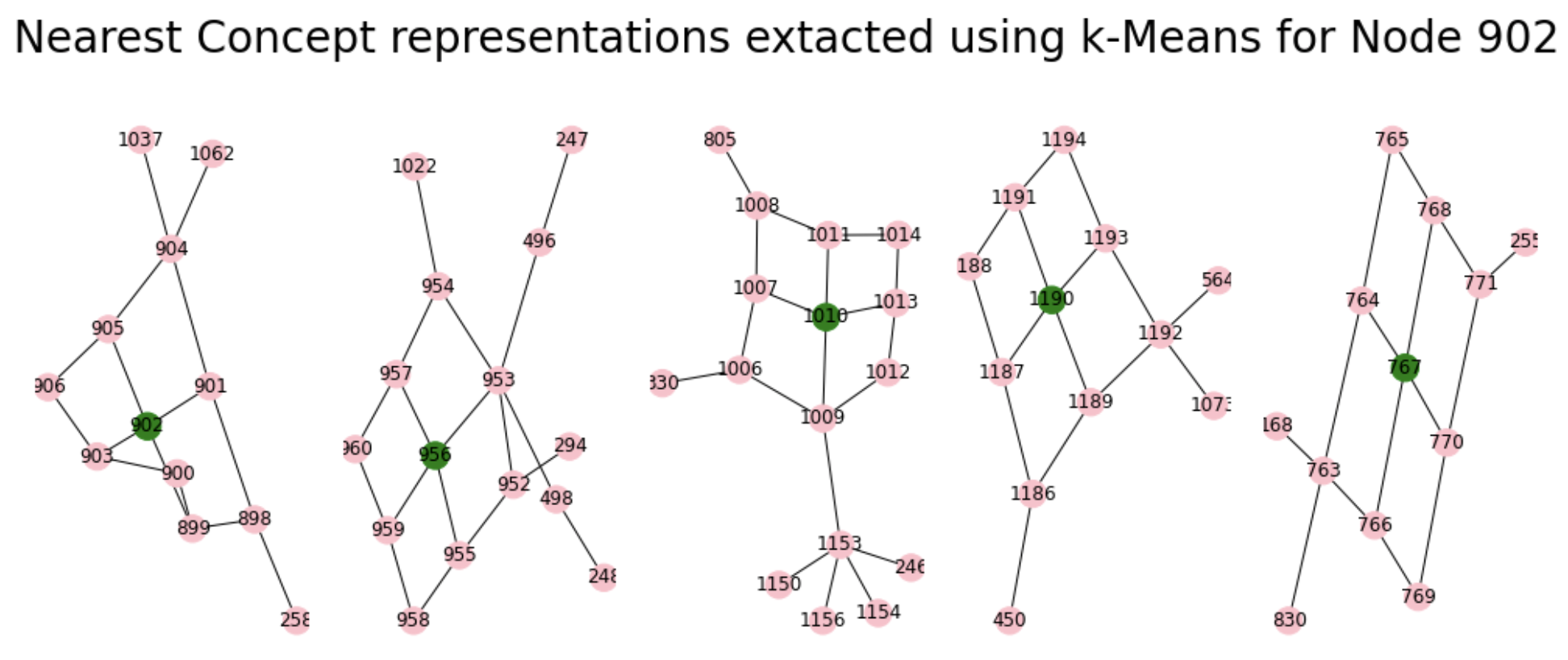}}
\caption{The concept representations nearest to the Tree-Grid node 901 explained, visualised to capture cluster variance. The pink nodes are the neighbourhood of the clustered nodes in green.}
\label{tree_grid_exp2}
\end{center}
\vskip -0.2in
\end{figure}

\begin{figure}[ht]
\vskip 0.2in
\begin{center}
\centerline{\includegraphics[width=0.6\columnwidth, trim={0 0 0 3cm},clip]{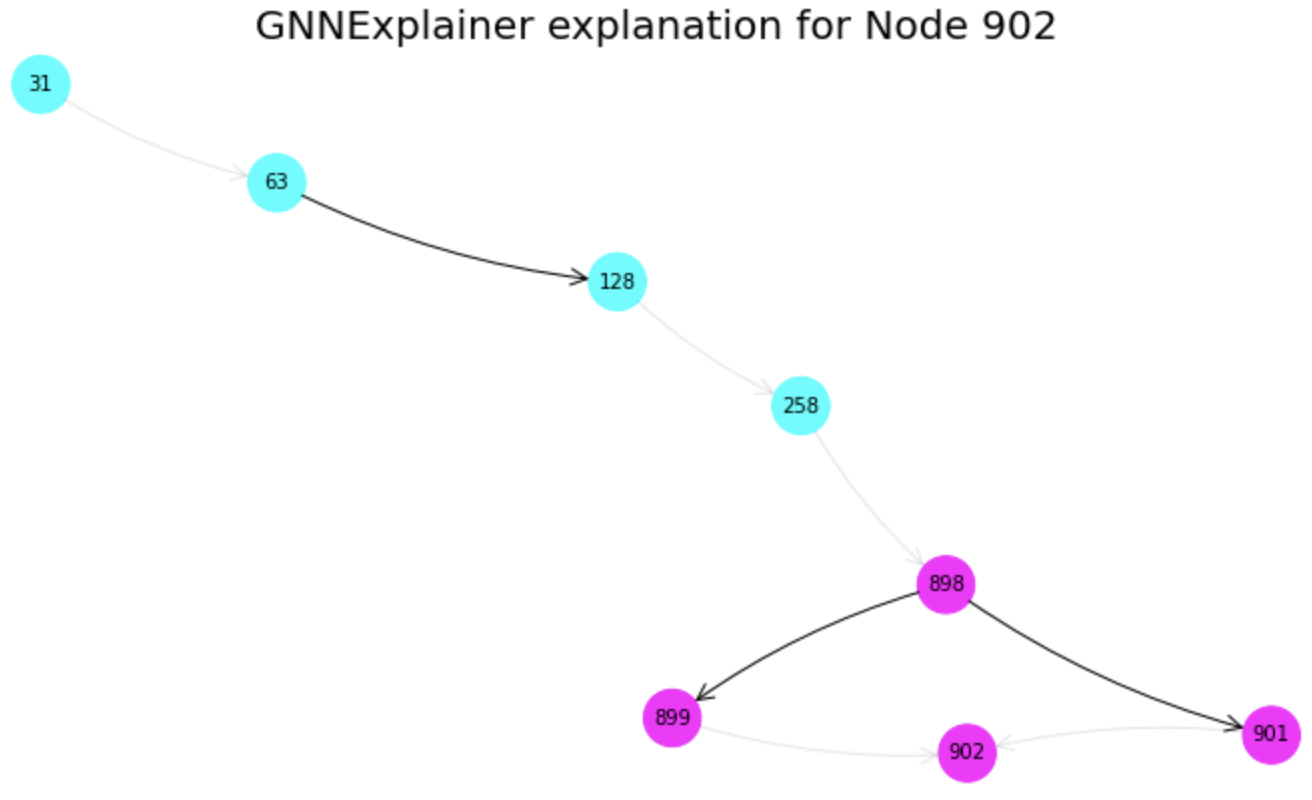}}
\caption{GNNExplainer explanation produced for Tree-Grid node 901, where nodes part of the grid are shown in purple.}
\label{tree_grid_gnnexplainer}
\end{center}
\vskip -0.2in
\end{figure}

Figures \ref{tree_grid_exp1} and \ref{tree_grid_exp2} show an explanation extracted for predicting the node 902. Figure \ref{tree_grid_exp1} shows the concept associated with the cluster, which is the grid structure with a middle node. Figure \ref{tree_grid_exp2} shows concept representations closer to the node being predicted. The structures presented are a little more varied, which can be attributed to the random edges in the graph. In general, the concept of a middle node in the grid is sufficient for explanation. In contrast, Figure \ref{tree_grid_gnnexplainer} shows the explanation produced by GNNExplainer \cite{Ying2019}, which only highlights a corner of the grid and more of the base graph. While the prediction shows the dependence of the nodes through directed edges, the proposed method provides a more coherent overview of similar nodes and allows to identify key structures in the dataset.

\subsubsection{Mutagenicity}

\begin{figure}[ht]
\vskip 0.2in
\begin{center}
\centerline{\includegraphics[width=\columnwidth, trim={3cm 0cm 3cm 2cm},clip]{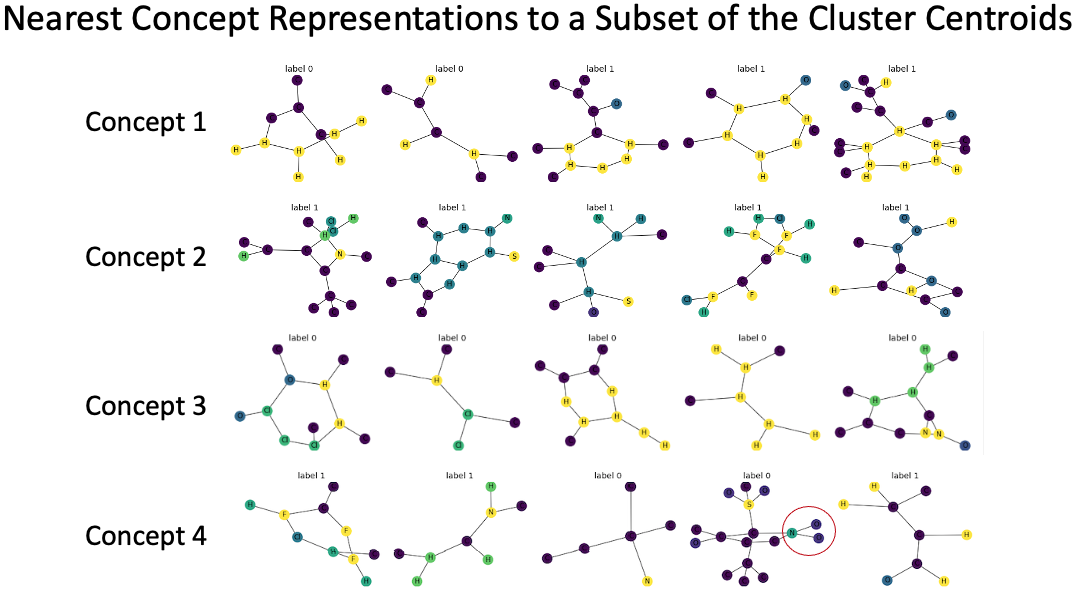}}
\caption{A subset of the concepts discovered for Mutagenicity. The nodes are labelled by the element they represent and coloured for distinction, however, the colouring scheme is not unique across concepts.}
\label{mutag}
\end{center}
\vskip -0.2in
\end{figure}

Figure \ref{mutag} shows a subset of the concepts extracted for Mutagenicity. It can be argued that the circular motif is recovered as a concept, but it cannot be assigned clearly. For example, concept 1 shows multiple examples of subgraphs with circular structures, however, the subgraphs are from graphs of different classes. Similarly, the circular structure can be observed in concept 2 and 3, however, concept 2 only shows examples of class 1, while concept 3 only shows examples of class 0. The other substructure $NO_2$ can only be observed in one subgraph visualised for concept 4. The motif is circled in red. In general, the concepts appear less pure than the ones extracted for the node classification datasets. There are two explanations for this. Firstly, it can be attributed to the lower accuracy of the model, which can be equated with less distinct clusters. Secondly, it could be indicative of an insufficient clustering and visualisation, which can be fine tuned by adapting $k$ and $n$, respectively.

\begin{figure}[ht]
\vskip 0.2in
\begin{center}
\centerline{\includegraphics[width=0.8\columnwidth, trim={0 0 0 2cm},clip]{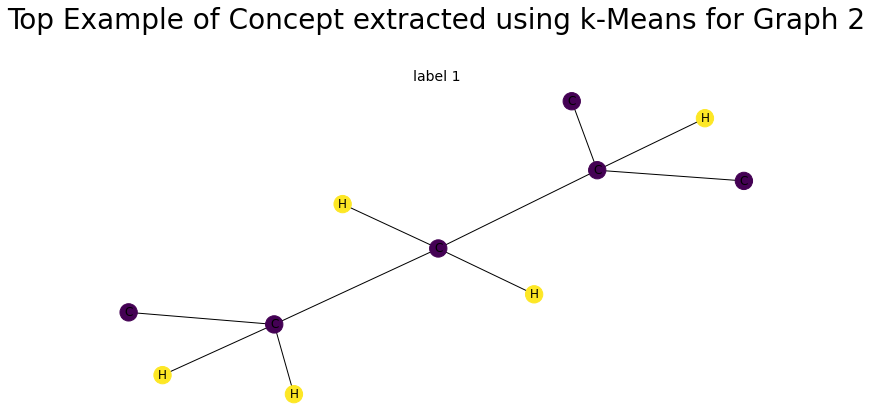}}
\caption{The concept with the highest representation in the graph explained. The nodes are labelled by the element they represent and coloured for distinction.}
\label{concept14}
\end{center}
\vskip -0.2in
\end{figure}

\begin{figure}[ht]
\vskip 0.2in
\begin{center}
\centerline{\includegraphics[width=0.8\columnwidth, trim={0 0 0 2cm},clip]{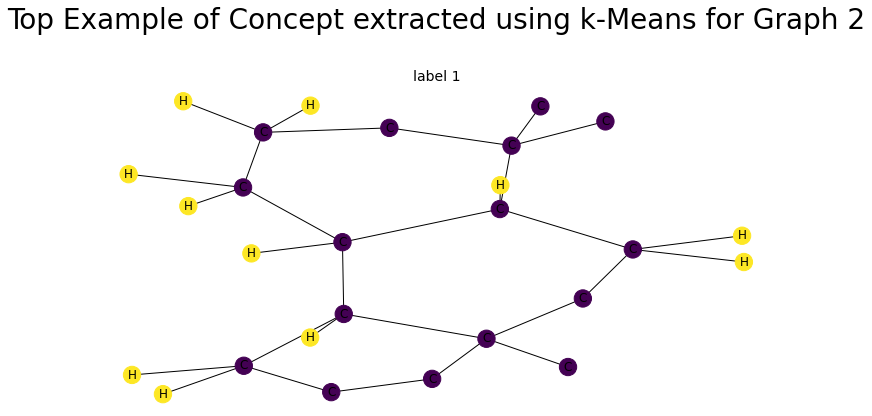}}
\caption{The second most dominant concept in the graph. The nodes are labelled by the element they represent and coloured for distinction.}
\label{concept5}
\end{center}
\vskip -0.2in
\end{figure}

As concept extraction is performed on the node level, an additional processing step is required to explain the prediction at the graph level. In order to reason about a full graph, we propose extracting the concept associated with each node and evaluating the contribution. For example, the graph chosen for explanation consists of 53 nodes. Across these 53 nodes, we identify 6 of the 30 concepts discovered applicable. The most dominant concept is concept 14, which is associated with 33 nodes. Figure \ref{concept14} visualises this concept. The concept highlights that a row of $C$ atoms with $H$ atoms attached are indicative of the prediction. Viewing this in conjunction with the second most dominant concept, visualised in Figure \ref{concept5}, it becomes evident that this forms a ring structure. 

\begin{figure}[ht]
\vskip 0.2in
\begin{center}
\centerline{\includegraphics[width=0.6\columnwidth, trim={0 0 0 1cm},clip]{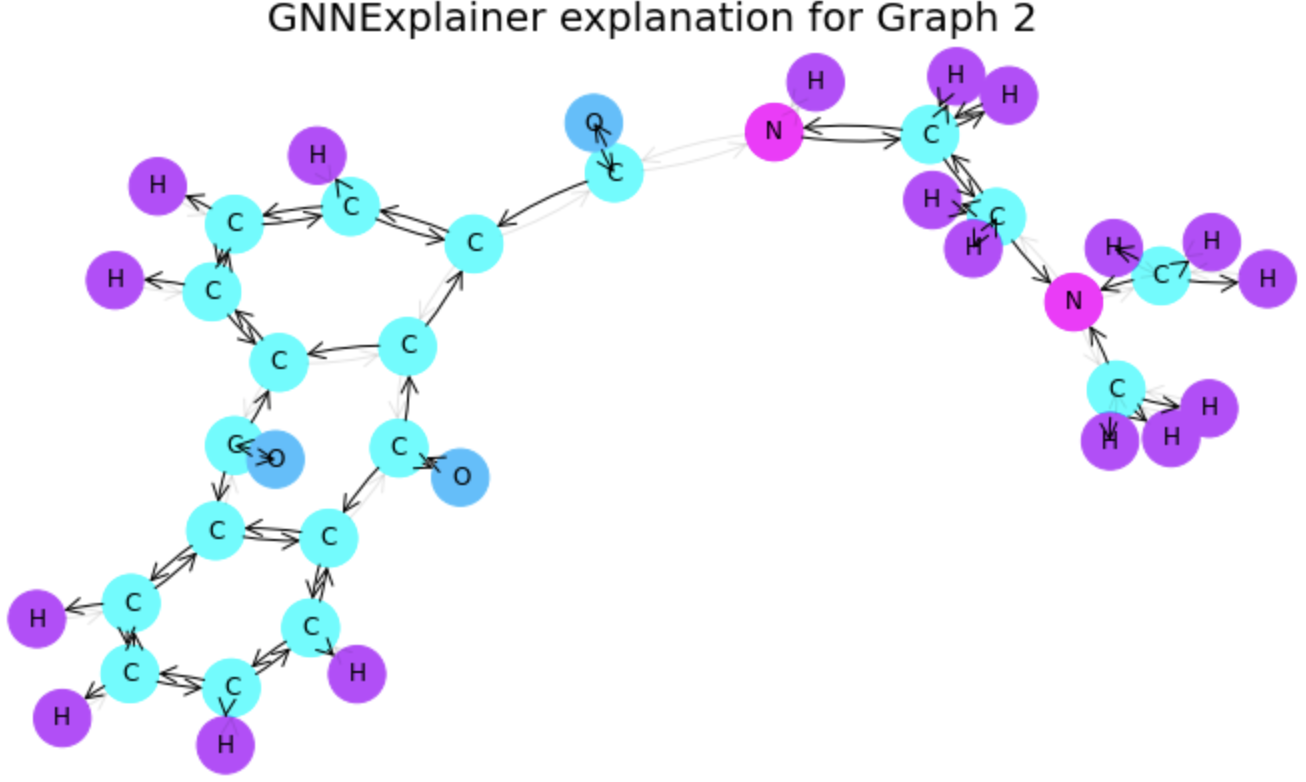}}
\caption{Explanation produced for Graph 2 using GNNExplainer. The nodes are labelled by the element they represent and coloured for distinction. The colour is chosen independently by GNNExplainer.}
\label{mutag_gnnexplainer}
\end{center}
\vskip -0.2in
\end{figure}

As the method is focused on extracting node-level concepts, the explanation produced cannot be compared directly with that of GNNExplainer \cite{Ying2019}. GNNExplainer provides a full subgraph, as shown in Figure \ref{mutag_gnnexplainer}. The explanation matches ours in some parts. GNNExplainer also identifies the ring structures as important, however, also shows further nodes. GNNExplainer provides a better overview of the dependency of the nodes in the computation for the prediction, however, the concept-based explanations extracted are easier to understand, as they focus the user on a smaller subgraph. Furthermore, a larger subgraph can be displayed by increasing the parameter $n$ used in the $n$-hop visualisation of neighbours. This is an advantage of our technique, as it allows for exploration by the user.

\subsubsection{REDDIT-BINARY}

\begin{figure}[ht]
\vskip 0.2in
\begin{center}
\centerline{\includegraphics[width=\columnwidth, trim={3cm 0cm 3cm 2cm},clip]{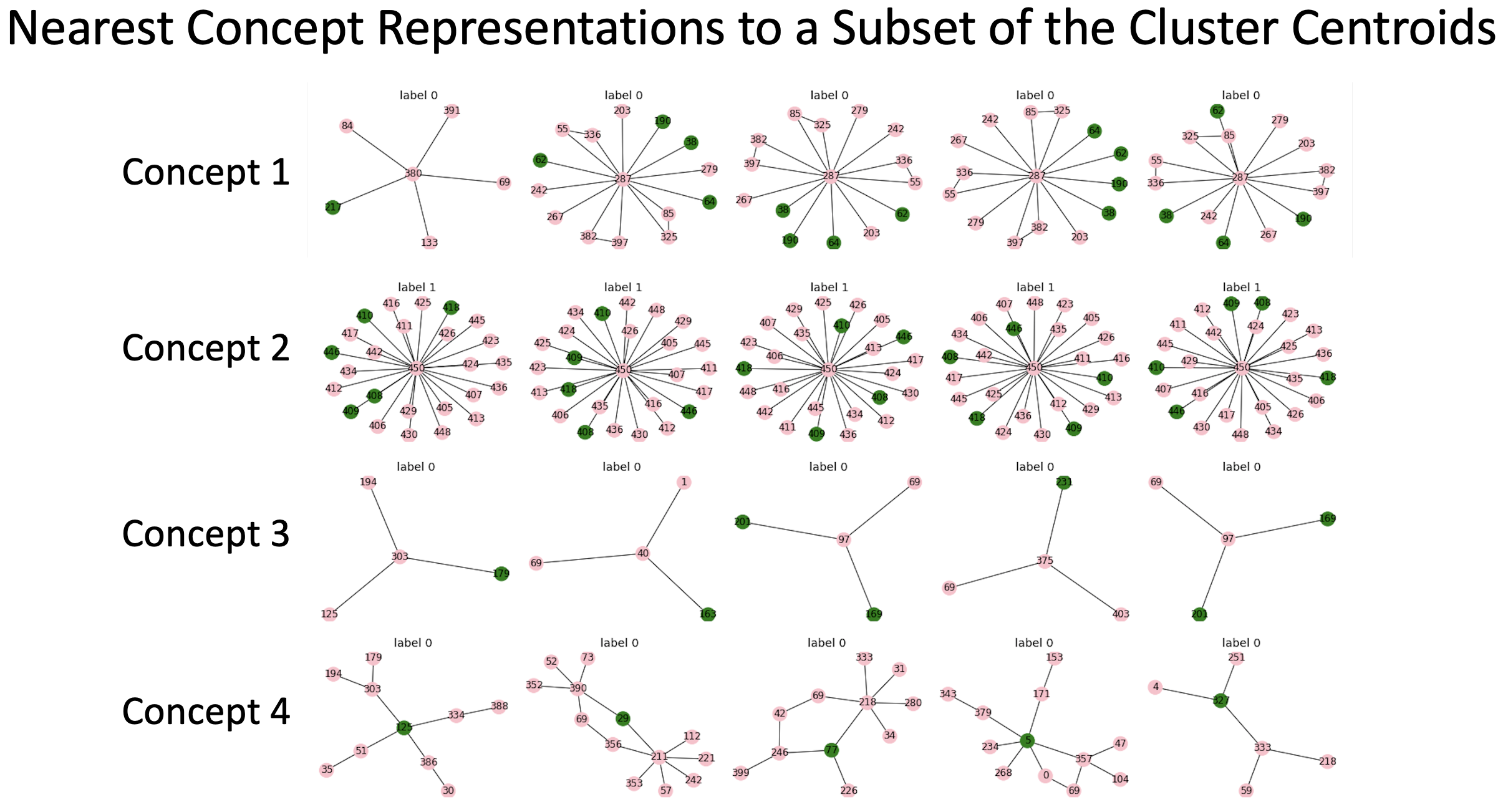}}
\caption{A subset of the concepts discovered for REDDIT-BINARY. The green nodes are the nodes clustered together from which neighbourhood expansion is performed (pink nodes).}
\label{reddit}
\end{center}
\vskip -0.2in
\end{figure}

We perform the same concept discovery and extraction for REDDIT-BINARY. Figure \ref{reddit} shows the concepts discovered. Concepts 1, 3 and 4 represent the expert response to a question. While the concepts discovered do not exactly match the motif, they are representative of the interactions. For example, the middle node in the star structure displayed as part of concept 3 can be identified as the expert responding to multiple questions. A differentiation must be made between concepts 1 and 3, and concept 4. In concept 4, the nodes grouped are the experts responding to questions, while in the other concepts the green nodes are questions. This is evident from their connectivity. In contrast, concept 2 shows the motif identifying reactions to a topic, where the green nodes are examples of users reacting to a topic. It looks very similar to concept 1, however, it can be differentiated by the class label displayed above. In conclusion, the desired motifs are successfully discovered and give an insight into the types of interactions on a global level.

\begin{figure}[ht]
\vskip 0.2in
\begin{center}
\centerline{\includegraphics[width=0.9\columnwidth, trim={0 0 0 2cm},clip]{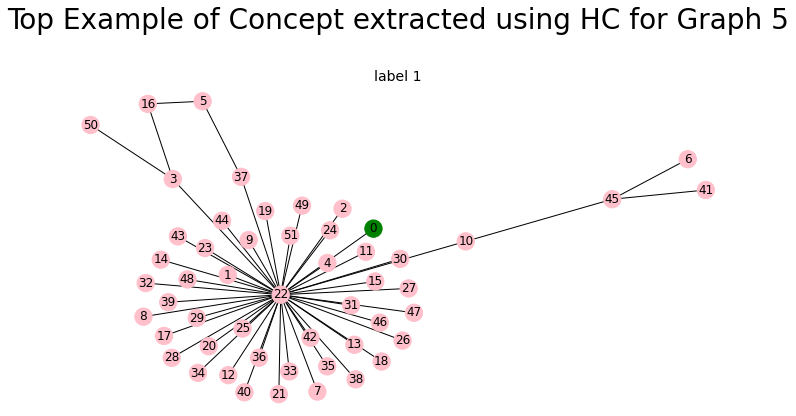}}
\caption{The concept with the highest representation in the graph explained. The green node represents the node this explanation is produced for, while, the pink nodes form the neighbourhood. However, as this is the strongest concept in the graph it applies to further surrounding nodes, such as nodes 2, 24 and 51.}
\label{redditconcept}
\end{center}
\vskip -0.2in
\end{figure}

\begin{figure}[ht]
\vskip 0.2in
\begin{center}
\centerline{\includegraphics[width=0.7\columnwidth, trim={0 0 0 1cm},clip]{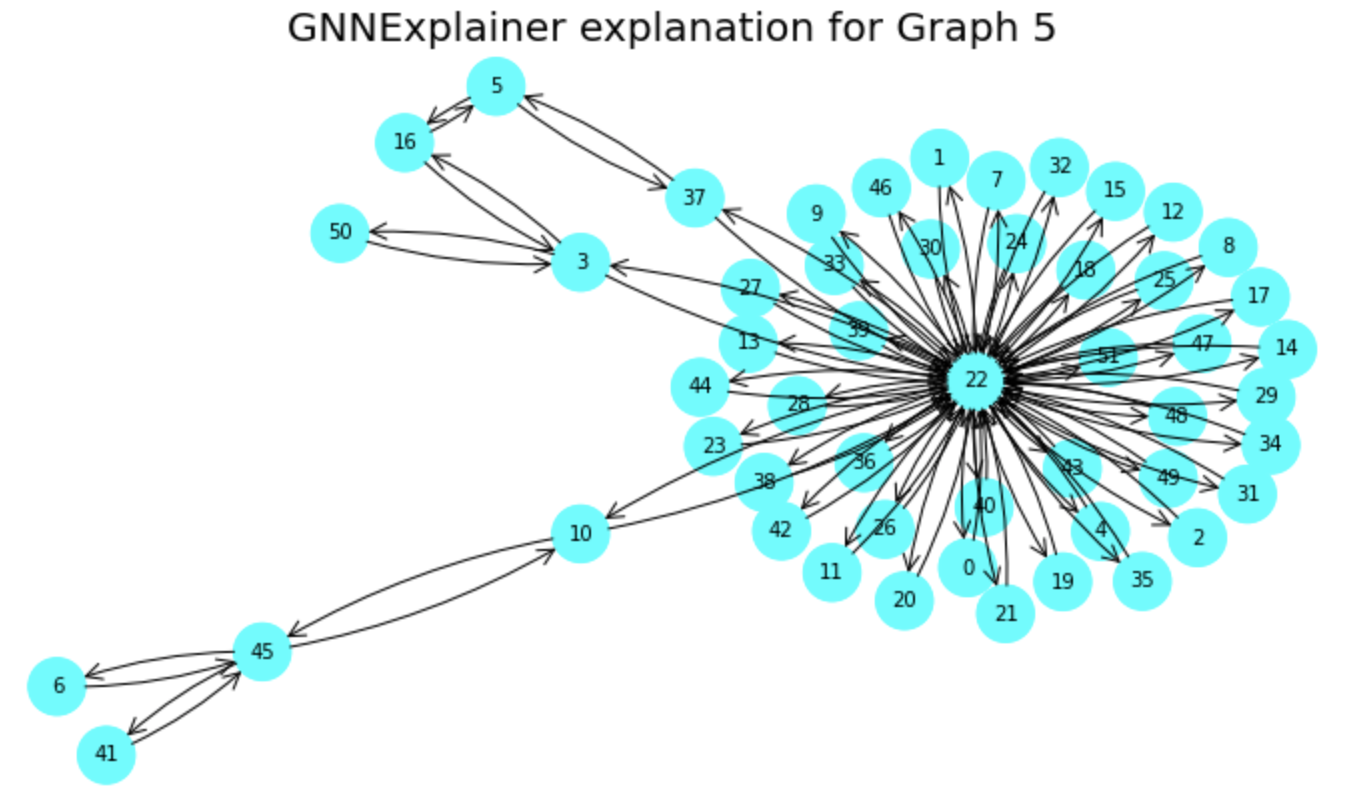}}
\caption{Explanation produced for Graph 5 using GNNExplainer. The colouring is chosen by GNNExplainer and carries no further meaning.}
\label{reddit_gnnexplainer}
\end{center}
\vskip -0.2in
\end{figure}

We also explain a graph chosen at random for REDDIT-BINARY. Figure \ref{redditconcept} displays the concept with the highest representation, which is 42 out of 52 nodes. The concept shows users reacting to a topic. The green node can be inferred as a user, who is reacting to a topic represented by node 22. As we can observe that a large number of nodes have the same connectivity as the green node, the strong presence of this concept is explained. GNNExplainer produces a similar looking explanation, as shown in Figure \ref{reddit_gnnexplainer}, but the explanation is local, while the concept-based explanation is applicable across graphs in the dataset. 

\subsubsection{Key Results}

In conclusion, the proposed method allows to successfully discover and extract concepts important for predictions on a global level. The method allows to extract fine-grained concepts and highlights structural similarities that are of importance for the prediction of similar nodes. The concept-based explanations are easier to understand as there are multiple examples, rather than a customised explanation that only applies to a single instance. However, the method appears to perform less well on graph classification tasks as clear concepts are harder to identify and require more reasoning by the user. Nevertheless, the global explanations allow a unique insight into the dataset.




\end{document}